\documentclass[screen]{acmart}

\AtBeginDocument{%
  \providecommand\BibTeX{{%
    \normalfont B\kern-0.5em{\scshape i\kern-0.25em b}\kern-0.8em\TeX}}}

\acmDOI{10.1145/1122445.1122456}

\setcopyright{acmcopyright}
\acmJournal{TOIS}
\acmYear{2020} \acmVolume{1} \acmNumber{1} \acmArticle{1} \acmMonth{1} \acmPrice{15.00}

\acmPrice{15.00}
\usepackage{booktabs}
\usepackage{subfigure}
\usepackage{color}
\usepackage{multirow}

\newcommand{\ie}{\emph{i.e.,}}
\newcommand{\aka}{\emph{a.k.a.,}}
\newcommand{\eg}{\emph{e.g.,}}

\newcommand{\ignore}[1]{}

\newcommand{\dubbelop}{$^{\blacktriangle}$}

\newcommand{\dubbelneer}{$^{\blacktriangledown}$}

\begin{document}

\title{Learning to Respond with Your Favorite Stickers: A Framework of \\ Unifying Multi-Modality and User Preference in Multi-Turn Dialog}

\author{Shen Gao}
\affiliation{%
  \institution{Wangxuan Institute of Computer Technology, Peking University}
}
\email{shengao@pku.edu.cn}

\author{Xiuying Chen}
\affiliation{%
  \institution{Wangxuan Institute of Computer Technology, Peking University}
}
\email{xy-chen@pku.edu.cn}

\author{Li Liu}
\affiliation{%
  \institution{Inception Institute of Artificial Intelligence}
}
\email{liuli1213@gmail.com}

\author{Dongyan Zhao}
\affiliation{%
	\institution{Wangxuan Institute of Computer Technology, Peking University}
}
\email{zhaody@pku.edu.cn}

\author{Rui Yan}
\authornote{Corresponding Author: Rui Yan (ruiyan@pku.edu.cn)}
\affiliation{
  \institution{\textsuperscript{1} Gaoling School of Artificial Intelligence, Renmin University of China; 
  \textsuperscript{2} Wangxuan Institute of Computer Technology, Peking University}
}
\email{ruiyan@pku.edu.cn}

\renewcommand{\shortauthors}{Shen Gao, et al.}

\begin{abstract}
  Stickers with vivid and engaging expressions are becoming increasingly popular in online messaging apps, and some works are dedicated to automatically select sticker response by matching the stickers image with previous utterances.
  However, existing methods usually focus on measuring the matching degree between the dialog context and sticker image, which ignores the user preference of using stickers.
  Hence, in this paper, we propose to recommend an appropriate sticker to user based on multi-turn dialog context and sticker using history of user.
  Two main challenges are confronted in this task.
  One is to model the sticker preference of user based on the previous sticker selection history.
  Another challenge is to jointly fuse the user preference and the matching between dialog context and candidate sticker into final prediction making.
  To tackle these challenges, we propose a \emph{Preference Enhanced Sticker Response Selector} (PESRS) model.
  Specifically, PESRS first employs a convolutional based sticker image encoder and a self-attention based multi-turn dialog encoder to obtain the representation of stickers and utterances.
  Next, deep interaction network is proposed to conduct deep matching between the sticker and each utterance.
  Then, we model the user preference by using the recently selected stickers as input, and use a key-value memory network to store the preference representation.
  PESRS then learns the short-term and long-term dependency between all interaction results by a fusion network, and dynamically fuse the user preference representation into the final sticker selection prediction.
  Extensive experiments conducted on a large-scale real-world dialog dataset show that our model achieves the state-of-the-art performance for all commonly-used metrics.
  Experiments also verify the effectiveness of each component of PESRS.
\end{abstract}



\maketitle

\section{Introduction}
\label{sec:intro}
\begin{figure}
    \centering
    \includegraphics[scale=0.15]{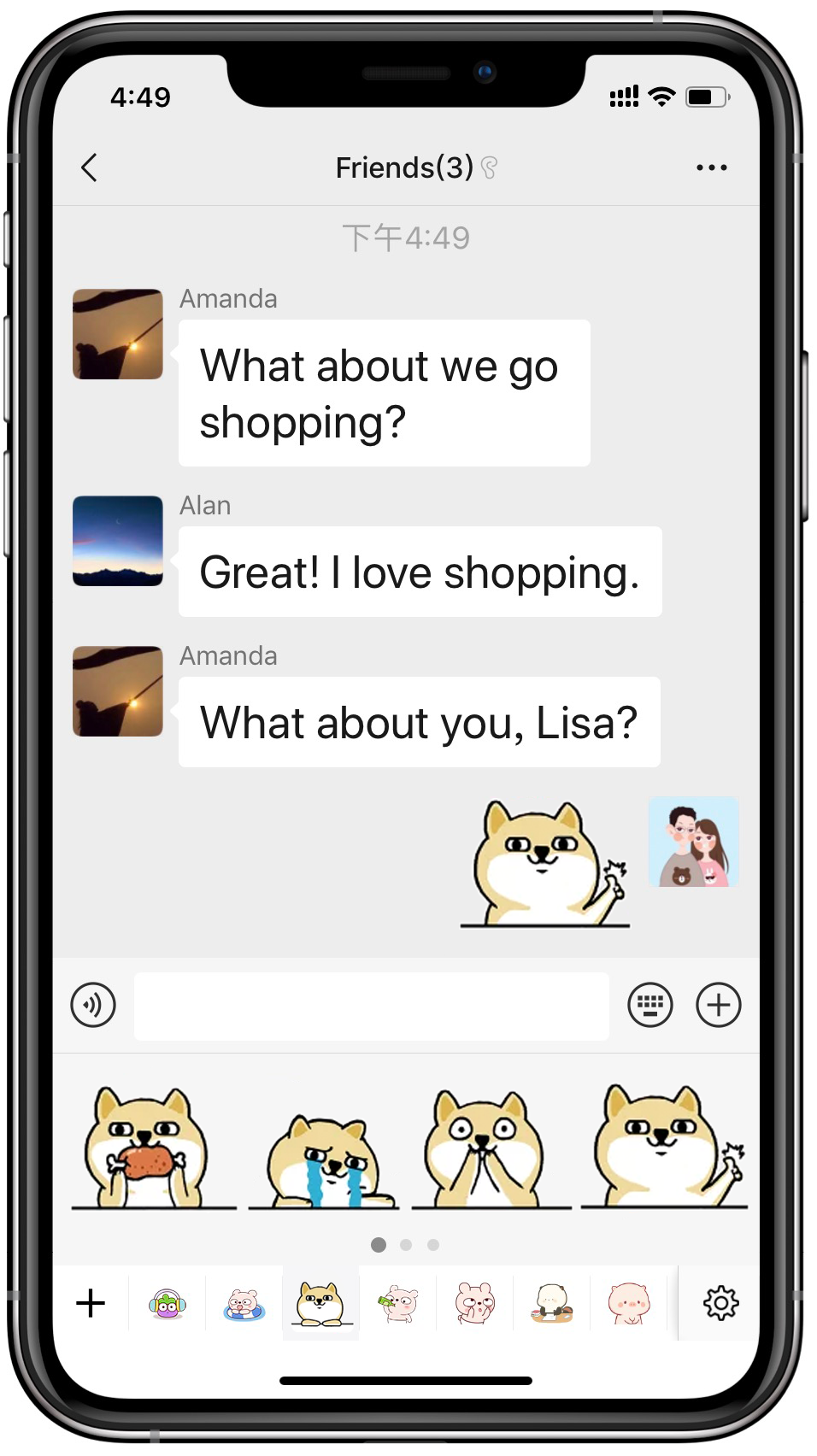}
    \caption{
         An example of stickers in a multi-turn dialog. Sticker response selector automatically selects the proper sticker based on multi-turn dialog history. 
    }
    \label{fig:example}
\end{figure}

Images (\aka graphicon) are another important approach for expressing feelings and emotions in addition to using text in communication.
In mobile messaging apps, these images can generally be classified into emojis and stickers.
Emoji is a kind of small picture which is already stored in most of the keyboard of the mobile operational systems, \ie iOS or Android.
Emojis are pre-designed by the mobile phone vendor (now it is managed by standards organization) and the number of emoji is limited, and users can not design emoji by themselves.
Different with the inflexible emojis, sticker is image or graphicon essentially~\cite{Seta2018BiaoqingTC,Herring2017NicePC,Ge2018CommunicativeFO}, which users can draw or modify images as a sticker and upload to the chatting app by themselves.
The using of stickers on online chatting usually brings diversity of expressing emotion.
Since emojis are sometimes used to help reinforce simple emotions in a text message due to their small size, and their variety is limited.
Stickers, on the other hand, can be regarded as an alternative for text messages, which usually include cartoon characters and are of high definition.
They can express much more complex and vivid emotion than emojis.
Most messaging apps, such as WeChat, Telegram, WhatsApp, and Slack provide convenient ways for users to download stickers for free, or even share self-designed ones.
We show a chat window including stickers in Figure~\ref{fig:example}. 

Stickers are becoming more and more popular in online chat.
First, sending a sticker with a single click is much more convenient than typing text on the 26-letter keyboard of a small mobile phone screen.
Second, there are many implicit or strong emotions that are difficult to express in words but can be captured by stickers with vivid facial expressions and body language.
However, the large scale use of stickers means that it is not always straightforward to think of the sticker that best expresses one's feeling according to the current chatting context.
Users need to recall all the stickers they have collected and selected the appropriate one, which is both difficult and time-consuming.

Consequently, much research has focused on recommending appropriate emojis to users according to the chatting context.
Existing works such as~\cite{xie2016neural}, are mostly based on emoji recommendation, where they predict the probable emoji given the contextual information from multi-turn dialog systems.
In contrast, other works~\cite{barbieri2017emojis,barbieri2018multimodal} recommend emojis based on the text and images posted by a user.
As for sticker recommendation, existing works such as~\cite{laddha2019understanding} and apps like Hike or QQ directly match the text typed by the user to the short text tag assigned to each sticker.
However, since there are lots of ways of expressing the same emotion, it is very hard to capture all variants of an utterance as tags.

To overcome the drawbacks, we propose a sticker response selector (SRS) for sticker selection in our early work~\cite{gao2020sticker}, where we address the task of sticker response selection in multi-turn dialog.
We focus on the two main challenges in this work:
(1) Since existing image recognition methods are mostly built with real-world images, and how to capture the semantic meaning of sticker is challenging.
(2) Understanding multi-turn dialog history information is crucial for sticker recommendation, and jointly modeling the candidate sticker with multi-turn dialog is challenging.
Herein, we propose a novel sticker recommendation model, namely \emph{sticker response selector} (SRS), for sticker response selection in multi-turn dialog.
Specifically, SRS first learns representations of dialog context history using a self-attention mechanism and learns the sticker representation by a convolutional neural network (CNN). 
Next, SRS conducts deep matching between the sticker and each utterance and produces the interaction results for every utterance.
Finally, SRS employs a fusion network which consists of a sub-network fusion RNN and fusion transformer to learn the short and long term dependency of the utterance interaction results.
The final matching score is calculated by an interaction function.
To evaluate the performance of our model, we propose a large number of multi-turn dialog dataset associated with stickers from one of the popular messaging apps. 
Extensive experiments conducted on this dataset show that SRS significantly outperforms the state-of-the-art baseline methods in commonly-used metrics.

\begin{figure}[h]
    \centering
    \includegraphics[scale=0.15]{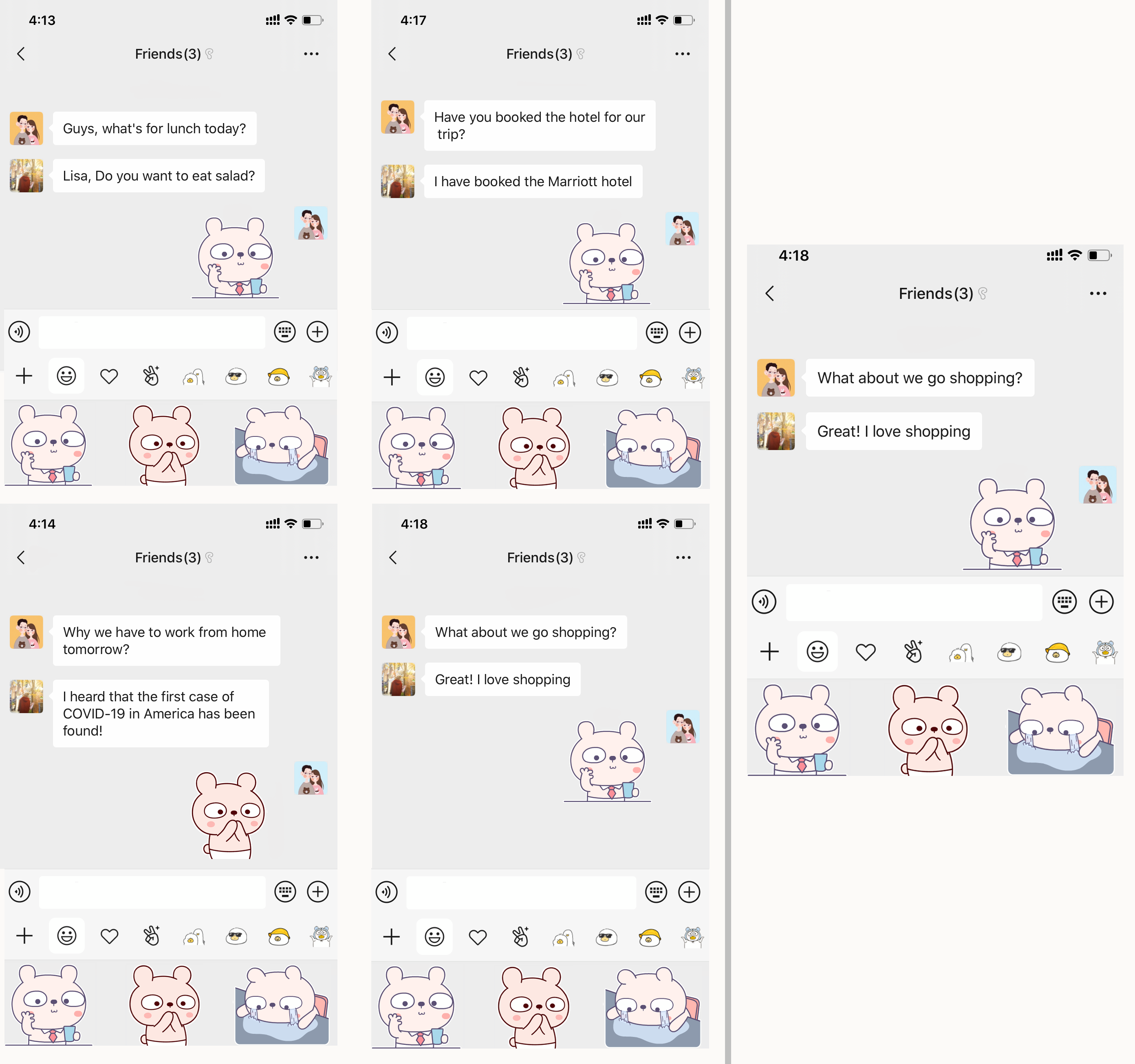}
    \caption{
        User's history dialog context and the selected sticker. Four figures in the left history dialog context and the selected sticker, and the right one the current dialog context with the user selected sticker. User tends to use the same sticker when the dialog context is semantically similar. 
    }
    \label{fig:user-preference-case}
\end{figure}

However, the user's sticker selection does not only depend on the matching degree between dialog context and candidate sticker image, but also depends on the user's preference of using sticker.
When users decide to use a sticker as their response in multi-turn dialog, they may choose their favorite one from all appropriate stickers as the final response. 
We assume that user tends to use the recently used sticker in their dialog history, and the recently-used-sticker can represent the user's preference of sticker selection.
An example is shown in Figure~\ref{fig:user-preference-case}.
To verify this assumption, we retrieve 10 recently-used-stickers of each user and calculate the proportion of whether the currently used sticker appeared in these 10 stickers.
The result shows that 54.09\% of the stickers exist in the 10 recently used sticker set.
Hence, we reach to the conclusion that users have strong personal preference when selecting the sticker as their response for the current dialog context.
However, in some cases, this also indicates a tendency to re-use stickers, but not necessarily a preference.

Motivated by this observation, in this work, we take one step further and improve our previously proposed SRS framework with user preference modeling.
Overall, we propose a novel sticker recommendation model which considers the user preference, namely \emph{Preference Enhanced Sticker Response Selector} (PESRS).
Specifically, PESRS first employs a convolutional network to extract features from the candidate stickers.
Then, we retrieve the recent user sticker selections then a user preference modeling module is employed to obtain a user preference representation.
Next, we conduct the deep matching between the candidate sticker and each utterance as the same as SRS.
Finally, we use a gated fusion method to combine the deep matching result and user preference into final sticker prediction.

The key to the success of PESRS lies in how to design the user preference modeling module, which should not only identify the user's favorite sticker and but also consider the current dialog context.
Motivated by this, we first propose a recurrent neural network (RNN) based position-aware sticker modeling module which encodes the recently used stickers in chronological order.
Then, we employ a key-value memory network to store these sticker representations as values and the corresponding dialog context as keys.
Finally, we use the current dialog context to query the key-value memory and obtain the dynamic user preference of the current dialog context.

We empirically compare PESRS and SRS on the public dataset\footnote{https://github.com/gsh199449/stickerchat} proposed by our early work~\cite{gao2020sticker}.
This is a large-scale real-world Chinese multi-turn dialog dataset, which dialog context is multiple text utterances and the response is a sticker image.
Experimental results show that on this dataset, our newly proposed PESRS model can significantly outperform the existing methods. 
Particularly, PESRS yields 4.8\% and 7.1\% percentage point improvement in terms of $MAP$ and $R_{10}@1$ compared with our early work SRS.
In addition to the comprehensive evaluation, we also evaluate our proposed user preference memory by a fine-grained analysis.
The analysis reveals how the model leverages the user's recent sticker selection history and provides us insights on why they can achieve big improvement over state-of-the-art methods.

This work is a substantial extension of our previous work reported at WWW 2020. 
The extension in this article includes the user preference modeling framework for the existing methods, a proposal of a new framework for sticker selection in the multi-turn dialog.
Specifically, the contributions of this work include the following:

\begin{itemize}
    \item We propose a position-aware sticker modeling module which can model the user's sticker selection history.
    \item We propose a key-value memory network to store the user's recently used stickers and its corresponding dialog context.
    \item Finally, we use the current dialog context to query the key-value memory and obtain a user preference representation, and then fuse the user preference representation into final sticker prediction dynamically.  
    \item Experiments conducted on a large-scale real-world dataset show that our model outperforms all baselines, including state-of-the-art models. Experiments also verify the effectiveness of each module in PESRS as well as its interpretability.
\end{itemize}

The rest of the paper is organized as follows:
We summarize related work in \S\ref{sec:related}. 
\S\ref{sec:dataset} introduces the data collection method and some statistics of our proposed multi-turn dialog sticker selection dataset.
We then formulate our research problem in \S \ref{sec:formulation} and elaborate our approach in \S\ref{sec:model}. 
\S\ref{sec:exp-setup} gives the details of our experimental setup and \S\ref{sec:exp-result} presents the experimental results. 
Finally, \S\ref{sec:conclusion} concludes the paper.

\section{Related Work}\label{sec:related}

We outline related work on sticker recommendation, user modeling, visual question answering, visual dialog, and multi-turn response selection.

\subsection{Sticker and Emoji Recommendation}

Most of the previous works emphasize the use of emojis instead of stickers.
For example, \cite{barbieri2017emojis,barbieri2018multimodal} use a multimodal approach to recommend emojis based on the text and images in an Instagram post.
\cite{Guibon2018EmojiRI} propose a MultiLabel-RandomForest algorithm to predict emojis based on the private instant messages.
\cite{Zhao2020CAPERCP} conduct emoji prediction on social media text (\eg Sina Weibo and Twitter), and they tackle this task as ranking among all emojis.
The total number of unique emojis in their dataset is 50, which is much smaller than the number of stickers.
What is more, emojis are limited in variety, while there exists an abundance of different stickers.
\cite{Zhou2018MojiTalk} incorporates the emoji information into the dialog generation task, and they use the emoji classification as an auxiliary task to facilitate the dialog generation to produce utterance with proper emotion.
The most similar work to ours is \cite{laddha2019understanding}, where they generate recommended stickers by first predicting the next message the user is likely to send in the chat, and then substituting it with an appropriate sticker.

However, more often than not the implication of the stickers cannot be fully conveyed by text and, in this paper, we focus on directly generating sticker recommendations from dialog history.

\subsection{User Modeling}

User modeling~\cite{Ren2019Lifelong,zolna2017User,Huang2019Explainable,Zhu2017What,Yang2017Multi} is a hot research topic especially in recommendation task, which models the preference of user based on the user history interaction data.
Specifically, in the e-commerce recommendation task~\cite{Lei2019TiSSA,Ren2019RepeatNet,Huang2019TaxonomyAware}, the user modeling systems use the purchase history or click records to model the user's intrinsic interest and temporal interest~\cite{Yu2019Adaptive,Pi2019Practice}.
Most of the research typically utilize user-item binary relations, and assume a flat preference distribution over items for each user.
They neglect the hierarchical discrimination between user intentions and user preferences.
\citet{Zhu2020Sequential} propose a novel key-array memory network with user-intention-item triadic relations, which takes both user intentions and preferences into account for the next-item recommendation.
As for the user modeling in the news recommendation task, there are much side information can be used to obtain better user preference representation.
\citet{Wu2019Neural} propose a neural news recommendation approach which can exploit heterogeneous user behaviors, including the search queries and the browsed webpages of the user.

However, to model the user preference of sticker selection, we should not only model the sticker selection history, and the dialog context of each selected sticker should also be considered when modeling the user preference.

\subsection{Memory Networks}

The memory network proposed by \citet{Sukhbaatar2015EndToEndMN} generally consists of two components.
The first one is a memory matrix to save information (\ie memory slots) and the second one is a neural network to read/write the memory slots.
The memory network has shown better performance than traditional long-short term memory network in several tasks, such as question answering~\cite{Sukhbaatar2015EndToEndMN,Pavez2018Working,Ma2018Visual,Gao2018MotionAppearance}, machine translation~\cite{Maruf2018Document}, text summarization~\cite{Kim2019Abstractive,Chen2019Learning,Gao2020From}, dialog system~\cite{Chu2018Learning,Wu2019Globaltolocal} and recommendation~\cite{Ebesu2018Collaborative,Wang2018Neural,Zhou2019TopicEnhanced}.
The reason is that the memory network can store the information in a long time range and has more memory storage units than LSTM which has the single hidden state.
Follow memory network, there are many variations of memory network have been proposed, \ie key-value memory network~\cite{Miller2016KeyValueMN} and dynamic memory network~\cite{Xiong2016DynamicMN,Kumar2016AskMA}.
Our method is mainly based on the key-value memory network~\cite{Miller2016KeyValueMN}, which employs the user history dialog contexts as the memory keys and the corresponding selected stickers the memory values.
However, there are two main differences between our PESRS model and the previous key-value memory network.
First, the user history data is in chronological order, we should consider the time information when storing them into the memory.
To recommend more accurate stickers, the model should not only consider the user preference information stored in the memory, but also incorporates the matching result between current dialog context and candidate stickers.
The second difference lies in that we propose a dynamic fusion layer that considers both the memory read output and the matching result of the current context.
Compared with these methods, we not only implement a key-value memory network, but also provide a sticker selection framework that could incorporate the user's preference.

\subsection{Visual Question Answering}
Sticker recommendation involves the representation of and interaction between images and text, which is related to the Visual Question Answering (VQA) task~\cite{Goyal2018Think,Gao2019Multi,Chao2018Cross,Wang2017Explicit,Noh2019Transfer,Li2018Visual,Su2018Learning}.
Specifically, VQA takes an image and a corresponding natural language question as input and outputs the answer.
It is a classification problem in which candidate answers are restricted to the most common answers appearing in the dataset and requires deep analysis and understanding of images and questions such as image recognition and object localization~\cite{malinowski2015ask,xiong2016dynamic,wu2016ask,goyal2017making}.
Current models can be classified into three main categories: early fusion models, later fusion models, and external knowledge-based models.
One state-of-the-art VQA model is \cite{li2019beyond}, which proposes an architecture, positional self-attention with co-attention, that does not require a recurrent neural network (RNN) for video question answering.
\cite{guo2019image} proposes an image-question-answer synergistic network, where candidate answers are coarsely scored according to their relevance to the image and question pair in the first stage. 
Then, answers with a high probability of being correct are re-ranked by synergizing with images and questions.

The difference between sticker selection and VQA task is that the sticker selection task focus more on multi-turn multimodal interaction between stickers and utterances.

\subsection{Visual Dialog}
Visual dialog extends the single turn dialog task~\cite{Tao2018Get,Guo2019Dual,Murahari2019Improving} in VQA to a multi-turn one, where later questions may be related to former question-answer pairs. %
To solve this task, \cite{lu2017best} transfers knowledge from a pre-trained discriminative network to a generative
network with an RNN encoder, using a perceptual loss.
\cite{wu2018you} combines reinforcement learning and generative adversarial networks (GANs) to generate more human-like responses to questions, where the GAN helps overcome the relative paucity of training data, and the
tendency of the typical maximum-likelihood-estimation-based approach to generate overly terse answers.
\cite{jain2018two} demonstrates a simple symmetric discriminative baseline that can be applied to both predicting an answer as well as predicting a question in the visual dialog.

Unlike visual dialog tasks, in a sticker recommendation system, the candidates are stickers rather than text.

\subsection{Multi-turn Response Selection}
Multi-turn response selection~\cite{Tao2019One,Feng2019Learning,Yan2018Coupled,Yan2017Joint,Yan2016LearningTR,Li2019Insufficient,Chan2019Modeling} takes a message and utterances in its previous turns as input and selects a response that is natural and relevant to the whole context.
In our task, we also need to take previous multi-turn dialog into consideration.
Previous works include \cite{zhou2016multi}, which uses an RNN to represent context and response, and measure their relevance.
More recently, \cite{Wu2017SequentialMN} matches a response with each utterance in the context on multiple levels of granularity, and the vectors are then combined through an RNN.
The final matching score is calculated by the hidden states of the RNN.
\cite{zhou2018multi} extends this work by considering the matching with dependency information.
More recently, \cite{tao2019multi} proposes a multi-representation fusion network where the representations can be fused into matching at an early stage, an intermediate stage, or at the last stage.

Traditional multi-turn response selection deals with pure natural language processing, while in our task, we also need to obtain a deep understanding of images. %
\section{Dataset}
\label{sec:dataset}

In this section, we introduce our multi-turn dialog dataset with sticker as response in detail.

\subsection{Data Collection}

We collect the large-scale multi-turn dialog dataset with stickers from one of the most popular messaging apps, Telegram\footnote{https://telegram.org/}.
In this app, a large amount of sticker sets are published, and everyone can use the sticker when chatting with a friend or in a chat group.
Specifically, we select 20 public chat groups consisting of active members, which are all open groups that everyone can join it without any authorities.
The chat history of these groups is collected along with the complete sticker sets.
These sticker sets include stickers with similar style.
All stickers are resized to a uniform size of $128 \times 128$ pixels.
We use 20 utterances before the sticker response as the dialog context, and then we filter out irrelevant utterance sentences, such as URL links and attached files.
Due to privacy concern, we also filter out user information and anonymize user IDs.
To construct negative samples, 9 stickers other than the ground truth sticker are randomly sampled from the sticker set.
After pre-processing, there are 320,168 context-sticker pairs in the training dataset, 10,000 pairs in the validation, and 10,000 pairs in test datasets respectively.
We make sure that there is no overlap between these three datasets, there is no the same dialog context in any two datasets.
Two examples are shown in Figure~\ref{fig:dataset-case}.
We publish this dataset to communities to facilitate further research on dialog response selection task.

\subsection{Statistics and Analysis}

\begin{table}[t]
\centering
\caption{Statistics of Response Selection Dataset.}
\label{tab:stat-dataset}
\begin{tabular}{llll}
\toprule
 & Train & Valid & Test \\
\midrule
\# context-stickers pairs & 320,168 & 10,000 & 10,000 \\
Avg. words of context utterance & 7.54 & 7.50 & 7.42 \\
Avg. users participate & 5.81 & 5.81 & 5.79 \\
\bottomrule
\end{tabular}
\end{table}
In total, there are 3,516 sets of sticker which contain 174,695 stickers.
The average number of stickers in a sticker set is 49.64.
Each context includes 15.5 utterances on average.
The average number of users who participate in the dialog context over each dataset is shown in the third row of Table~\ref{tab:stat-dataset}.

\begin{figure*} 
    \centering 
    \subfigure[The distribution of history length in training dataset.]{ 
        \label{fig:distribution-history-len}
        \includegraphics[scale=0.38]{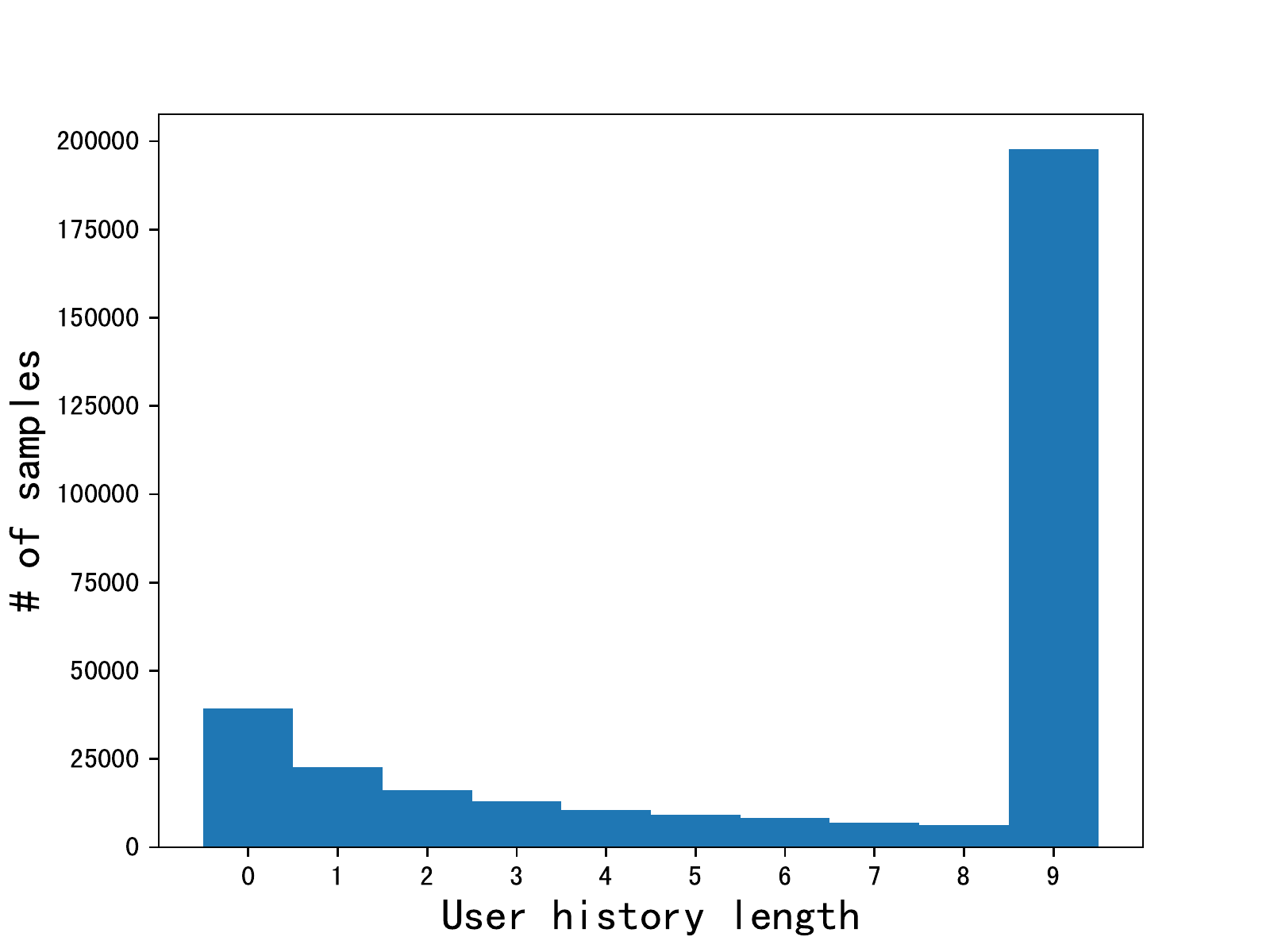}
    } 
    \subfigure[Similarity distribution among all stickers in test dataset.]{ 
        \label{fig:similarity-distribution}
        \includegraphics[scale=0.40]{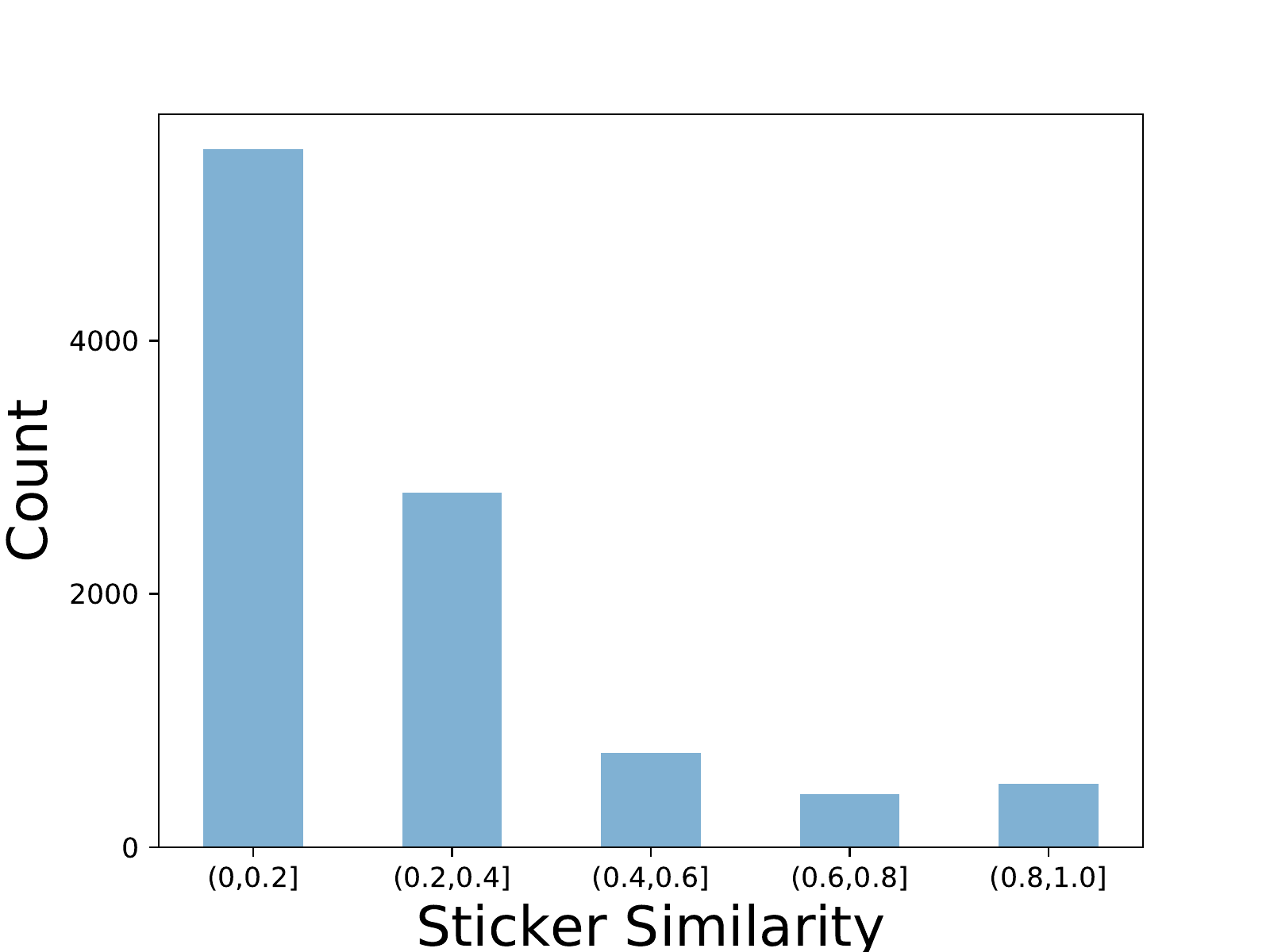}
    } 
    \caption{Statistics of dataset.}
\end{figure*}

Since not all the users have history dialog data, we calculate the percentage of how many data samples in our dataset have history data.
There are 290939 data samples in our training dataset which have at lease one history sticker selection history, and the percentage is 88.12\%.
We set the maximum of retrieved history data pair (consisting of dialog context and selected sticker) for one data sample to 10, and the average of history length in our training dataset is 6.82.
We also plot the distribution of history length in Figure~\ref{fig:distribution-history-len}.

\subsection{Sticker Similarity}

\begin{figure*}[t]
    \centering
    \includegraphics[scale=0.50]{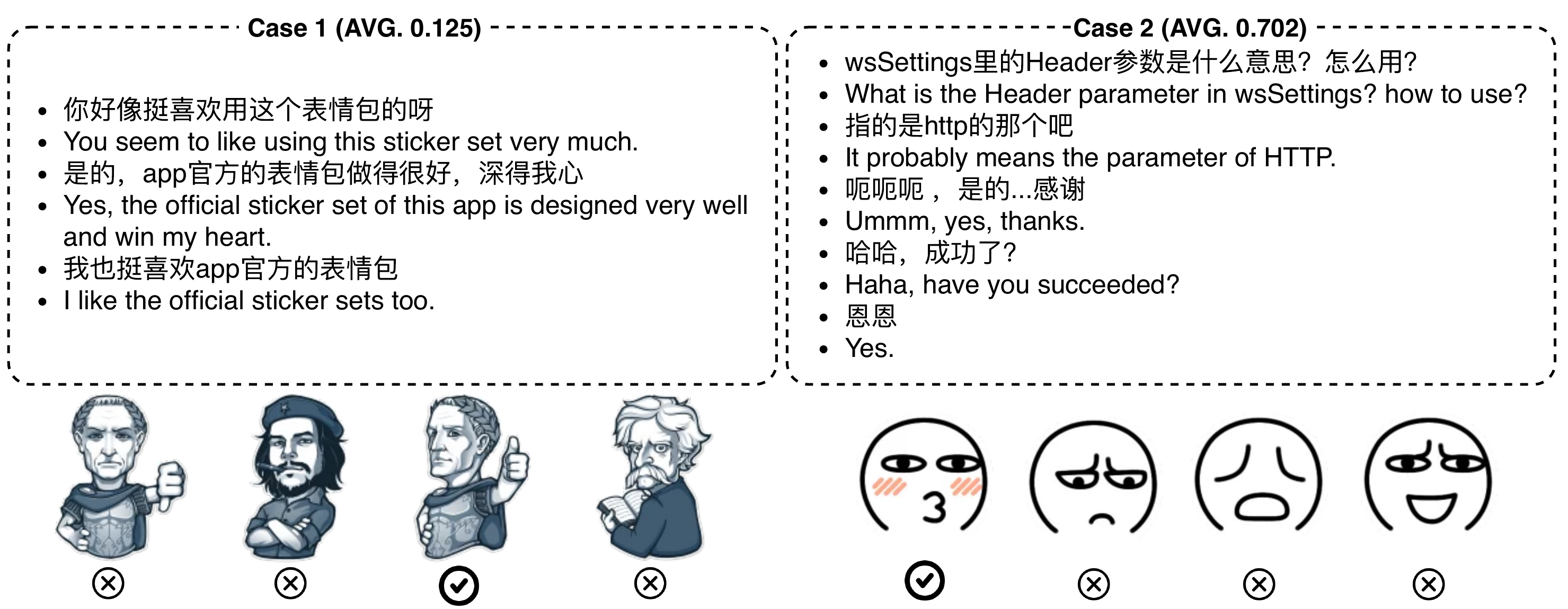}
    \caption{
        Example cases in the dataset with different similarity scores.
    }
    \label{fig:dataset-case}
\end{figure*}

Stickers in the same set always share a same style or contain the same cartoon characters.
Intuitively, the more similar the candidate stickers are, the more difficult it is to choose the correct sticker from candidates.
In other words, the similarity between candidate stickers determines the difficulty of the sticker selection task.
To investigate the difficulty of this task, we calculate the average similarity of all the stickers in a specific sticker set by the Structural Similarity Index (SSIM) metric~\cite{wang2004image,avanaki2008exact}.
We first calculate the similarity between the ground truth sticker and each negative sample, then average the similarity scores.
The similarity distribution among test data is shown in Figure~\ref{fig:similarity-distribution}, where the average similarity is 0.258.
The examples in Figure~\ref{fig:dataset-case} are also used to illustrate the similarity of stickers more intuitively, where the left one has a relatively low similarity score, and the right one has a high similarity score.
\section{Problem formulation}
\label{sec:formulation}

Before presenting our approach for sticker response selection in multi-turn dialog, we first introduce our notations and key concepts. 
Table~\ref{tbl:notations} lists the main notations we use.

\begin{table}[!t]
 \caption{Glossary.}
 \label{tbl:notations}
 \centering
 \begin{tabular}{ll}
  \toprule
  Symbol & Description \\
  \midrule   
  $s$ & multi-turn dialog context \\
  $u_{i}$ & $i$-th utterance in $s$ \\
  $T_u$ & number of utterances in dialog context \\
  $x^i_j$ & $j$-th word in $i$-th utterance $u_{i}$ \\
  $T_x^i$ & number of words in the $i$-th utterance \\
  $C$ & candidate sticker set \\
  $c_{i}$ & $i$-th candidate sticker in $c$ \\
  $T_c$ & number of stickers in candidate sticker set $c$ \\
  $y_i$ & the selection label of $i$-th sticker $c_i$ \\
  $\hat{s}^k$ & $k$-th multi-turn dialog context in history \\
  $\hat{u}^k_{i}$ & $i$-th utterance in $k$-th history context $\hat{s}^k$ \\
  $\hat{x}^{k,i}_j$ & $j$-th word in $i$-th utterance $\hat{u}^k_{i}$ of $k$-th history context \\
  $\hat{c}_{k}$ & user selected sticker of $k$-th history \\
  $T_h$ & number of history dialog context and selected sticker \\
  \bottomrule
 \end{tabular}
\end{table}

Similar to the multi-turn dialog response selection~\cite{Wu2017SequentialMN,zhou2018multi}, we assume that there is a multi-turn dialog context $s=\{u_{1},\dots,u_{T_u}\}$ and a candidate sticker set $C=\{c_{1},...c_{T_c}\}$, where $u_{i}$ represents the $i$-th utterance in the multi-turn dialog.
In the $i$-th utterance $u_i=\{x^i_1,\dots,x^{i}_{T_x^i}\}$, $x^i_j$ represents the $j$-th word in $u_i$, and $T_x^i$ represents the total number of words in $u_i$ utterance.
In dialog context $s$, $c_{i}$ represents a sticker image with a binary label $y_i$, indicating whether $c_i$ is an appropriate response for $s$.
$T_u$ is the number of utterance in the dialog context and $T_c$ is the number of candidate stickers.
For each candidate set, there is only one ground truth sticker, and the remaining ones are negative samples.

To model the user preference, we use $T_h$ history dialog contexts with user selected sticker $\{(\hat{s}^1, \hat{c}_1), \dots, (\hat{s}^{T_h}, \hat{c}_{T_h})\}$, where $\hat{s}^i$ denotes the $i$-th history dialog context and $\hat{c}_i$ denotes the user selected sticker at $i$-th history dialog context.
In the remaining of the paper, we use the word \textbf{current} to denotes the dialog context $s$ and sticker $c_i$ which the model needs to predict the sticker selection, and we use the word \textbf{history} to denote the dialog context and sticker which user has generated before.
In the $k$-th history, there is a dialog context $\hat{s}^k=\{\hat{u}^k_{i},\dots,\hat{u}^k_{T_u}\}$ which contains up to $T_u$ utterances as the same as current dialog context $s$, and a user selected sticker $\hat{c}_{k}$.
For each dialog history, we pad the dialog context which number of utterances is less than $T_u$ to $T_u$.
Our goal is to learn a ranking model that can produce the correct ranking for each candidate sticker $c_i$; that is, can select the correct sticker among all the other candidates.
For the rest of the paper, we take the $i$-th candidate sticker $c_{i}$ as an example to illustrate the details of our model and omit the candidate index $i$ for brevity.
In some of the sticker selection scenarios, the stickers in the preceding dialog context may affect the current decision of sticker selection. 
But in most cases, the sticker selection is influenced by a few utterances before. 
Thus, in this paper, we focus on modeling the text utterances in dialog context. 
And we will consider the information provided by the stickers in the preceding context in our future work. 

\section{PESRS model}
\label{sec:model}

\begin{figure*}
    \centering
    \includegraphics[scale=0.7]{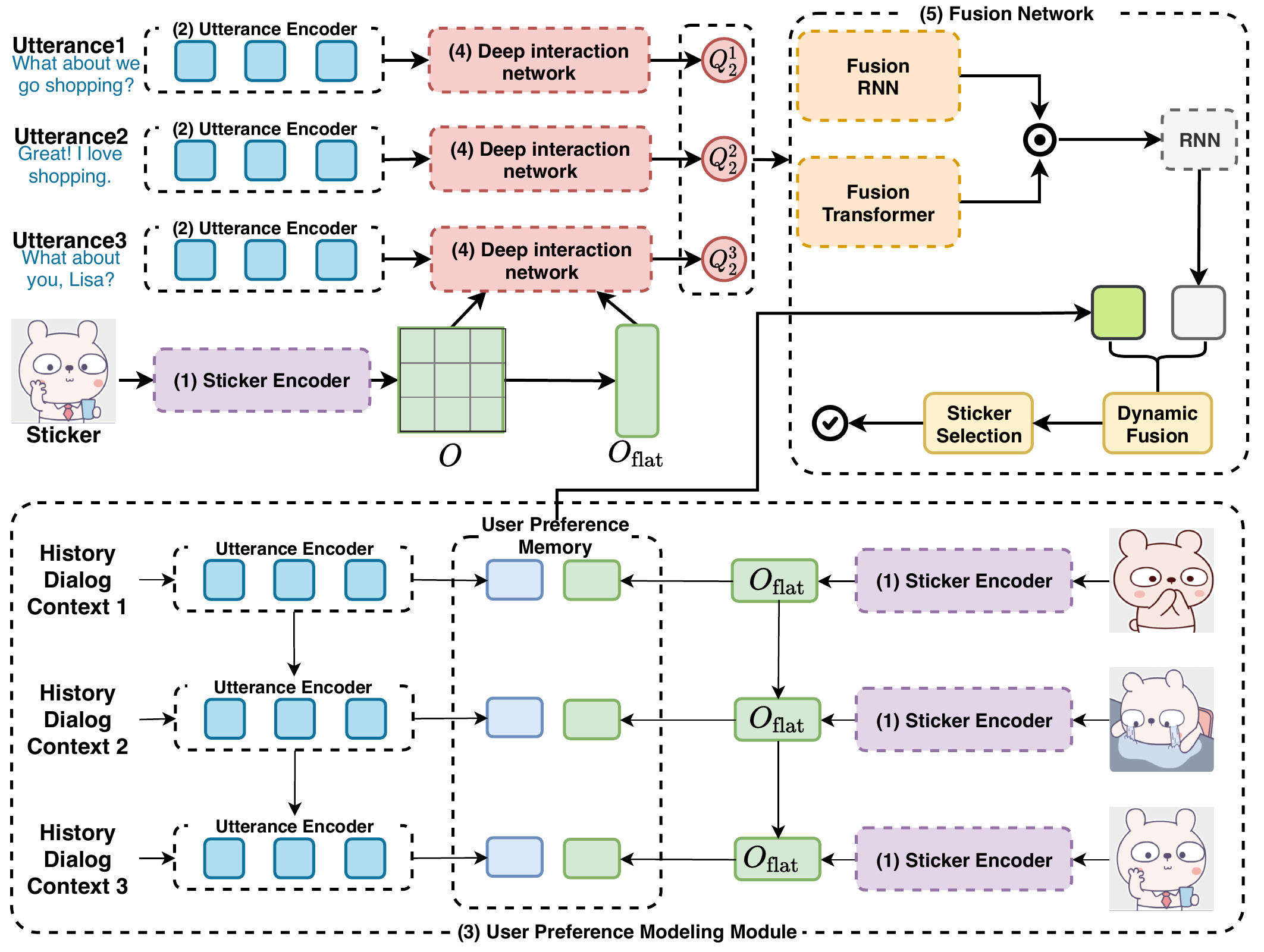}
    \caption{
     Overview of PESRS. 
     We divide our model into five ingredients: 
     (1) \textit{Sticker encoder} learns sticker representation by a convolutional neural network; 
     (2) \textit{Utterance encoder} learns representation of each utterance by self-attention based Transformer; 
     (3) \textit{User preference modeling module} obtains the position-aware history representations and store them into a key-value memory network;
     (4) \textit{Deep interaction network} conducts deep matching interaction between sticker representation and utterance representation in different levels of granularity;
     (5) \textit{Fusion network} combines the long-term and short-term dependency feature between interaction results produced by (4) and the user preference representation produced by (3) into final sticker prediction layer.
    }
    \label{fig:model}
\end{figure*}

\subsection{Overview}

In this section, we propose our \emph{preference enhanced sticker response selector}, abbreviated as PESRS. 
An overview of PESRS is shown in Figure~\ref{fig:model}, which can be split into five main parts:

\begin{itemize}
    \item \textit{Sticker encoder} is a convolutional neural network (CNN) based image encoding module that learns a sticker representation. %
    \item \textit{Utterance encoder} is a self-attention mechanism-based module encoding each utterance $u_{i}$ in the multi-turn dialog context $s$.
    \item \textit{User preference modeling module} is a key-value memory network that stores the representation of history dialog context and corresponding selected sticker.
    \item \textit{Deep interaction network} module conducts deep matching between each sticker representation and each utterance, and outputs each interaction result.
    \item \textit{Fusion network} learns the short-term dependency by the fusion RNN and the long-term dependency by the fusion Transformer, and finally outputs the matching score by combining the current interaction results with user preference representation using a gated fusion layer.
\end{itemize}

\subsection{Sticker Encoder}
\label{subsec:sticker_encoder}

Much research has been conducted to alleviate gradient vanishing~\cite{he2016deep} and reduce computational costs~\cite{he2015delving} in image modeling tasks.
We utilize one of these models, \ie the Inception-v3~\cite{szegedy2016rethinking} model rather than plain CNN to encode sticker image:
\begin{align}
   O, O_{\text{flat}} &= \text{Inception-v3}(c) , \label{eq:inceptionv3}
\end{align}
where $c$ is the sticker image.
The sticker representation is $O \in \mathbb{R}^{p \times p \times d}$ which conserves the two-dimensional information of the sticker, and will be used when associating stickers and utterances in \S\ref{deep_int}.
We use the original image representation output of Inception-v3 $O_{\text{flat}} \in \mathbb{R}^{d}$ as another sticker representation.
Most imaging grounded tasks~\cite{Jing2018OnTA,Wu2018ChainOR,Wu2018ObjectDifferenceAA} employ the pre-trained image encoding model to produce the image representation.
However, existing pre-trained CNN networks including Inception-v3 are mostly built on real-world photos.
Thus, directly applying the pre-trained networks on stickers cannot speed up the training process.
In this dataset, sticker author give each sticker $c$ an emoji tag which denotes the general emotion of the sticker.
Hereby, we propose an auxiliary sticker classification task to help the model converge quickly, which uses $O_{\text{flat}}$ to predict which emoji is attached to the corresponding sticker.
More specifically, we feed $O_{\text{flat}}$ into a linear classification layer and then use the cross-entropy loss $\mathcal{L}_s$ as the loss function of this classification task.

\subsection{Utterance Encoder}

To model the semantic meaning of the dialog context, we learn the representation of each utterance $u_i$.
First, we use an embedding matrix $e$ to map a one-hot representation of each word in each utterance $u_i$ to a high-dimensional vector space.
We also add the positional embedding to the original word embedding, and we use the $e(x^i_j)$ to denote the embedding representation of word $x^i_j$.
The positional embedding is the same as Transformer~\cite{vaswani2017attention}.
From these embedding representations, we use the attentive module with positional encoding from Transformer~\cite{vaswani2017attention} to model the interactions between the words in an utterance.
Attention mechanisms have become an integral part of compelling sequence modeling in various tasks~\cite{bahdanau2014neural,fan2018hierarchical,Gao2019Abstractive,li2019beyond}.
In our sticker selection task, we also need to let words fully interact with each other words to model the dependencies of words in the input sentence.
The self attentive module in the Transformer requires three inputs: the query $Q$, the key $K$ and the value $V$.
To obtain these three inputs, we use three linear layers with different parameters to project the embedding of dialog context $e(x^i_j)$ into three spaces:
\begin{align}
    Q^i_j &= FC(e(x^i_j)), \label{equ:transformer-q-linear} \\
    K^i_j &= FC(e(x^i_j)) , \\
    V^i_j &= FC(e(x^i_j)).
\end{align}
The self attentive module then takes each $Q^i_j$ to attend to $K^i_\cdot$, and uses these attention distribution $\alpha^{i}_{j, \cdot} \in \mathbb{R}^{T_x^i}$ as weights to gain the weighted sum of $V^i_j$, as shown in Equation~\ref{equ:transformer-sum}.
\begin{align}
    \alpha^i_{j,k} &= \frac{\exp\left( Q^i_j \cdot K^i_k \right)}{\sum_{n=1}^{T_x^i} \exp\left(Q^i_j \cdot K^i_n\right)}, \label{equ:attention}\\
    \beta^i_{j} &= \sum_{k=1}^{T_x^i} \alpha^i_{j,k} \cdot V^i_{k}, \label{equ:transformer-sum}
\end{align}
Next, we add the original word embedding $e(x^i_j)$ on $\beta^i_{j}$ as the residential connection layer, shown in Equation~\ref{equ:drop-add}:
\begin{align}
    \hat{h}^i_j = \text{Dropout} \left( e(x^i_j) + \beta^i_j \right), \label{equ:drop-add}
\end{align}
where $\alpha^i_{j,k}$ denotes the attention weight between $j$-th word to $k$-th word in $i$-th utterance.
To prevent vanishing or exploding of gradients, a layer normalization operation~\cite{lei2016layer} is also applied on the output of the feed-forward layers with ReLU activation as shown in Equation~\ref{equ:ffn}: 
\begin{align}
    h^i_j = \text{norm} \left( max(0, \hat{h}^i_j \cdot W_1 + b_1) \cdot W_2 + b_2 + \hat{h}^i_j \right), \label{equ:ffn}
\end{align}
where $W_1, W_2, b_1, b_2$ are all trainable parameters of the feed-forward layer.
$h^{i}_j$ denotes the hidden state of $j$-th word for the $i$-th utterance in the Transformer.
We also employ the multi-head attention is our model which conducts these operation multiple times and then concatenate the outputs as the final representation.
For brevity, we omit these multi-head operation in our equations.

\subsection{Deep Interaction Network}
\label{deep_int}

\begin{figure*}
    \centering
    \includegraphics[scale=0.75]{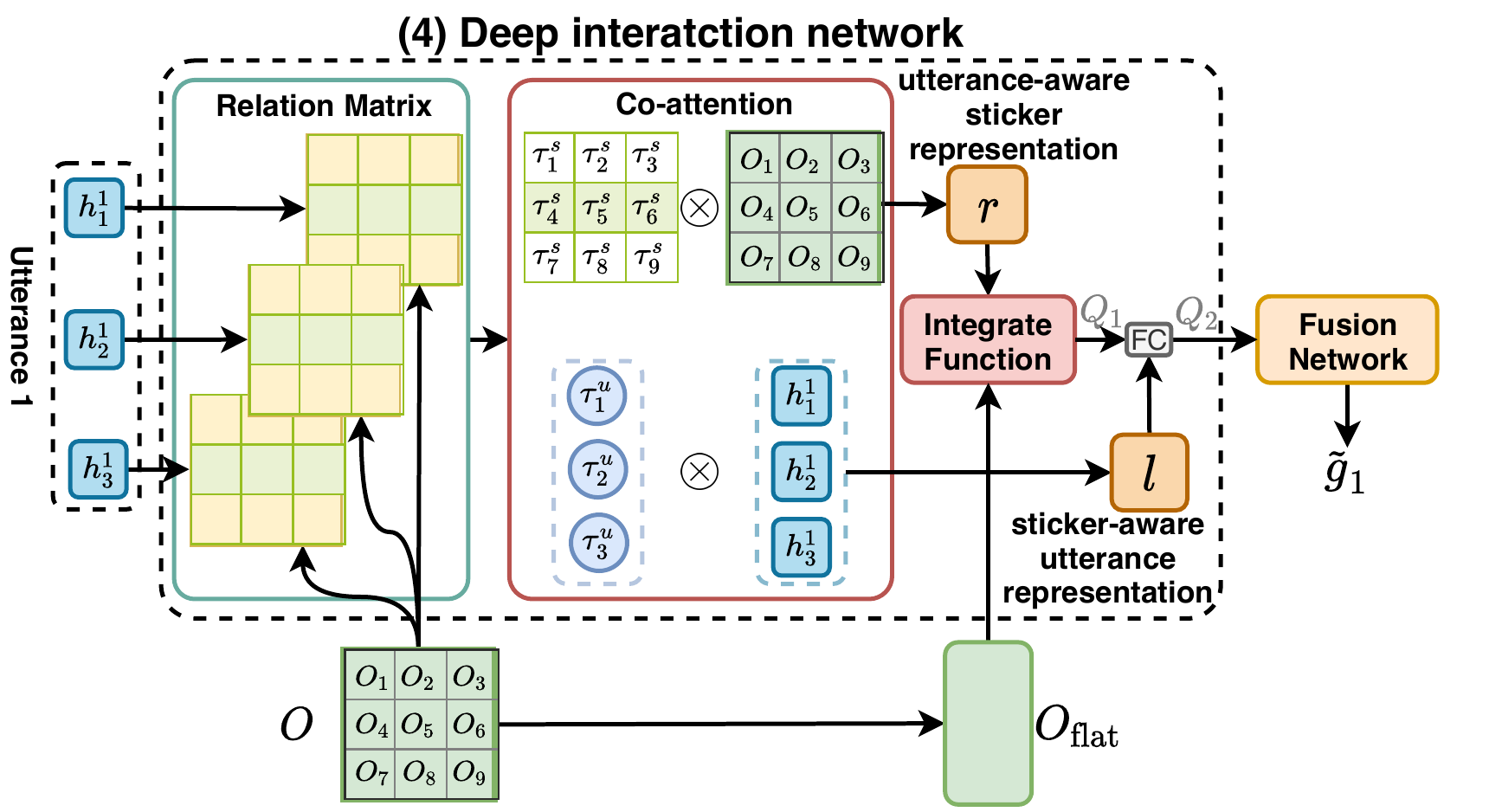}
    \caption{Framework of deep interaction network.}
    \label{fig:interaction}
\end{figure*}

Now that we obtain the representation of the sticker and each utterance, we can conduct a deep matching between these components to model the bi-directional relationship between the words in dialog context and the sticker patches.
On one hand, there are some emotional words in dialog context history that match the expression of the stickers such as ``happy'' or ``sad''.
On the other hand, specific parts of the sticker can also match these corresponding words such as dancing limbs or streaming eyes.
Hence, we employ a bi-directional attention mechanism between a sticker and each utterance, that is, from utterance to sticker and from sticker to utterance, to analyze the cross-dependency between the two components.
The interaction is illustrated in Figure~\ref{fig:interaction}.

We take the $i$-th utterance as an example and omit the index $i$ for brevity.
The two directed attentions are derived from a shared relation matrix, $M \in  \mathbb{R}^{(p^2) \times  T_{u}}$, calculated by sticker representation $O \in \mathbb{R}^{p \times p \times d}$ and utterance representation $h \in \mathbb{R}^{T_{u} \times d}$.
The score $M_{kj} \in \mathbb{R}$ in the relation matrix $M$ indicates the relation between the $k$-th sticker representation unit $O_k$, $k \in [1,p^2]$ and the $j$-th word $h_j$, $j \in [1, T_{u}]$ and is computed as:
\begin{align}
    M_{kj} &= \sigma(O_k, h_j) , \\
    \sigma(x, y) &= w^\intercal [x \oplus y \oplus (x \otimes y)] , \label{eq:alpha}
\end{align}
where $\sigma$ is a trainable scalar function that encodes the relation between two input vectors. 
$\oplus$ denotes a concatenation operation and $\otimes$ is the element-wise multiplication.

Next, a two-way max pooling operation is conducted on $M$, \ie let $\tau_j^u = \max(M_{:j}) \in \mathbb{R}$ represent the attention weight on the $j$-th utterance word by the sticker representation, corresponding to the ``utterance-wise attention''.
This attention learns to assign high weights to the important words that are closely related to sticker.
We then obtain the weighted sum of hidden states as ``\textbf{sticker-aware utterance representation}'' $l$:
\begin{equation}\label{equ:sa-utterance}
l = \sum^{T_{u}}_j {\tau_j^u h_j} .
\end{equation}

Similarly, sticker-wise attention learns which part of a sticker is most relevant to the utterance.
Let $\tau_k^s = \max(M_{k:}) \in \mathbb{R}$ represent the attention weight on the $k$-th unit of the sticker representation. We use this to obtain the weighted sum of $O_{k}$, \ie the ``\textbf{utterance-aware sticker representation}'' $r$:
\begin{equation}\label{equ:ua-sticker}
r = \sum^{p^2}_k {\tau_k^s O_{k}} .
\end{equation}

After obtaining the two outputs from the co-attention module, we combine the sticker and utterance representations and finally get the ranking result.
We first integrate the utterance-aware sticker representation $r$ with the original sticker representation $O_{\text{flat}}$ using an \textbf{integrate function}, named $IF$:
\begin{align}
    Q_1 &= \text{IF} \left( O_{\text{flat}}, r \right) , \\
    \text{IF}(x, y) &= \text{FC} \left(x \oplus y \oplus \left( x \otimes y \right) \oplus (x+y) \right) , \label{triliner}
\end{align}
where $\text{FC}$ denotes the fully-connected (FC) layer and we use the ReLU~\cite{Nair2010RectifiedLU} as the activation function, $\oplus$ represents the vector concatenation along the final dimension of the vector and $\otimes$ denotes the elementwise product operation.
We add the sticker-aware utterance representation $l$ into $Q_1$ together and then apply a fully-connected layer with ReLU activation:
\begin{align}\label{equ:q2}
    Q_2 = \text{FC} (Q_1 \oplus l) .
\end{align}

\subsection{User Preference Modeling Module}
\label{subsec:preference}

\begin{figure*}
    \centering
    \includegraphics[scale=0.6]{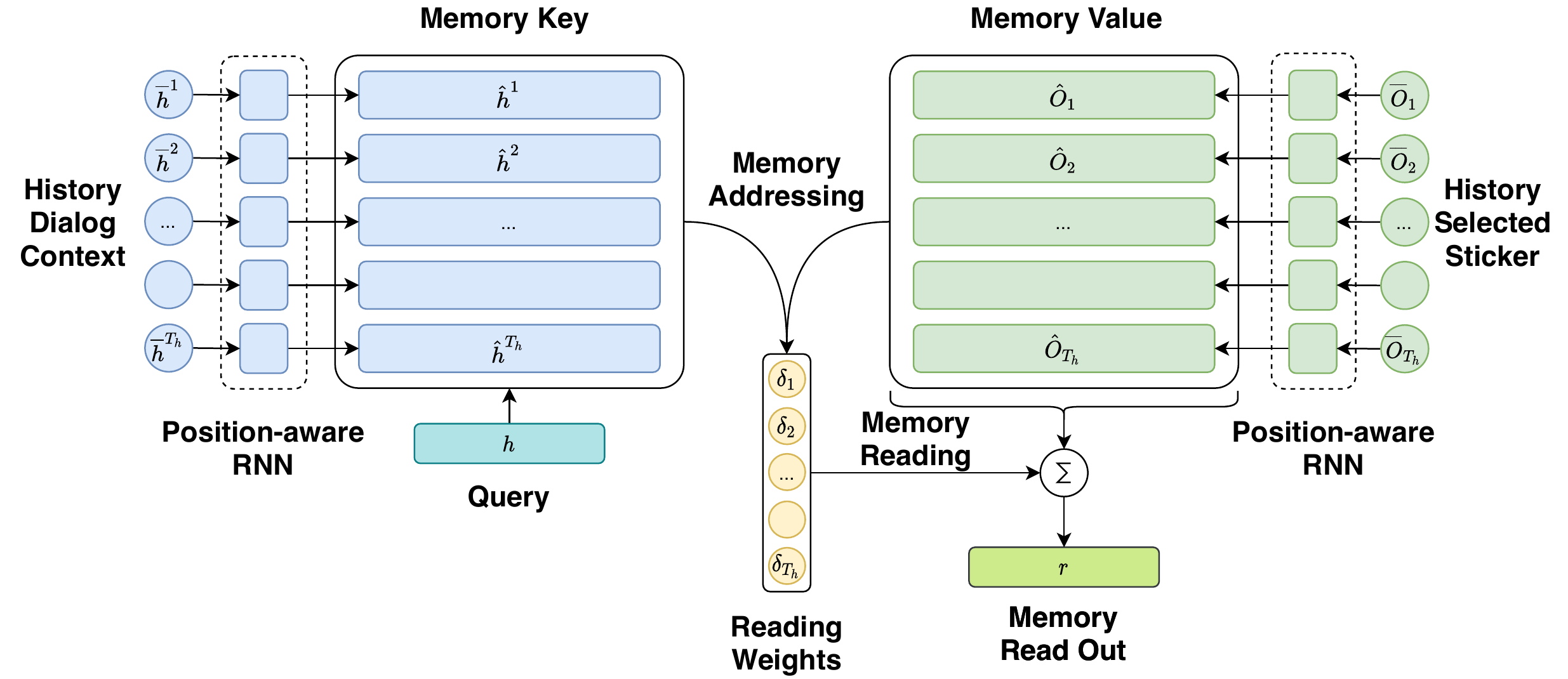}
    \caption{Framework of user preference modeling module which consists a position-aware RNN and a key-value memory network. We use the history dialog contexts as the keys and the history selected stickers as the values. Finally, we use the representation of the current dialog context as the query the user preference memory.}
    \label{fig:preference-memory}
\end{figure*}

Users have their preference when selecting the sticker as the response of the multi-turn dialog context.
Hence, to recommend the sticker more accurately, our model should consider the user's preference when giving the final sticker recommendation.
Intuitively, the sticker that selected by the user recently contains the user's preference, and these history data can help our model to build the preference representation.
As for constructing the preference modeling module, our motivation is to find the semantically similar dialog contexts in the history data, and then use the corresponding selected stickers of these dialog contexts to facilitate the final sticker prediction of the current dialog context.
Hence, we propose the user preference memory and the architecture of this module as shown in Figure~\ref{fig:preference-memory}.
The proposed user preference memory unit inherits from memory networks~\cite{Gao2019Product,Tao2019Log2Intent,Wang2018Neural,Chen2018Sequential}, and generally has two steps: (1) memory addressing and (2) memory reading. 
The user preference memory consists of a set of history multi-turn dialog contexts and selected stickers. 
Though one action is corresponding to a dialog context, it should attend to the different history contexts (\ie memory slots) upon the current dialog context.
Thus, we address and read the memory unit as follows.

\subsubsection{History Encoding} \label{subsubsec:history-encoding}

To store the history dialog contexts and selected stickers, we encode them into vector spaces using the same method as used when encoding current dialog context and candidate stickers.
Concretely, first, attentive module is employed to encode all the dialog contexts $\{\hat{s}^1, \dots, \hat{s}^{T_h}\}$:
\begin{align}
    \overline{h}^k_{i} = \text{mean-pooling} (\text{Transformer} (\hat{u}^k_{i})), \label{equ:transformer-history}
\end{align}
where $\overline{h}^k_{i}$ is the vector representation of the $i$-th utterance in the $k$-th history, the $\text{Transformer}$ is the same operation as shown in Equation~\ref{equ:transformer-q-linear}-Equation~\ref{equ:ffn}.
Different with the Transformer used in Equation~\ref{equ:transformer-q-linear}, the query, key and value in Equation~\ref{equ:transformer-history} are all $\hat{u}^k_{i}$, where we conduct self-attention over all history dialog contexts.
Then, we use a max-pooling layer to obtain the vector representation $\overline{h}^k$ of the $k$-th dialog context in history:
\begin{align}
    \overline{h}^k = \max (\{\overline{h}^k_{1}, \dots, \overline{h}^k_{T_h}\}).
\end{align}

Next, we use the same image encoder $\text{Inception-v3}$ in \S~\ref{subsec:sticker_encoder} to encode all the stickers $\{\hat{c}_{1}, \dots, \hat{c}_{T_h}\}$ of each history dialog context into vector representations $\{\overline{O}_{1}, \dots, \overline{O}_{T_h}\}$:
\begin{align}
    \overline{O}_k &= \text{Inception-v3}(\hat{c}_{k}) , \label{eq:inceptionv3}
\end{align}
where sticker representation $\overline{O}_{k} \in \mathbb{R}^{d}$ is a one-dimensional vector, we drop the output $O$ and use the output $O_{flat}$ as the $\overline{O}_{k}$.

Intuitively, it is much easier for the user to recall their recently used stickers than the stickers they used a long time ago.
Thus, we propose a recurrent neural network (RNN) based position-aware user history modeling layer which incorporates the position feature into the history data representation, \eg history dialog context representation $\overline{h}^k$ and history selected sticker representation $\overline{O}_{k}$.
We first concatenate the position of history dialog context as an additional feature to the vector representation of dialog representation $\overline{h}^k$ and sticker representation $\overline{O}_{k}$.
Then, we employ an RNN to encode these representations in chronological order:
\begin{align}
    \hat{h}^k &= \text{RNN} (t_k \oplus \overline{h}^k, \hat{h}^{k-1}) ,\\
    \hat{O}_{k} &= \text{RNN} (t_k \oplus \overline{O}_{k}, \hat{O}_{k-1}) .
\end{align}
Finally, we obtain the position-aware history data representations $\{\hat{h}^1, \dots, \hat{h}^{T_h}\}$ and $\{\hat{O}_{1}, \dots, \hat{O}_{T_h}\}$ and we will introduce how to store them into user preference memory.

\subsubsection{Memory Addressing}

After obtaining all the vector representations of history sticker and dialog context, we employ a key-value memory network and store them into each key-value slot, as shown in Figure~\ref{fig:preference-memory}.
In this memory network, we use the dialog contexts as the keys and use the corresponding selected stickers as the values.

First, we construct the query from the current dialog context, which will be used to retrieve the user preference representation from the memory network.
We apply a max-pooling layer on the representations of each utterance in the current dialog context:
\begin{align}
    h^i_m &= \text{mean-pooling}(\{h^i_1, \dots, h^i_{T_x}\}), \\
    h &= \max \{h^1_m, \dots, h^{T_u}_m\} , \label{eq:mem-query}
\end{align}
where $h^i_m$ is the representation of $i$-th utterance in the current dialog context, $h \in \mathbb{R}^{d}$ is used as the query and it represents the overall information of the current dialog context.
Next, we use $h$ to calculate the read weights over each memory slot:
\begin{align}
    \delta_k = \text{softmax} ( h W_\delta \hat{h}^k ) ,
\end{align}
where $\delta_k \in [0, 1]$ is the read weight for the $k$-th memory slot and $W_\delta$ is a trainable parameter.

\subsubsection{Memory Reading}

After obtaining the read weights $\{\delta_1, \dots, \delta_{T_h}\}$ for all the memory slots, we can write the semantic output for preference memory by:
\begin{align}
    r = \sum^{T_h}_k { \delta_k \hat{O}_{k} },
\end{align}
where $r$ in essence represents a semantic preference representation and will be used when predicting the sticker in current dialog context.

\subsection{Fusion Network}

\begin{figure}
    \centering
    \includegraphics[scale=0.7]{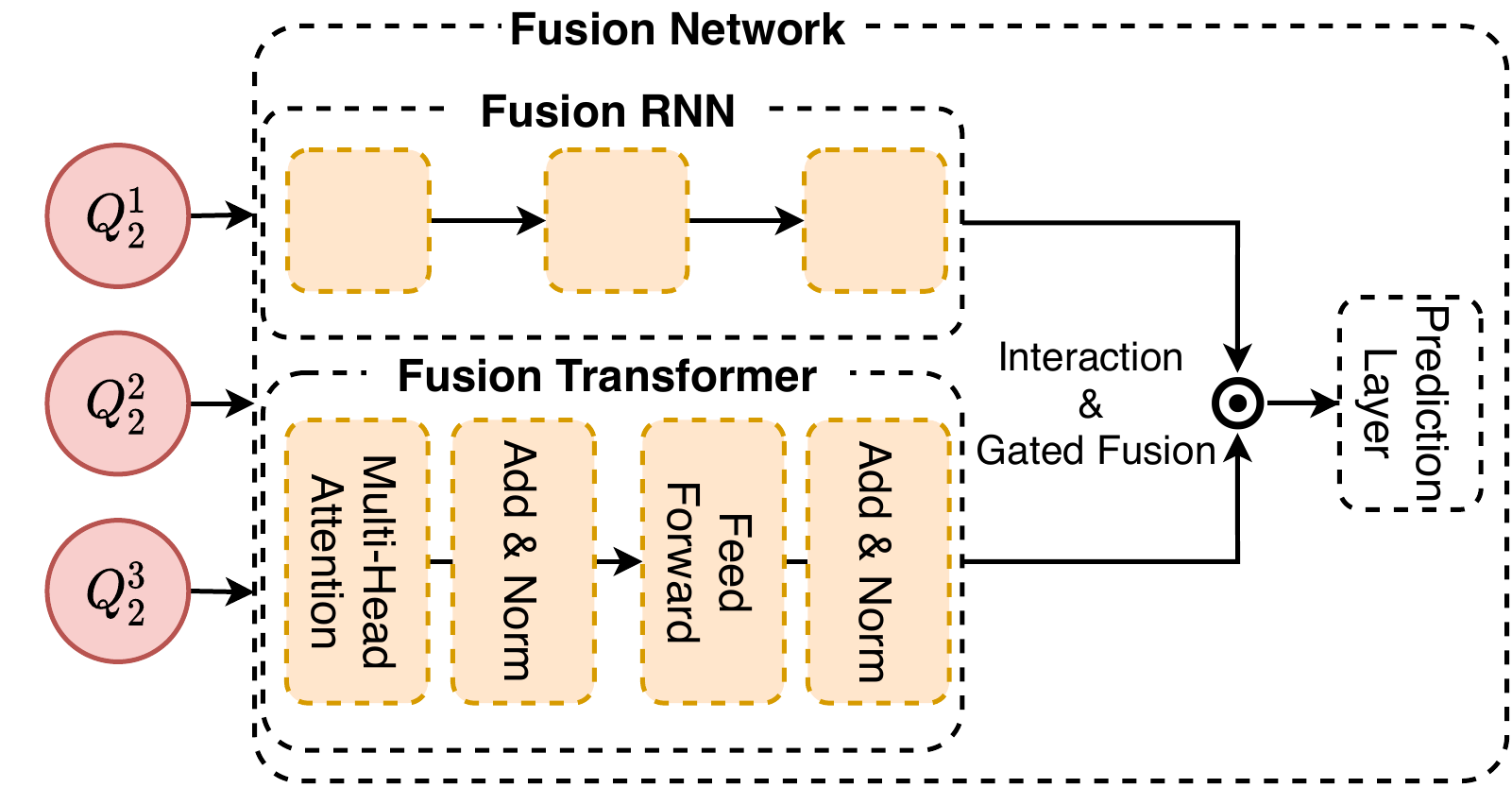}
    \caption{Framework of fusion network. The black circle in the right of the figure indicates the interaction combination and gated fusion.} 
    \label{fig:fusion}
\end{figure}

Up till now, we have obtained the user preference representation and interaction result between each utterance and the candidate sticker.
Here we again include the utterance index $i$ which has been omitted in previous subsections, and $Q_2$ now becomes $Q_2^i$. 
Since the utterances in a multi-turn dialog context are in chronological order, we employ a \textbf{Fusion RNN} and a \textbf{Fusion Transformer} to model the short-term and long-term interaction between utterance $\{Q_2^1, \dots, Q_2^{T_u}\}$.
Fusion RNN (shown in the top part of Figure~\ref{fig:fusion}) is based on the recurrent network which can capture short-term dependency over each utterance interaction result.
Fusion Transformer (shown in the bottom part of Figure~\ref{fig:fusion}) is based on the self-attention mechanism which is designed for capturing the important elements and the long-term dependency among all the interaction results.

\subsubsection{Fusion RNN} \label{subsubsec:fusion-rnn}

Fusion RNN first reads the interaction results for each utterance $\{Q_2^1, \dots, Q_2^{T_u}\}$ and then transforms into a sequence of hidden states.
In this paper, we employ the gated recurrent unit (GRU)~\cite{Chung2014EmpiricalEO} as the cell of fusion RNN, which is popular in sequential modeling~\cite{Gao2019How, Wu2017SequentialMN,tao2019multi}:
\begin{align}
g_i = \text{RNN} \left( Q_2^i, g_{i-1} \right) , \label{equ:fusion-rnn}
\end{align}
where $g_i$ is the hidden state of the fusion RNN and $g_{0}$ is the initial state of RNN which is initialized randomly.
Finally, we obtain the sequence of hidden states $\{g_1, \dots, g_{T_u}\}$.
One can replace GRU with similar algorithms such as Long-Short Term Memory network (LSTM)~\cite{hochreiter1997long}.
We leave the study as future work.

\subsubsection{Fusion Transformer}

To model the long-term dependency and capture the salience utterance from the context, we employ the self-attention mechanism introduced in Equation~\ref{equ:attention}-\ref{equ:ffn}.
Concretely, given $\{Q_2^1, \dots, Q_2^{T_u}\}$, we first employ three linear projection layers with different parameters to project the input sequence into three different spaces:
\begin{align}
\mathcal{Q}^i &= \text{FC} ( Q_2^i ), \\
\mathcal{K}^i &= \text{FC} ( Q_2^i ), \\
\mathcal{V}^i &= \text{FC} ( Q_2^i ).
\end{align}
Then we feed these three matrices into the self-attention algorithm illustrated in Equation~\ref{equ:attention}-\ref{equ:ffn}.
Finally, we obtain the long-term interaction result $\{\hat{g}_1, \dots, \hat{g}_{T_u}\}$.

\subsubsection{Long Short Interaction Combination}

To combine the interaction representation generated by fusion RNN and fusion Transformer, we employ the SUMULTI function proposed by \cite{Wang2016ACM} to combine these representations, which has been proven effective in various tasks:
\begin{align}
    \overline{g}_i = \text{ReLU}(\mathcal{W}^s 
    \begin{bmatrix}
    (\hat{g}_i - g_i) \otimes (\hat{g}_i - g_i) \\
    \hat{g}_i \otimes g_i
   \end{bmatrix}
    + \mathbf{b}^s),
\end{align}
where $\otimes$ is the element-wise product.
The new interaction sequence $\{\overline{g}_1, \dots, \overline{g}_{T_u}\}$ is then boiled down to a matching vector $\Tilde{g}_{T_u}$ by another GRU-based RNN:
\begin{align}
\Tilde{g}_i = \text{RNN}(\Tilde{g}_{i-1}, \overline{g}_i) .
\end{align}
We use the final hidden state $\Tilde{g}_{T_u}$ as the representation of the overall interaction result between the whole utterance context and the candidate sticker.

\subsubsection{Gated Fusion}

In the final prediction, our model combines the current dialog context interaction result and user preference representation to predict the final result.
However, in each case, the information required for current dialog context interaction and user preference representation is not necessarily the same.
If the current dialog context is very similar to the history dialog context, the historical information should play a greater role in prediction.
To incorporate the user preference information into final sticker prediction, we employ a gated fusion which dynamically fuses the current context interaction result and user preference representation together by using a gate $f_g$.
To dynamically fuse these two information sources, we calculate a gate $f_g \in [0, 1]$ which decide which part should the model concentrates on when making the final sticker selection decision:
\begin{equation}
    f_g = \sigma (\text{FC}([r \oplus \Tilde{g}_{T_u}])) , \label{eq:dynamic-fusion}
\end{equation}
where $\sigma$ is the sigmoid function and $\oplus$ denotes the vector concatenation operation.
Next, we apply a weighted sum operation using the gate $f_g$ on current context interaction result $\Tilde{g}_{T_u}$ and user preference representation $r$, as shown in Equation~\ref{eq:prediction-layer}.
Finally, we apply a fully-connected layer to produce the matching score $\hat{y}$ of the candidate sticker:
\begin{equation}
    \hat{y} = \sigma(\text{FC} (f_g * \Tilde{g}_{T_u} + (1 - f_g) * r)) , \label{eq:prediction-layer}
\end{equation}
where $\hat{y} \in (0,1)$ is the matching score of the candidate sticker.

\subsection{Learning}

Recall that we have a candidate sticker set $C=\{c_{1},...c_{T_c}\}$ which contains multiple negative samples and one ground truth sticker.
We use hinge loss as our objective function:
\begin{align}
\mathcal{L} = \sum^{N} \max \left( 0 , \hat{y}_{\text{negative}}- \hat{y}_{\text{positive}} +\text{margin} \right),
 \label{eq:loss-generator}
\end{align}
where $\hat{y}_{\text{negative}}$ and $\hat{y}_{\text{positive}}$ corresponds to the predicted labels of the negative sample and ground truth sticker, respectively.
The margin is the margin rescaling in hinge loss.
The gradient descent method is employed to update all the parameters in our model to minimize this loss function. %
\newcommand{\cbkgrnd}{\cellcolor{blue!15}}
\section{Experimental Setup}
\label{sec:exp-setup}

\subsection{Research Questions}
We list nine research questions that guide the experiments: 

\begin{itemize}
    \item \textbf{RQ1} (See \S~\ref{subsec:Overall}): What is the overall performance of PESRS compared with all baselines?
    \item \textbf{RQ2} (See \S~\ref{subsec:ablation}):  What is the effect of each module in PESRS? 
    \item \textbf{RQ3} (See \S~\ref{subsec:number}): How does the performance change when the number of utterances changes?
    \item \textbf{RQ4} (See \S~\ref{subsec:attention}): 
    Can co-attention mechanism successfully capture the salient part on the sticker image and the important words in dialog context? 
    \item \textbf{RQ5} (See \S~\ref{subsec:features}): What is the influence of the similarity between candidate stickers?
    \item \textbf{RQ6} (See \S~\ref{subsec:hidden}): What is the influence of the parameter settings?
    \item \textbf{RQ7} (See \S~\ref{subsec:history-len}): What is the influence of the user history length?
    \item \textbf{RQ8} (See \S~\ref{subsec:most-select}): What is the performance of using the user's most selected sticker as the response?
    \item \textbf{RQ9} (See \S~\ref{subsec:emoji-analysis}): Can sticker encoder capture the semantic meaning of sticker?
\end{itemize}

\subsection{Comparison Methods}

We first conduct an ablation study to prove the effectiveness of each component in PESRS as shown in Table~\ref{tab:ablations}.
Specifically, we remove each key part of our PESRS to create ablation models and then evaluate the performance of these models.

\begin{table}[t]
\centering
\caption{Ablation models for comparison.}
\label{tab:ablations}
\begin{tabular}{ll}
\toprule
Acronym & Gloss \\
\midrule
PESRS w/o Classify &  \multicolumn{1}{p{8cm}}{\small PESRS w/o emoji classification task}\\
PESRS w/o DIN &  \multicolumn{1}{p{8cm}}{\small PESRS w/o \textbf{D}eep \textbf{I}nteraction \textbf{N}etwork}\\
PESRS w/o FR &  \multicolumn{1}{p{8cm}}{\small PESRS w/o \textbf{F}usion \textbf{R}NN}\\
PESRS FR2T &  \multicolumn{1}{p{8cm}}{\small Change the \textbf{F}usion \textbf{R}NN in PESRS to Transformer with positional encoding}\\
PESRS w/o UPM &  \multicolumn{1}{p{8cm}}{\small PESRS w/o \textbf{U}ser \textbf{P}reference \textbf{M}emory } \\
PESRS w/o TAR &  \multicolumn{1}{p{8cm}}{\small PESRS w/o \textbf{T}ime-\textbf{A}ware \textbf{R}NN } \\
\bottomrule
\end{tabular}
\end{table}

Next, to evaluate the performance of our model, we compare it with the following baselines.
Note that, we adapt VQA and multi-turn response selection models to the sticker response selection task by changing their input text encoder to image encoder.
Since we incorporate the user history data into our model, we also compare with the user modeling method which has been widely used in the recommendation tasks.

\noindent (1) \textbf{SMN}: 
\cite{Wu2017SequentialMN} proposes a sequential matching network to address response selection for the multi-turn conversation problem.
SMN first matches a response with each utterance in the context.
Then vectors are accumulated in chronological order through an RNN.
The final matching score is calculated with RNN.

\noindent (2) \textbf{DAM}: 
\cite{zhou2018multi} extends the transformer model~\cite{vaswani2017attention} to the multi-turn response selection task, where representations of text segments are constructed using stacked self-attention.
Then, truly matched segment pairs are extracted across context and response. 

\noindent (3) \textbf{MRFN}: 
\cite{tao2019multi} proposes a multi-representation fusion network which consists of multiple dialog utterance representation methods and generates multiple fine-grained utterance representations.
Next, they argue that these representations can be fused into final response candidate matching at an early stage, at the intermediate stage or the last stage.
They evaluate all stages and find fusion at the last stage yields the best performance.
This is the state-of-the-art model on the multi-turn response selection task.

\noindent (4) \textbf{Synergistic}:
\cite{guo2019image} devises a novel synergistic network on VQA task.
First, candidate answers are coarsely scored according to their relevance to the image-question pair. 
Afterward, answers with high probabilities of being correct are re-ranked by synergizing with image and question.
This model achieves the state-of-the-art performance on the Visual Dialog v1.0 dataset~\cite{das2017visual}.

\noindent (5) \textbf{PSAC}: 
\cite{li2019beyond} proposes the positional self-attention with co-attention architecture on VQA task, which does not require RNNs for video question answering. 
We replace the output probability on the vocabulary size with the probability on candidate sticker set.

\noindent (6) \textbf{SRS}:
We propose the first sticker selection method consists of the sticker and dialog context encoding module, deep matching network and information fusion layer in our previous work~\cite{gao2020sticker}.
This method achieves the state-of-the-art performance on the multi-turn dialog-based sticker selection dataset.

\noindent (7) \textbf{LSTUR}:
\cite{An2019Neural} proposes a long- and short-term user modeling method to represent the long- and short-term user preference and then apply this method to the news recommendation task.
Experiments on a real-world dataset demonstrate their approach can effectively improve the performance of neural news recommendation method. 
To adapt this method on our sticker selection task, we replace their news encoding network with the sticker image encoding network, Inception-v3, as the same as we used in our model.
Since there are countless users in our task, we can not obtain a static user embedding as they used in their model.
For fair, comparison, we replace the user embedding in their model to current dialog context.

For the first three multi-turn response selection baselines, we replace the candidate utterance embedding RNN or Transformer network with the image encoding CNN network Inception-v3, which is the same as used in our proposed model.
This Inception-v3 network is initialized using a pre-trained model\footnote{\url{https://github.com/tensorflow/models/tree/master/research/slim}} for all baselines and PESRS.

\subsection{Evaluation Metrics}

Following~\cite{tao2019multi,zhou2018multi}, we employ recall at position $k$ in $n$ candidates $R_n@k$ as an evaluation metric, which measures if the positive response is ranked in the top $k$ positions of $n$ candidates.
Following~\cite{zhou2018multi}, we also employ mean average precision (MAP)~\cite{baeza2011modern} as an evaluation metric.
The statistical significance of differences observed between the performance of two runs is tested using a two-tailed paired t-test and is denoted using \dubbelop\ (or \dubbelneer) for strong significance at $\alpha=0.01$.

\subsection{Implementation Details}

We implement our experiments using TensorFlow~\cite{abadi2016tensorflow} on an NVIDIA GTX 2080Ti GPU. 
If the number of words in an utterance is less than 30, we pad zeros, otherwise, the first 30 words are kept.
The word embedding dimension is set to 100 and the number of hidden units is 100.
The batch size is set to 32.
9 negative samples are randomly sampled from the sticker set containing the ground truth sticker, and we finally obtain 10 candidate stickers for the model to select.
We initialize all the parameters randomly using a Gaussian distribution in [-0.02, 0.02].
We use Adam optimizer~\cite{Kingma2015AdamAM} as our optimizing algorithm, and the learning rate is $1 \times 10^{-4}$.

\section{Experimental result}
\label{sec:exp-result}

\subsection{Overall Performance}
\label{subsec:Overall}

\begin{table}[t]
\centering
\caption{RQ1: Automatic evaluation comparison. Significant differences are with respect to MRFN.}
\begin{tabular}{@{}l cc cc @{}}
\toprule
& MAP & $R_{10}@1$ & $R_{10}@2$  & $R_{10}@5$   \\
\midrule
\multicolumn{5}{@{}l}{\emph{Visual Q\&A methods}}\\
Synergistic & 0.593 \phantom{0}  & 0.438\phantom{0} & 0.569\phantom{0} & 0.798\phantom{0} \\
PSAC & 0.662\phantom{0}  & 0.533\phantom{0} & 0.641\phantom{0} & 0.836\phantom{0} \\
\midrule
\multicolumn{5}{@{}l}{\emph{Multi-turn response selection methods}}\\
SMN & 0.524\phantom{0}  & 0.357\phantom{0} & 0.488\phantom{0} & 0.737\phantom{0} \\
DAM & 0.620\phantom{0}   & 0.474\phantom{0} & 0.601\phantom{0} & 0.813\phantom{0} \\
MRFN & 0.684\phantom{0}  & 0.557\phantom{0} & 0.672\phantom{0} & 0.853\phantom{0}\\
LSTUR & 0.689  & 0.558 & 0.68 & 0.874 \\
\midrule
SRS & 0.709  & 0.590 & 0.703 & 0.872 \\
PESRS & \textbf{0.743}  & \textbf{0.632}\dubbelop & \textbf{0.740}\dubbelop & \textbf{0.897} \\
\bottomrule
\end{tabular}
\label{tab:comp_auto_baselines}
\end{table}

\begin{figure*} 
    \centering 
    \subfigure[$MAP$ score]{ 
        \label{figs:MAP.png} %
        \includegraphics[scale=0.40]{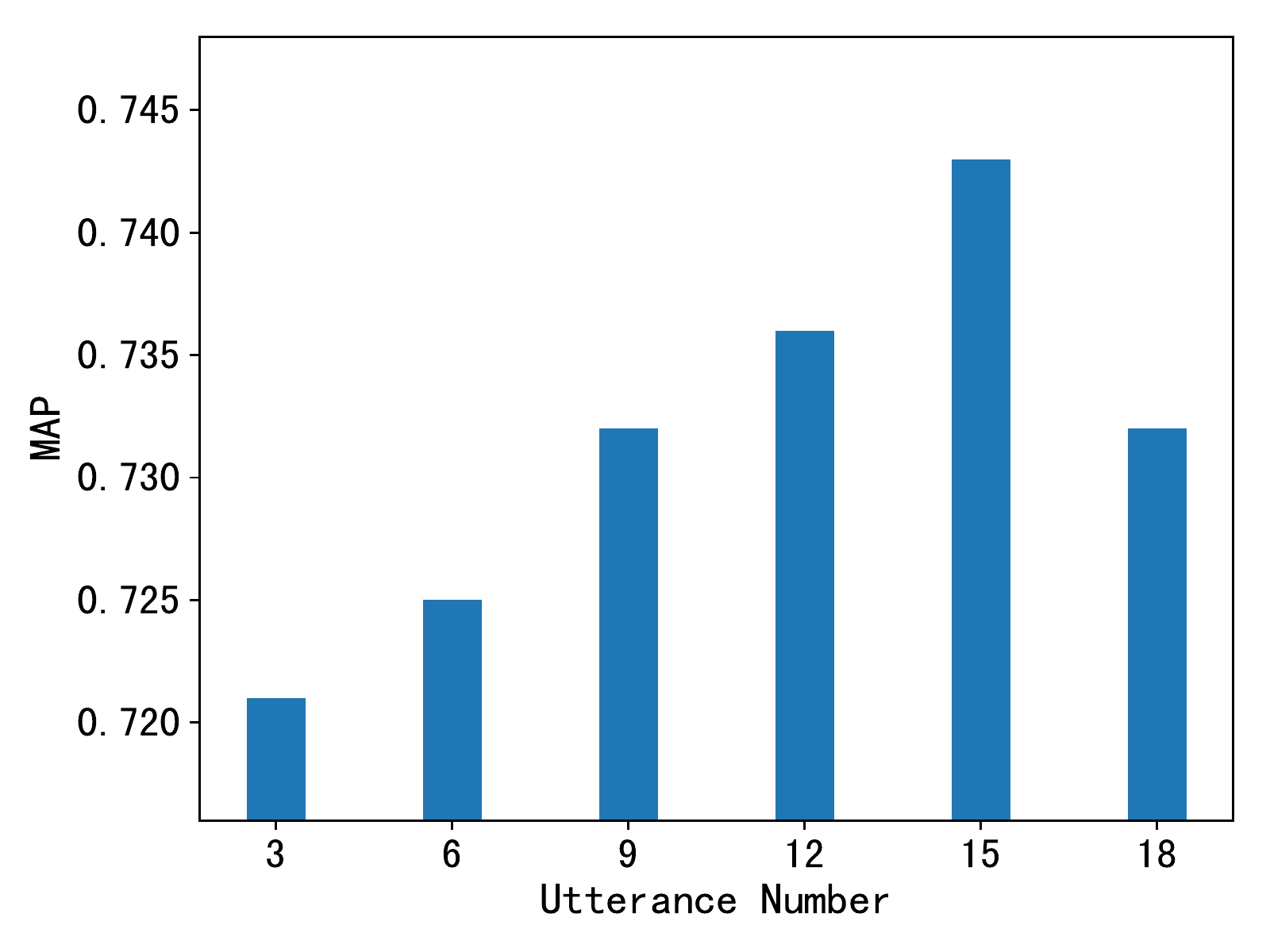}
    } 
    \subfigure[$R_{10}@1$ score]{ 
        \label{figs:r1.png} %
        \includegraphics[scale=0.40]{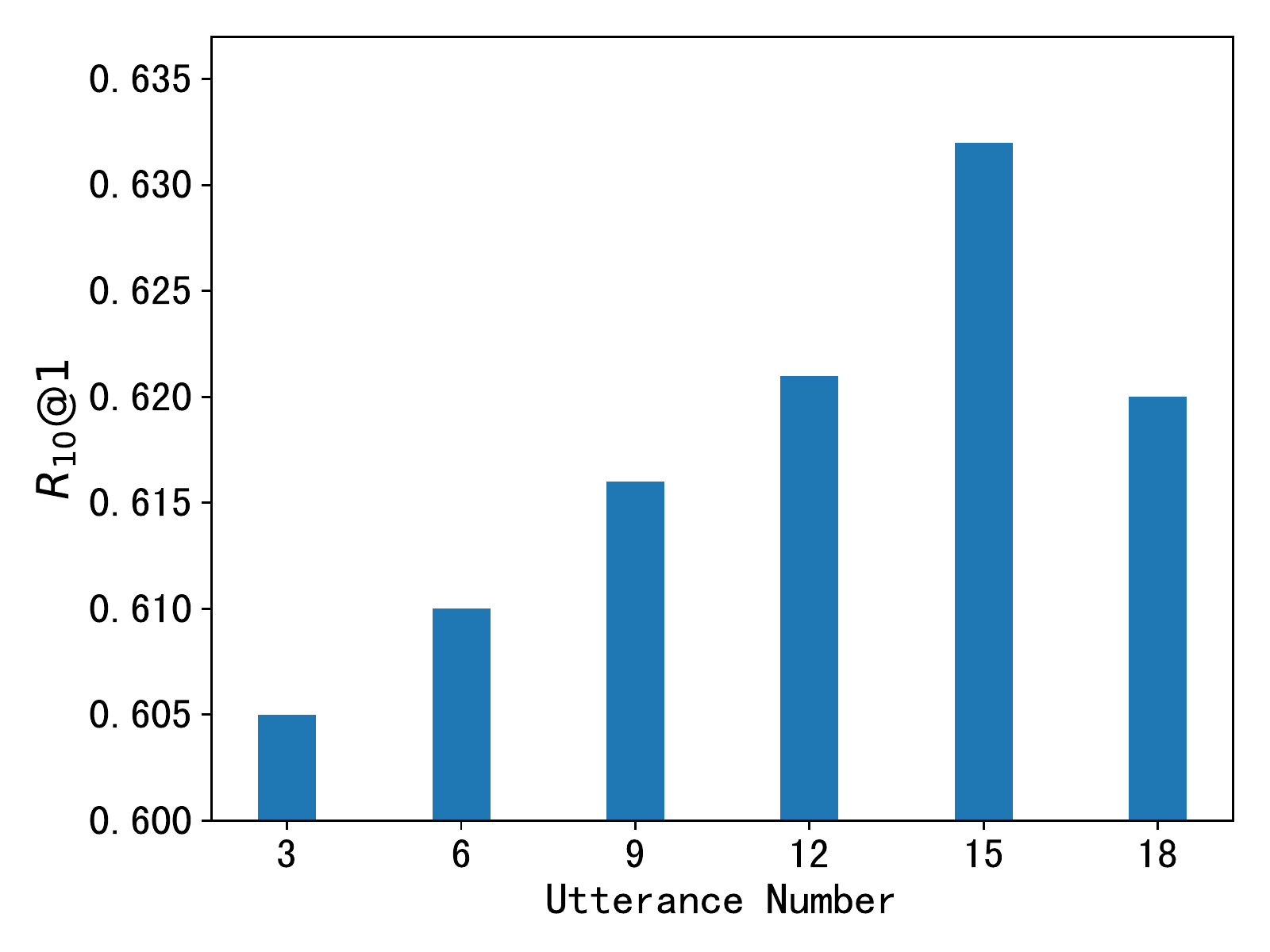}
    } 
    \subfigure[$R_{10}@2$ score]{ 
        \label{figs:r2.png} %
        \includegraphics[scale=0.40]{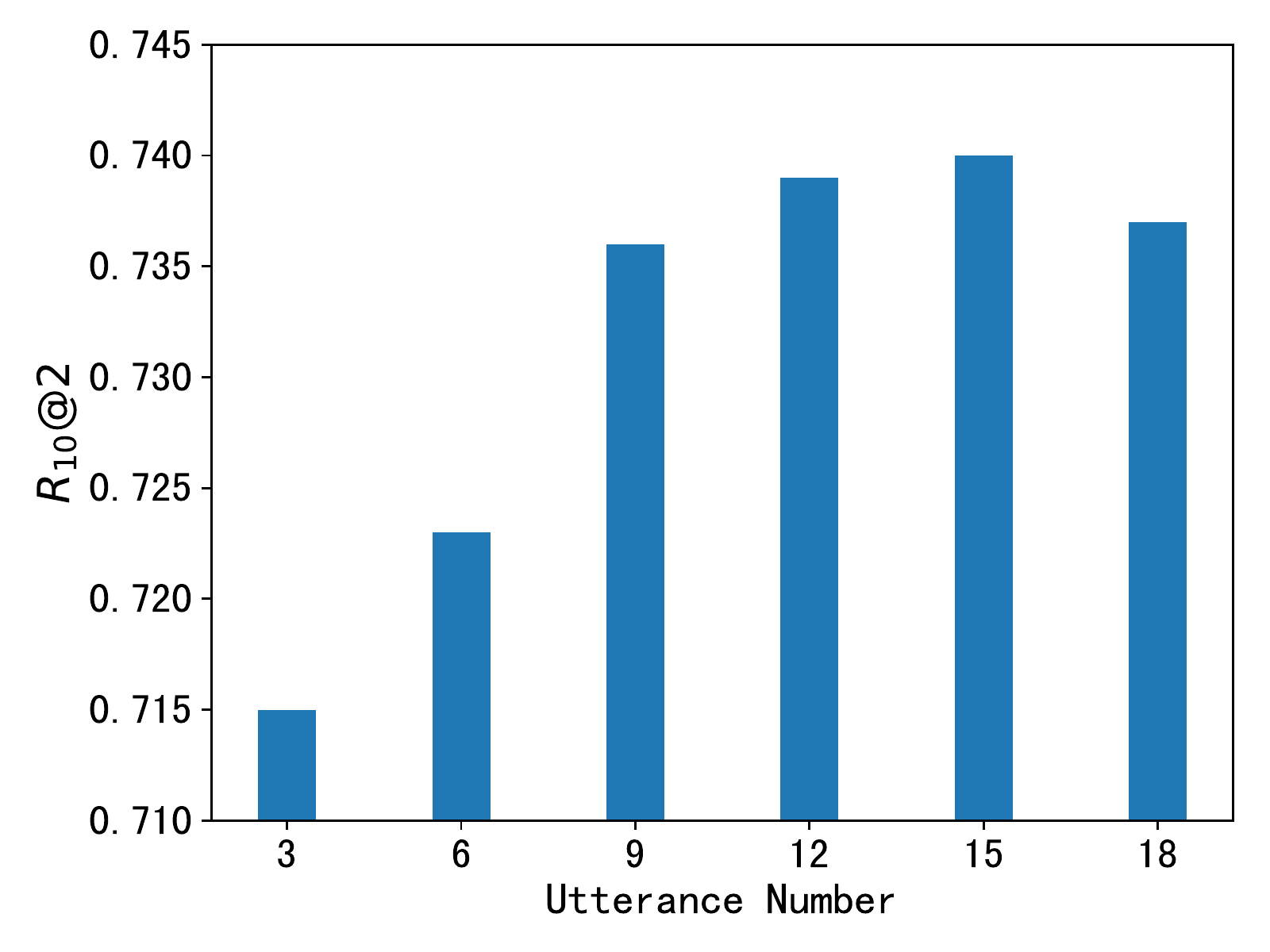}
    } 
    \subfigure[$R_{10}@5$ score]{ 
        \label{figs:r5.png} %
        \includegraphics[scale=0.40]{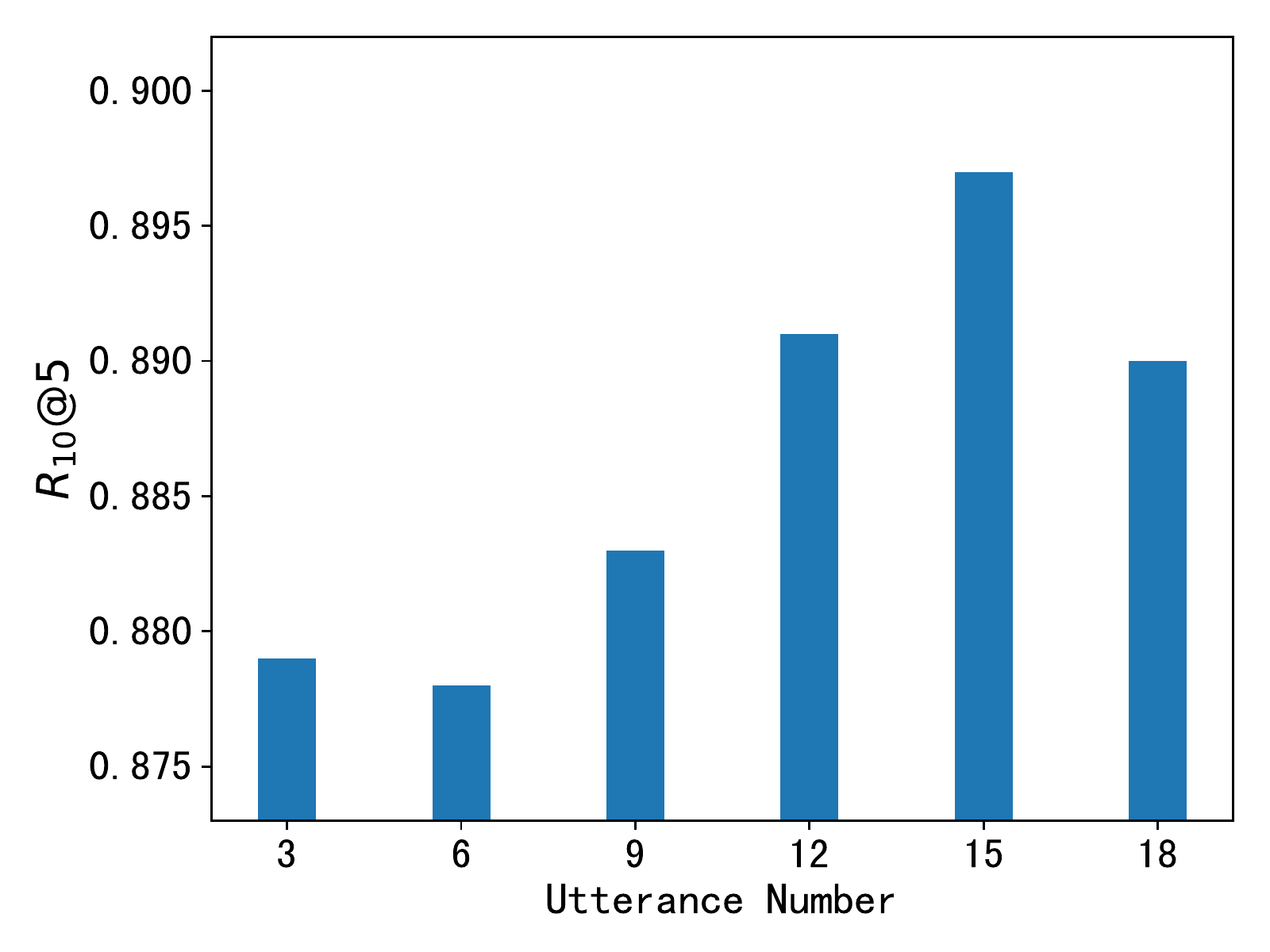}
    } 
    \caption{
         RQ3: Performance of PRSRS on all metrics when reading different number of utterances.
    }
    \label{fig:turns}
\end{figure*}

For research question \textbf{RQ1}, we examine the performance of our model and baselines in terms of each evaluation metric, as shown in Table~\ref{tab:comp_auto_baselines}.
First, the performance of the multi-turn response selection models is generally consistent with their performances on text response selection datasets.
SMN~\cite{Wu2017SequentialMN}, an earlier work on multi-turn response selection task with a simple structure, obtains the worst performance on both sticker response and text response selection.
DAM~\cite{zhou2018multi} improves the SMN model and gets better performance.
MRFN~\cite{tao2019multi} is the state-of-the-art text response selection model and achieves the best performance among baselines in our task as well.
Second, VQA models perform generally worse than multi-turn response selection models, since the interaction between multi-turn utterances and sticker is important, which is not taken into account by VQA models. %
Third, our previously proposed SRS achieves better performance with 3.36\%, 5.92\% and 3.72\% improvements in MAP, $R_{10}@1$ and $R_{10}@2$ respectively, over the state-of-the-art multi-turn selection model, \ie MRFN, and with 6.80\%, 10.69\% and 8.74\% significant increases (with p-value < 0.05) over the state-of-the-art visual dialog model, PSAC. 
Finally, comparing with our previously proposed sticker selection method SRS, our newly proposed model PESRS which incorporates the user preference information achieves the state-of-the-art performance with 4.8\%, 7.1\% and 5.3\% improvements in $MAP$, $R_{10}@1$ and $R_{10}@2$ respectively, over our previous method SRS which is just based on the multi-modal matching between utterance and sticker image.
That demonstrates the superiority of incorporating the user preference information into sticker selection model.

\subsection{Ablation Study}
\label{subsec:ablation}

\begin{table}[t]
    \centering
    \caption{RQ2: Evaluation of different ablation models.}
    \begin{tabular}{@{}lcc cc@{}}
        \toprule
        & MAP & $R_{10}@1$ & $R_{10}@2$ & $R_{10}@5$ \\
        \midrule
        PESRS w/o Classify & 0.714 & 0.598  & 0.707 & 0.866  \\
        PESRS w/o DIN & 0.728  & 0.612  & 0.725 & 0.888 \\
        PESRS w/o FR & 0.727  & 0.609 & 0.725 & 0.886 \\
        PESRS FR2T & 0.725  & 0.610 & 0.719 & 0.881 \\
        PESRS w/o UPM & 0.709  & 0.590 & 0.703 & 0.872 \\
        PESRS w/o TAR & 0.710  & 0.589 & 0.706 & 0.873 \\
        PESRS & \textbf{0.743}  & \textbf{0.632}\dubbelop & \textbf{0.740}\dubbelop & \textbf{0.897} \\
        \bottomrule
    \end{tabular}
    \label{tab:comp_rouge_ablation}
\end{table}

For research question \textbf{RQ2}, we conduct ablation tests on the use of the sticker classification loss (introduced in \S~\ref{subsec:sticker_encoder}), the deep interaction network (introduced in \S~\ref{deep_int}), the fusion RNN (introduced in \S~\ref{subsubsec:fusion-rnn}), the user preference memory without position aware RNN (introduced in \S~\ref{subsec:preference}) and the full user preference memory (introduced in \S~\ref{subsubsec:history-encoding}) respectively.  
The evaluation results are shown in Table~\ref{tab:comp_rouge_ablation}.
The performances of all ablation models are worse than that of PESRS under all metrics, which demonstrates the necessity of each component in PESRS.
We also find that the sticker classification makes contribution to the overall performance.
And this additional task can also speed up the training process, and help our model to converge quickly.
We use 21 hours to train the PESRS until convergence, and we use 35 hours for training PESRS w/o Classify.
The fusion RNN brings a significant contribution (with p-value < 0.05), improving the $MAP$ and $R_{10}@1$ scores by 2.2\% and 3.8\%, respectively.
We also change the fusion RNN to a Transformer with positional encoding, which leads to a decrease of the performance that verifies the effectiveness of fusion RNN.
The deep interaction network also plays an important part. 
Without this module, the interaction between the sticker and utterance are hindered, leading to a 3.3\% drop in $R_{10}@1$.
Particularly, since the user preference memory capture the preference of user's sticker selection, we can see that when the user preference memory is removed from the model, the model suffers from dramatic performance drop in terms of all metrics.
And the position-aware user history encoding RNN also makes contribution to the PESRS model, improving the $MAP$ and $R_{10}@1$ scores by 4.6\% and 7.3\%, respectively.

\subsection{Analysis of Number of Utterances} \label{subsec:number}

For research question \textbf{RQ3}, in addition to comparing with various baselines, we also evaluate our model when reading different number of utterances to study how the performance relates to number of context turns.

Figure~\ref{fig:turns} shows how the performance of the PESRS changes with respect to different numbers of utterances turns.
In this experiment, we change the numbers of utterances turns in both current dialog context and history dialog contexts.
We observe a similar trend for PESRS on the first three evaluation metrics $MAP$, $R_{10}@1$ and $R_{10}@2$: they first increase until the utterance number reaches 15, and then fluctuate as the utterance number continues to increase.
There are two possible reasons for this phenomena.
The first reason might be that, when the information in the utterances is limited, the model can capture the features well, and thus when the amount of information increases, the performance gets better.
However, the capacity of the model is limited, and when the amount of information reaches its upper bound, it gets confused by this overwhelming information.
The second reason might be of the usefulness of the utterance context.
Utterances that occur too early before the sticker response may be irrelevant to the sticker and bring unnecessary noise.
As for the last metric, the above observations do not preserve.
The $R_{10}@5$ scores fluctuate when the utterance number is below 15, and drop when the utterance number increases.
The reason might be that $R_{10}@5$ is not a strict metric, and it is easy to collect this right sticker in the set of half of the whole candidates.
Thus, the growth of the information given to PESRS does not help it perform better but the noise it brings harms the performance.
On the other hand, though the number of utterances changes from 3 to 18, the overall performance of PESRS generally remains at a high level, which proves the robustness of our model.

\begin{figure*}[h]
    \centering
    \includegraphics[scale=0.50]{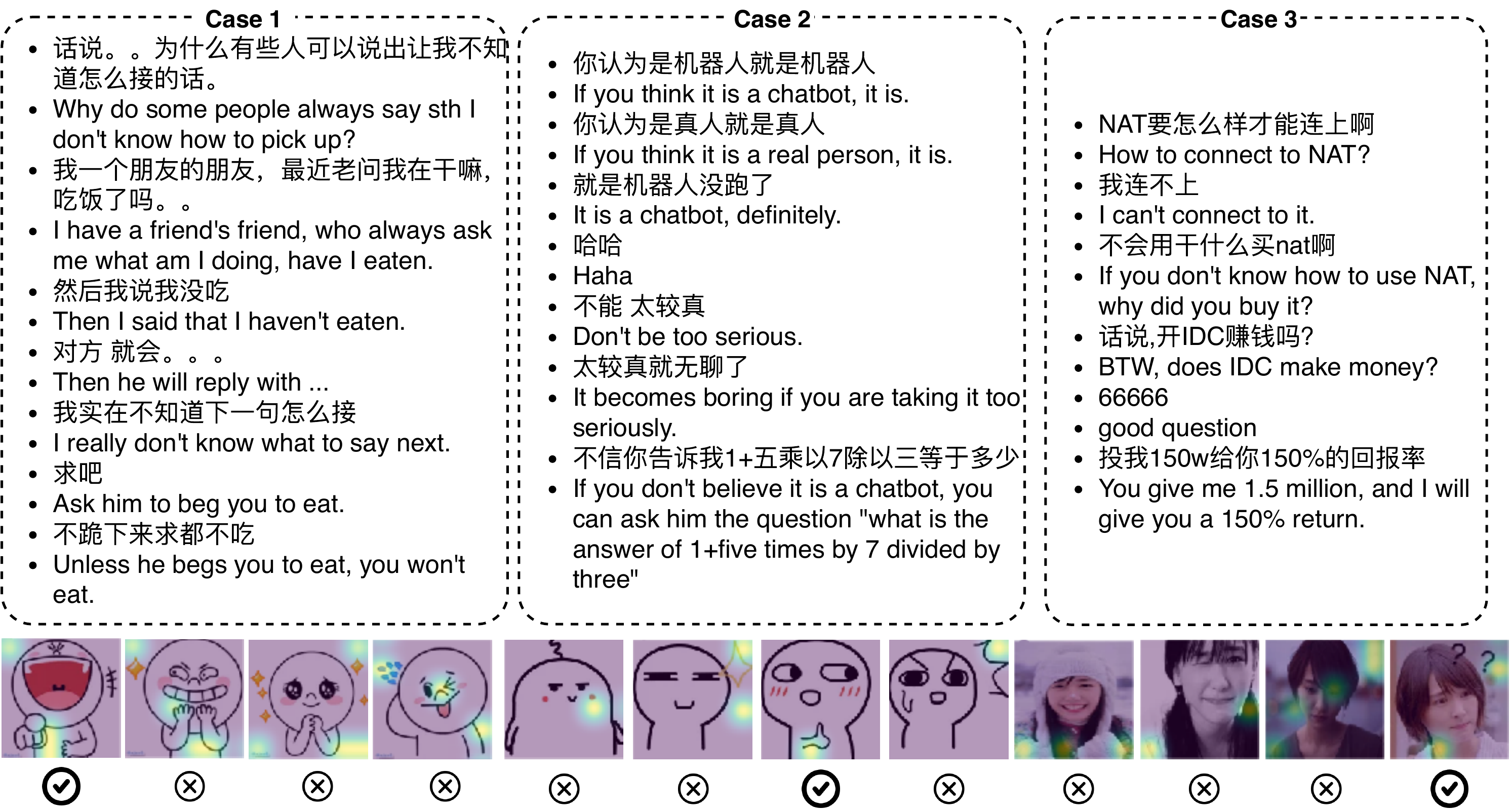}
    \caption{
        RQ4: Examples of sticker selection results produced by SRS. We show the selected sticker and three random selected candidate stickers with the attention heat map. The lighter the area on image is, the higher attention weight it gets. The first two cases are collected from a chitchat group, and the third one is collected from a VPN custom service group.
    }
    \label{fig:predict-case}
\end{figure*}

\begin{figure}[h]
    \centering
    \includegraphics[scale=0.6]{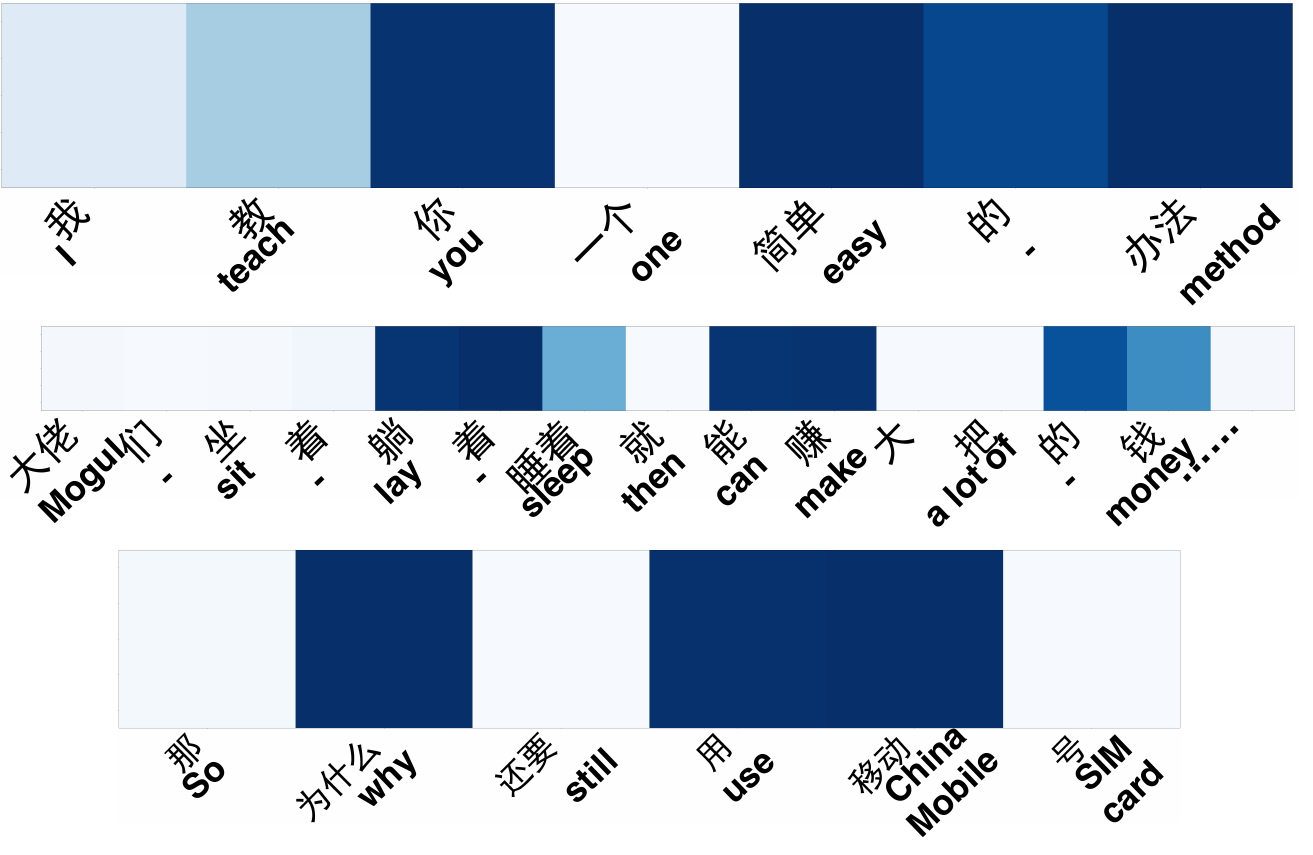}
    \caption{
        RQ4: Examples of the attention weights of the dialog utterance. We translate Chinese to English word by word. The darker the area is, the higher weight the word gets.
    }
    \label{fig:text-attention-case}
\end{figure}

\subsection{Analysis of Attention Distribution in Interaction Process}
\label{subsec:attention}

Next, we turn to address \textbf{RQ4}.
We also show three cases with the dialog context in Figure~\ref{fig:predict-case}.
There are four stickers under each dialog context, one is the selected sticker by our model and other three stickers are random selected candidate stickers.
As a main component, the deep interaction network comprises a bi-directional attention mechanism between the utterance and the sticker, where each word in the utterance and each unit in the sticker representation have a similarity score in the co-attention matrix.
To visualize the sticker selection process and to demonstrate the interpretability of deep interaction network, we visualize the sticker-wise attention $\tau^s$ (Equation~\ref{equ:ua-sticker}) on the original sticker image and show some examples in Figure~\ref{fig:predict-case}.
The lighter the area is, the higher attention it gets.

Facial expressions are an important part in sticker images.
Hence, we select several stickers with vivid facial expression in Figure~\ref{fig:predict-case}.
Take forth sticker in Case 1 for example where the character has a wink eye and a smiling mouth.
The highlights are accurately placed on the character's eye, indicating that the representation of this sticker is highly dependent on this part.
Another example is the last sticker of Case 3, there is two question marks on the top right corner of the sticker image which indicates that the girl is very suspicious of this.
In addition to facial expression, the characters gestures can also represent emotions.
Take the third sticker in Case 2 for example, the character in this sticker gives a thumbs up representing support and we can find that the attention lies on his hand, indicating that the model learns the key point of his body language.

Furthermore, we randomly select three utterances from the test dataset, and we also visualize the attention distribution over the words in an utterance, as shown in Figure~\ref{fig:text-attention-case}.
We use the weight $\tau_j^u$ for the $j$-th word (calculated in Equation~\ref{equ:sa-utterance}) as the attention weight.
We can find that the attention module always gives a higher attention weight on the salience word, such as the ``easy method'', ``make a lot of money'' and ``use Chine Mobile''.

\subsection{Influence of Similarity between Candidates}
\label{subsec:features}

\begin{figure}[h]
    \centering
    \includegraphics[scale=0.60]{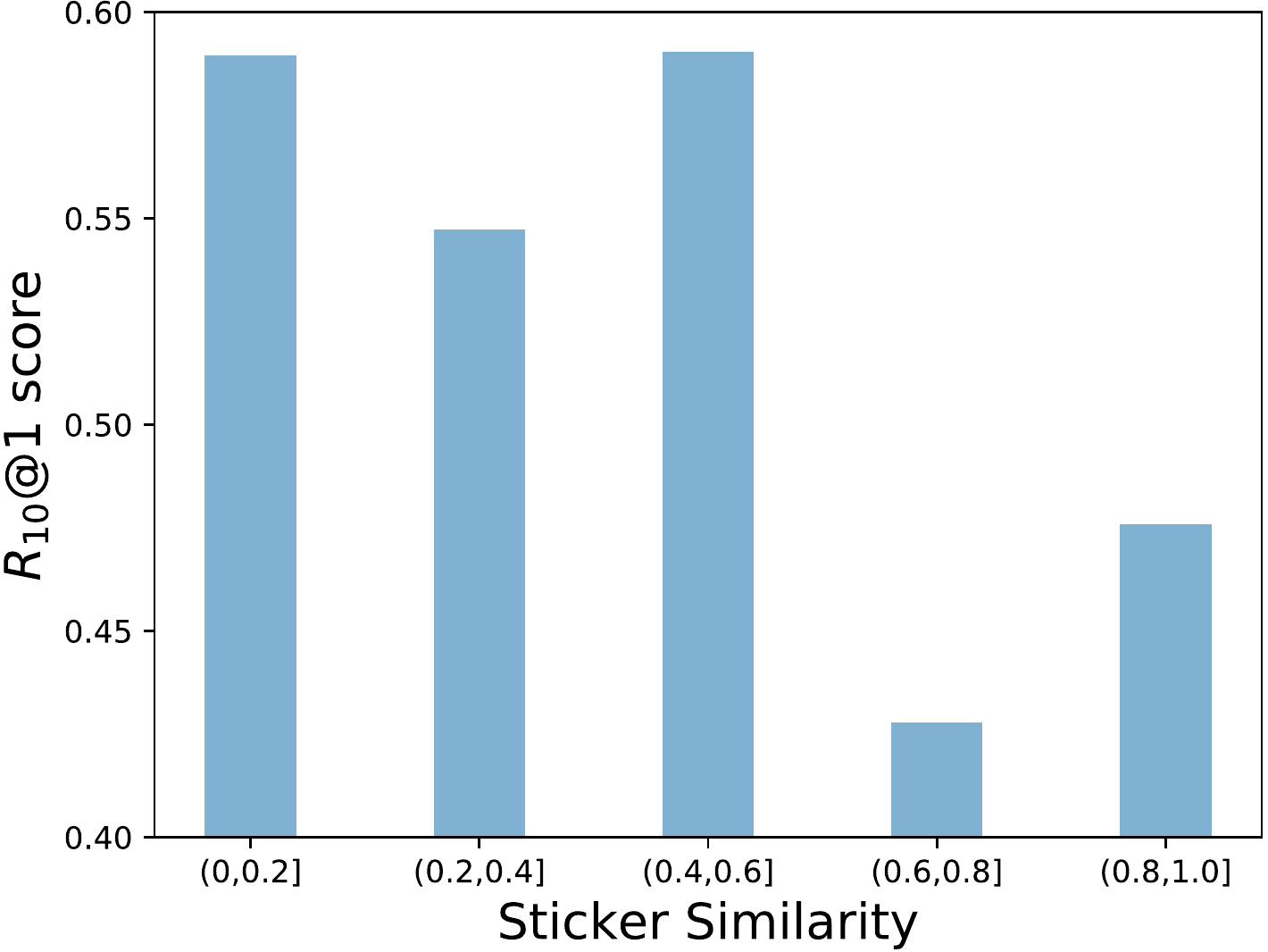}
    \caption{
        RQ5: Performance of SRS on groups of different candidate similarity.
    }
    \label{fig:similarity-recall}
\end{figure}

In this section, we turn to \textbf{RQ5} to investigate the influence of the similarities between candidates.
The candidate stickers are sampled from the same set, and stickers in a set usually have a similar style.
Thus, it is natural to ask: Can our model identify the correct sticker from a set of similar candidates?
What is the influence of the similarity between candidate stickers?
Hence, we use the Structural Similarity Index (SSIM) metric~\cite{wang2004image,avanaki2008exact} to calculate the average similarity among all candidates in a test sample and then aggregate all test samples into five groups according to their average similarities.
We calculate the $R_{10}@1$ of each group of samples, as shown in Figure~\ref{fig:similarity-recall}.
The x-axis is the average similarity between candidate stickers and the y-axis is the $R_{10}@1$ score.

Not surprisingly, our model gains the best performance when the average similarity of the candidate group is low and its performance drops as similarity increases.
However, we can also see that, though similarity varies from minimum to maximum, the overall performance can overall stay at high level. 
$R_{10}@1$ scores of all five groups are above 0.42, and the highest score reaches 0.59.
That is, our model is highly robust and can keep giving reasonable sticker responses.

\subsection{Robustness of Parameter Setting}\label{subsec:hidden}

\begin{figure}[h]
    \centering
    \includegraphics[scale=0.6]{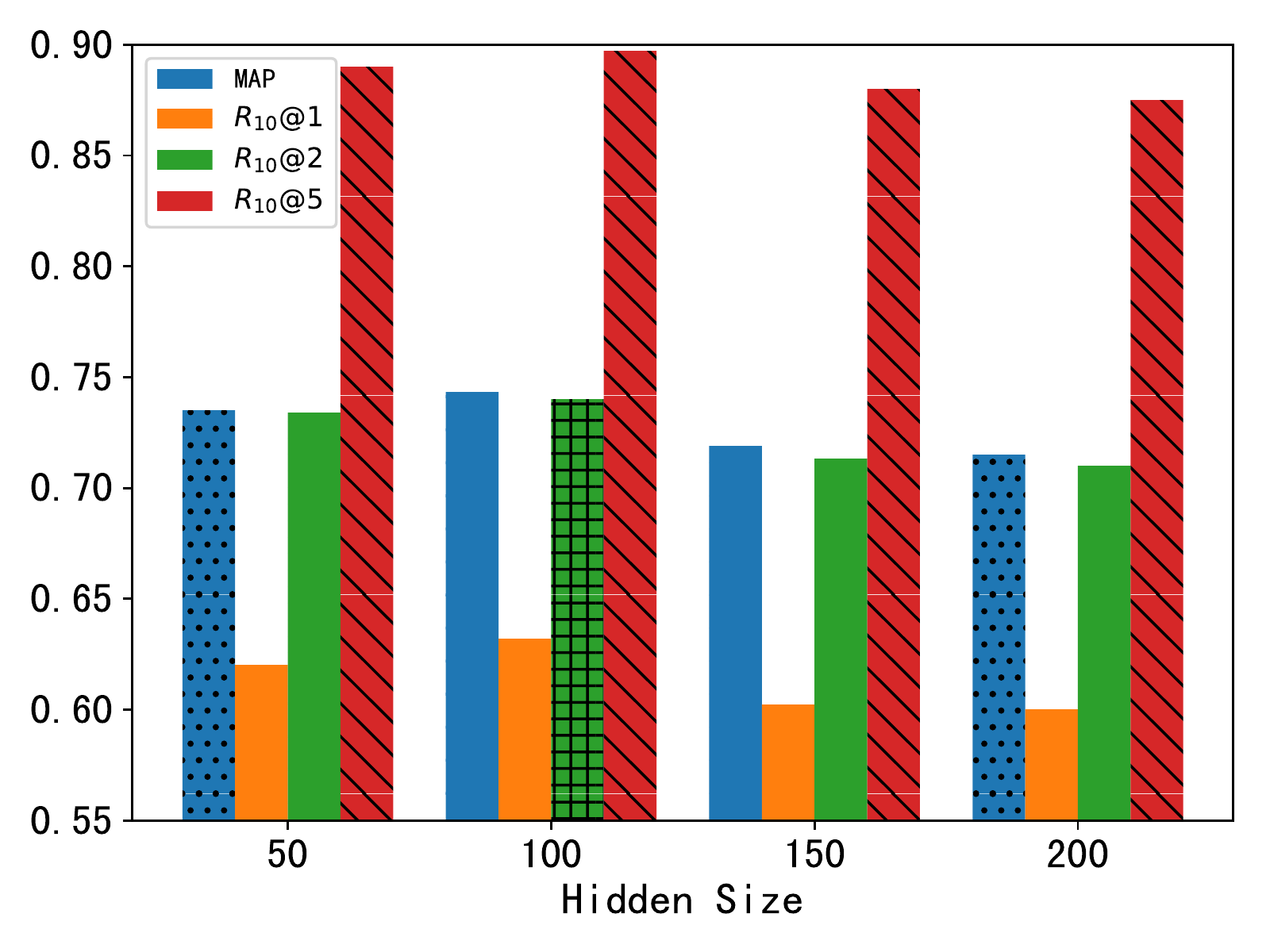}
    \caption{
        RQ6: Performance of PESRS with different parameter settings.
    }
    \label{fig:hidden}
\end{figure}

In this section, we turn to address \textbf{RQ6} to investigate the robustness of parameter setting.
We train PESRS model in different parameter setting as shown in Figure~\ref{fig:hidden}.
The hidden size of the RNN, CNN and the dense layer in our model is tuned from 50 to 200, and we use the MAP and $R_n@k$ to evaluate each model.
As the hidden size grows larger from 50 to 100, the performance rises too.
The increment of hidden size improves the MAP and $R_{10}@1$ scores by 1.1\% and 1.9\%.
When the hidden size continuously goes larger from 100 to 200, the performance is declined slightly.
The increment of hidden size leads to a 3.9\% and 5.3\% drop in terms of MAP and $R_{10}@1$ respectively.
Nonetheless, we can find that each metric maintained at a stable interval, which demonstrates that our PESRS is robust in terms of the parameter size.

\subsection{Influence of User History Length} \label{subsec:history-len}

\begin{figure*} 
    \centering 
    \subfigure[$MAP$ score]{ 
        \label{figs:MAP.png} %
        \includegraphics[scale=0.40]{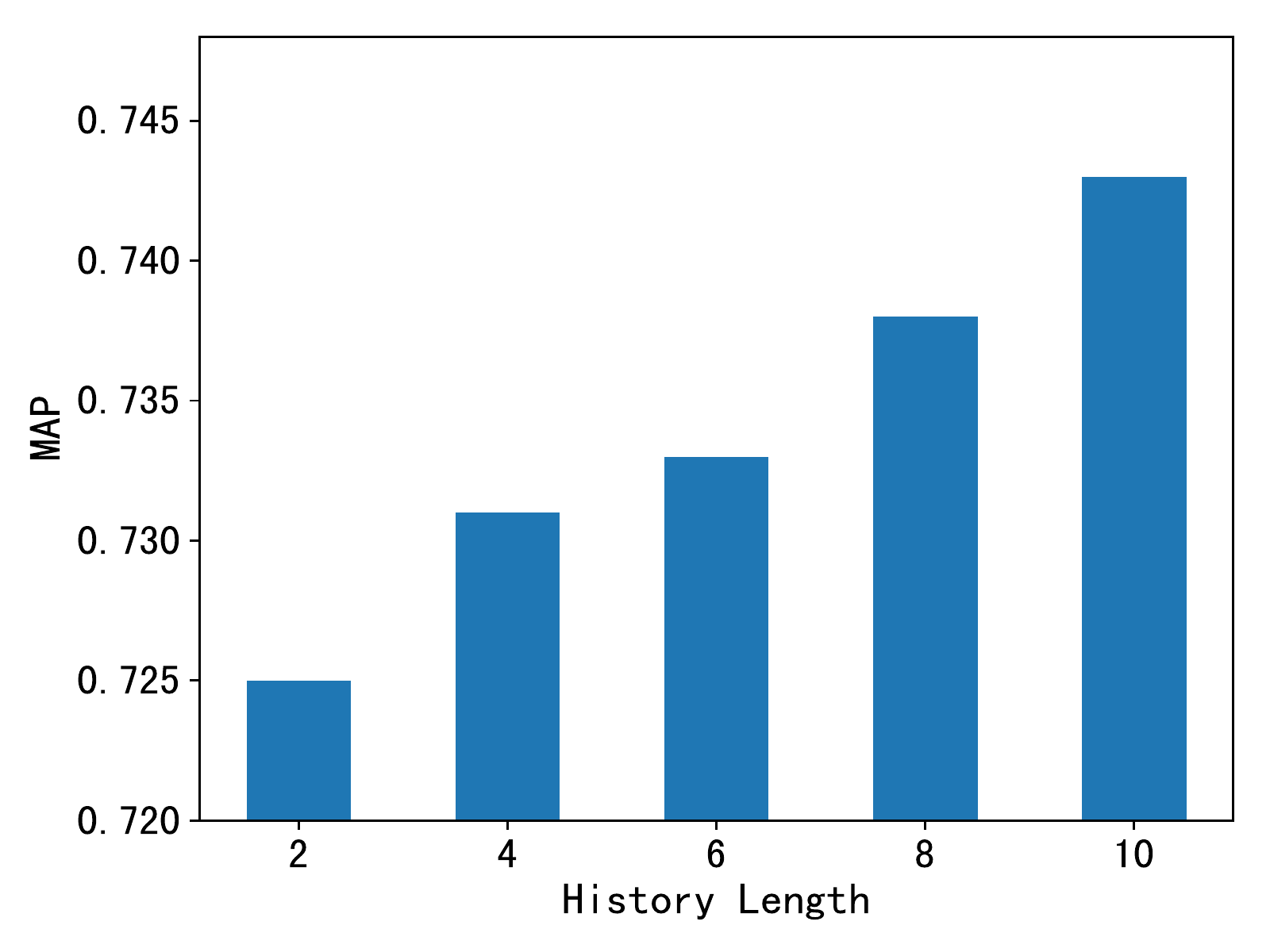}
    } 
    \subfigure[$R_{10}@1$ score]{ 
        \label{figs:r1.png} %
        \includegraphics[scale=0.40]{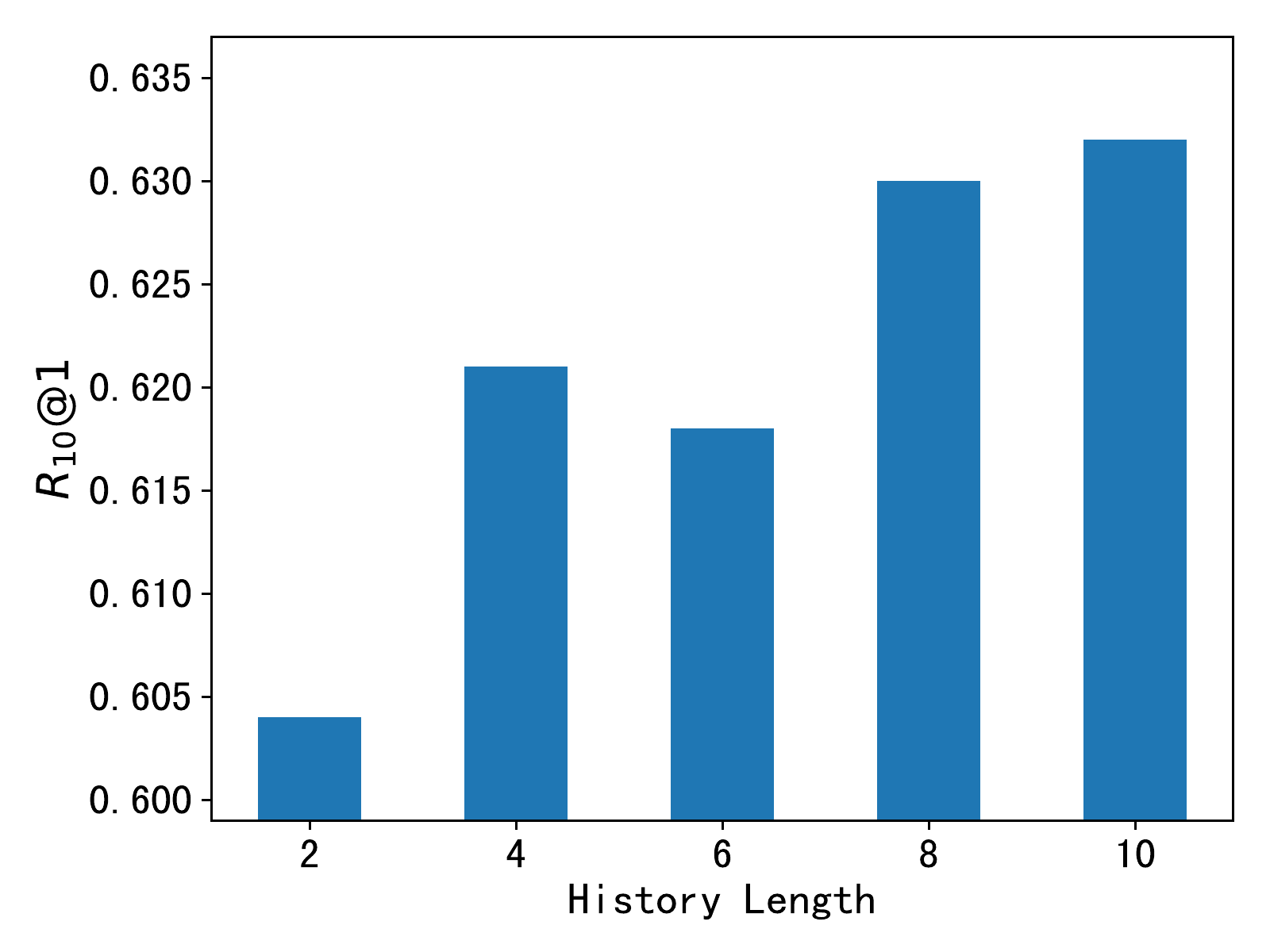}
    } 
    \subfigure[$R_{10}@2$ score]{ 
        \label{figs:r2.png} %
        \includegraphics[scale=0.40]{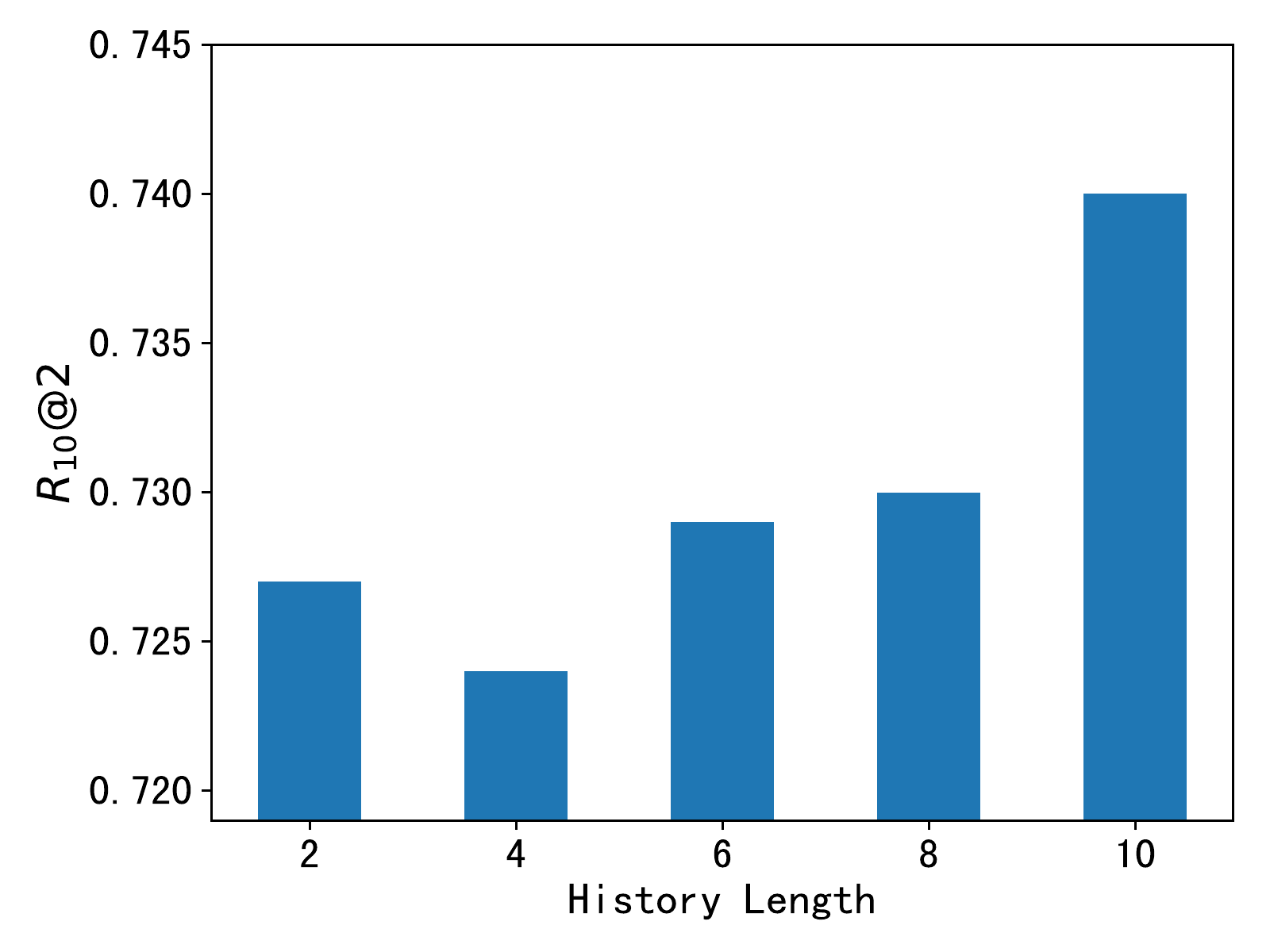}
    } 
    \subfigure[$R_{10}@5$ score]{ 
        \label{figs:r5.png} %
        \includegraphics[scale=0.40]{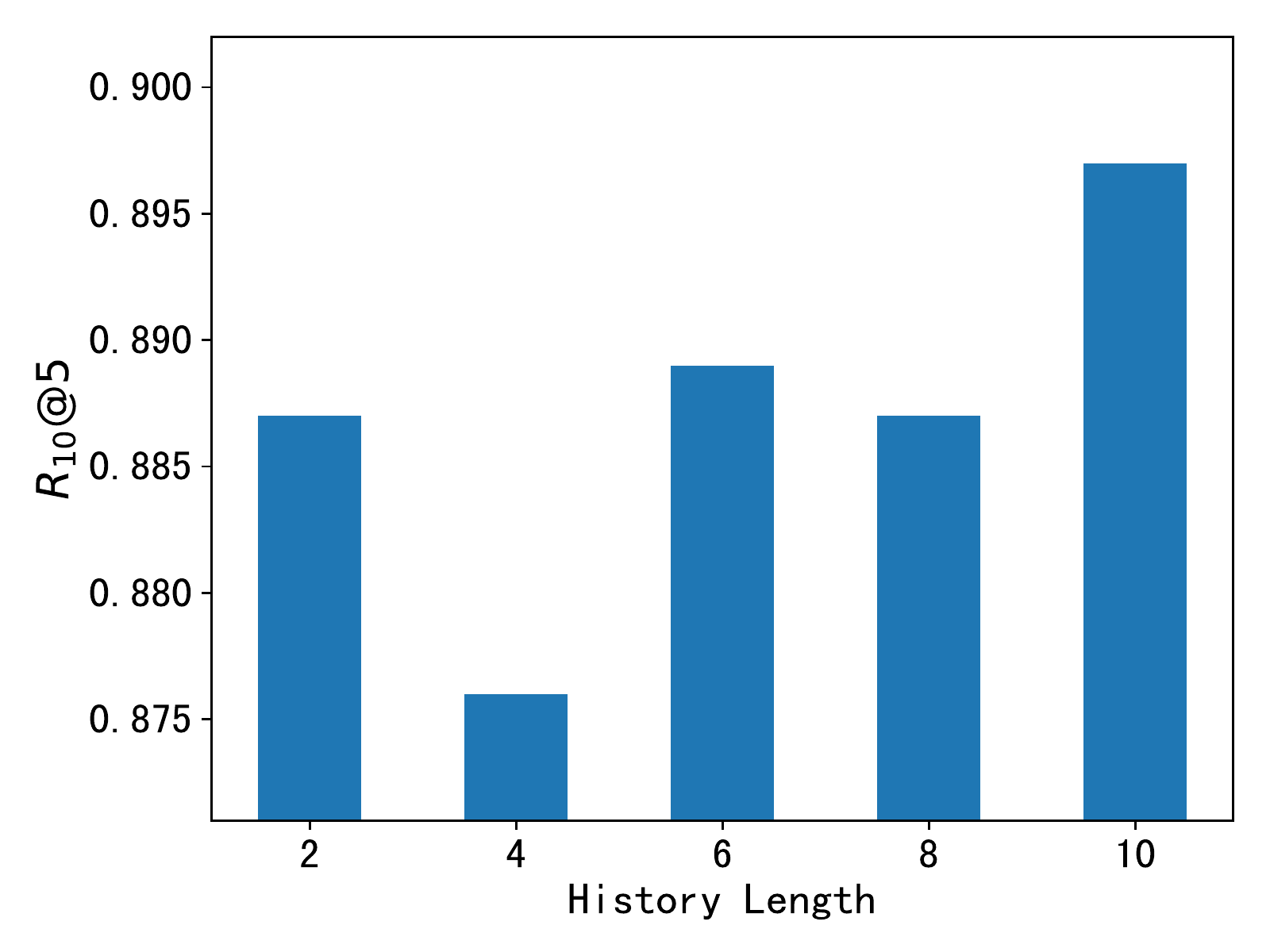}
    } 
    \caption{
        RQ7: Performance of PESRS with different user history length.
    }
    \label{fig:history-len}
\end{figure*}

Next, we address \textbf{RQ7} which focuses on the influence of using different lengths of user history.
We feed different lengths of user sticker selection history to the model, and we show the model performance of different lengths in Figure~\ref{fig:history-len}.
From this figure, we can find that the model performs worse when we just feed only $2$ user sticker selection history.
The sticker selection prediction performance of the model rises sharply as the history length increases. 
This indicates that it requires large amount of user behavior patterns for modeling the preference of the user. 
And the growth of user behavior sequence helps PESRS to better capture sticker selection patterns according to the dialog context.

\subsection{Analysis of User Preference Memory} \label{subsec:most-select}

Next, we turn to \textbf{RQ8} to investigate the effectiveness of user preference modeling module.
We propose a simple heuristic method and two variations of our user preference memory module.

To verify the necessity of using user preference modeling network, we use a simple heuristic method (\textit{MostSelected}) which just use the most selected sticker by user as the  sticker prediction of current dialog context.
This method does not consider the semantic matching degree of previous dialog context and current dialog context.
Consequently, the predicted sticker of this heuristic method is not flexible.

The first variation (\textit{AverageMem}) is to simply apply an average-pooling layer on all the previous selected sticker representations by the corresponding user:
\begin{align}
    r = \sum^{T_h}_k { \hat{O}_{k} }.
\end{align}
Then we use this as the user preference representation and feed $r$ to the final gated fusion layer, as shown in Equation~\ref{eq:dynamic-fusion}.

The second variation (\textit{WeightedMem}) is to remove the key addressing process, and directly apply an attention-then-weighted method on all the user previous selected stickers.
This variation can be split into two steps: (1) calculate attention weights and (2) weighted sum stickers.
We use the query vector $h$ (shown in Equation~\ref{eq:mem-query}) to calculate attention weights of each user previously selected stickers $\{\hat{O}_{1}, \dots, \hat{O}_{T_h}\}$, and the query vector $h$ is the same as used in our proposed user preference memory module:
\begin{align}
    \delta_k = \text{softmax} ( h W_\delta \hat{O}_{k} ) ,
\end{align}
where $\delta_k \in [0, 1]$ is the attention weight for the $k$-th selected sticker.
Then we apply the weighted-sum on all the user previously selected sticker representations:
\begin{align}
    r = \sum^{T_h}_k { \delta_k \hat{O}_{k} } .
\end{align}
Finally, we feed this preference representation $r$ into final gated fusion layer (Equation~\ref{eq:dynamic-fusion}).
Note that, the above two variations exclude the histories of dialogue contexts, and we employ these experiments to verify the effectiveness of incorporating histories of dialogue contexts.

\begin{table}[t]
    \centering
    \caption{RQ8: Performance of two variations user preference memory module.}
    \begin{tabular}{@{}lcc cc@{}}
        \toprule
        & MAP & $R_{10}@1$ & $R_{10}@2$ & $R_{10}@5$ \\
        \midrule
        SRS & 0.709  & 0.590 & 0.703 & 0.872 \\
        MostSelected & 0.545 & 0.419 & 0.490 & 0.679 \\
        AverageMem & 0.701 & 0.573 & 0.701 & 0.870 \\
        WeightedMem & 0.694 & 0.565 & 0.689 & 0.866 \\
        PESRS & \textbf{0.743}  & \textbf{0.632}\dubbelop & \textbf{0.740}\dubbelop & \textbf{0.897} \\
        \bottomrule
    \end{tabular}
    \label{tab:mem-analysis}
\end{table}

We conduct the experiments on these variations and compare with our proposed PESRS and SRS, as shown in Table~\ref{tab:mem-analysis}.
From this table, we can find that \textit{MostSelected} performs the worst among all the methods.
That demonstrates the necessity of exploring a learning-based method to leverage the user history data to recommend the proper sticker.
By comparing \textit{AverageMem} with the SRS which does not incorporate the user's history, we find that although \textit{AverageMem} and \textit{WeightedMem} leverages the user's history information, it can not take advantages from these data to boost the performance of sticker selection.
The reason is that these methods can not model the relationship between current dialog context and previous history data, thus it can not determine which history data may be helpful for the current context.

\subsection{Sticker Classification and Emotion Diversity}
\label{subsec:emoji-analysis}

\begin{figure}[h]
    \centering
    \includegraphics[scale=0.45]{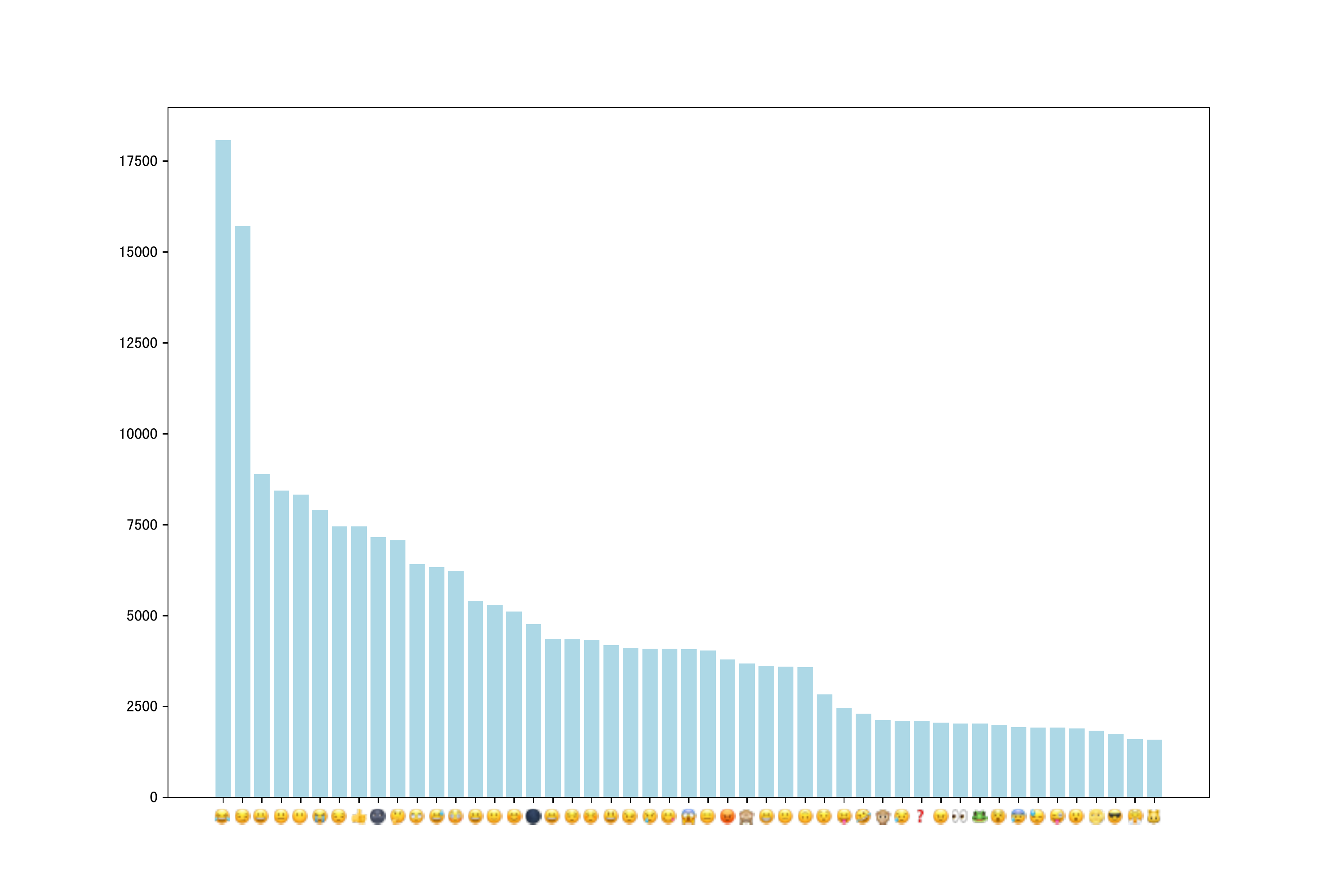}
    \caption{
        RQ9: Number of the used emoji labels over stickers in training dataset (top 50 emojis of 893 unique emojis in total). 
    }
    \label{fig:emoji-dist}
\end{figure}

Finally, we turn to \textbf{RQ9}.
In this dataset, the sticker authors give each sticker an emoji label that indicates the approximate emotion of the sticker.
However, this label is not a mandatory field when creating a sticker set in this online chatting platform.
Some authors use random emoji or one emoji label for all the stickers in the sticker set.
Thus, we cannot incorporate the emoji label and tackle the sticker selection task as an emoji classification task.
We randomly sample 20 sticker sets and employ human annotators to check whether the emoji label in sticker set is correct, and we find that there are 2 sticker set of them have wrong emoji labels for the stickers.
Since we introduce the auxiliary sticker classification (introduced in \S~\ref{subsec:sticker_encoder}) to help the model for accelerating convergence of the model training, we also report the sticker classification performance in this paper.
Note that, since the emoji label of the sticker may not be correct, therefore, the classification performance is \emph{not accurate}, the results are for reference only.
The results of the sticker classification are 65.74\%, 50.75\%, 47.02\%, 61.20\% for accuracy, F1, recall and precision, respectively.
These results indicate the sticker encoder can capture the semantic meanings of the sticker image.

To illustrate the diversity of the emotion expressed by the stickers, we use the emoji label as the indicator of the emotion and plot the distribution of the emoji label of stickers.
In Figure~\ref{fig:emoji-dist}, we only show the top 50 emoji labels used in all the sticker set in our training dataset and the total number of unique emoji label is 893.
From Figure~\ref{fig:emoji-dist}, we can find that there are many stickers with the emoji label \includegraphics[scale=0.08]{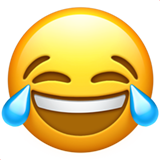} and \includegraphics[scale=0.08]{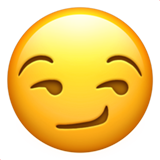}.
The reason is that some of the sticker authors assign \includegraphics[scale=0.08]{face-with-tears-of-joy_1f602.png} or \includegraphics[scale=0.08]{smirking-face_1f60f.png} as the emoji label to all the stickers in their sticker set, as we mentioned before (some authors use random emoji or one emoji label for all the stickers in the sticker set).
\section{Conclusion}\label{sec:conclusion}

In our previous work, we propose the task of multi-turn sticker response selection, which recommends an appropriate sticker based on multi-turn dialog context history without relying on external knowledge.
However, this method only focuses on measuring the matching degree between the dialog context and sticker image, which ignores the user preference of using stickers.
Hence, in this paper, we propose the \emph{Preference Enhanced Sticker Response Selector} (PESRS) to recommend an appropriate sticker to user based on multi-turn dialog context and sticker using history of user.
Specifically, PESRS first learns the representation of each utterance using a self-attention mechanism, and learns sticker representation by CNN.
Second, a deep interaction network is employed to fully model the dependency between the sticker and utterances.
The deep interaction network consists of a co-attention matrix that calculates the attention between each word in an utterance and each unit in a sticker representation.
Third, a bi-directional attention is used to obtain utterance-aware sticker representation and sticker-aware utterance representations.
Next, we retrieve the recent user sticker selections, and then propose a user preference modeling module which consists a position-aware history encoding network and a key-value based memory network to generate the user preference representation dynamically according to current dialog context.
Then, a fusion network models the short-term and long-term relationship between interaction results, and a gated fusion layer is applied to fuse the current dialog interaction results and user preference representation dynamically.
Finally, a fully-connected layer is applied to obtain the final sticker prediction using the output of gated fusion layer.
Our model outperforms state-of-the-art methods including our previous method SRS in all metrics and the experimental results also demonstrate the effectiveness of each module in our model.
In the near future, we aim to propose a personalized sticker response selection system. 
\section*{Acknowledgments}
We would like to thank the anonymous reviewers for their constructive comments. 
We would also like to thank Anna Hennig in Inception Institute of Artificial Intelligence for her help on this paper. 
This work was supported by the National Science Foundation of China (NSFC No. 61876196) and the National Key R\&D Program of China (2020AAA0105200).
Rui Yan is supported as a Young Fellow of Beijing Institute of Artificial Intelligence (BAAI).

\clearpage

\bibliographystyle{ACM-Reference-Format}
\bibliography{stickerPreference}


\begin{thebibliography}{95}


\ifx \showCODEN    \undefined \def \showCODEN     #1{\unskip}     \fi
\ifx \showDOI      \undefined \def \showDOI       #1{#1}\fi
\ifx \showISBNx    \undefined \def \showISBNx     #1{\unskip}     \fi
\ifx \showISBNxiii \undefined \def \showISBNxiii  #1{\unskip}     \fi
\ifx \showISSN     \undefined \def \showISSN      #1{\unskip}     \fi
\ifx \showLCCN     \undefined \def \showLCCN      #1{\unskip}     \fi
\ifx \shownote     \undefined \def \shownote      #1{#1}          \fi
\ifx \showarticletitle \undefined \def \showarticletitle #1{#1}   \fi
\ifx \showURL      \undefined \def \showURL       {\relax}        \fi
\providecommand\bibfield[2]{#2}
\providecommand\bibinfo[2]{#2}
\providecommand\natexlab[1]{#1}
\providecommand\showeprint[2][]{arXiv:#2}

\bibitem[\protect\citeauthoryear{Abadi, Barham, Chen, Chen, Davis, Dean, Devin,
  Ghemawat, Irving, Isard, et~al\mbox{.}}{Abadi et~al\mbox{.}}{2016}]%
        {abadi2016tensorflow}
\bibfield{author}{\bibinfo{person}{Mart{\'\i}n Abadi}, \bibinfo{person}{Paul
  Barham}, \bibinfo{person}{Jianmin Chen}, \bibinfo{person}{Zhifeng Chen},
  \bibinfo{person}{Andy Davis}, \bibinfo{person}{Jeffrey Dean},
  \bibinfo{person}{Matthieu Devin}, \bibinfo{person}{Sanjay Ghemawat},
  \bibinfo{person}{Geoffrey Irving}, \bibinfo{person}{Michael Isard},
  {et~al\mbox{.}}} \bibinfo{year}{2016}\natexlab{}.
\newblock \showarticletitle{Tensorflow: a system for large-scale machine
  learning.}. In \bibinfo{booktitle}{\emph{OSDI}}, Vol.~\bibinfo{volume}{16}.
  \bibinfo{pages}{265--283}.
\newblock


\bibitem[\protect\citeauthoryear{An, Wu, Wu, Zhang, Liu, and Xie}{An
  et~al\mbox{.}}{2019}]%
        {An2019Neural}
\bibfield{author}{\bibinfo{person}{Mingxiao An}, \bibinfo{person}{Fangzhao Wu},
  \bibinfo{person}{Chuhan Wu}, \bibinfo{person}{Kun Zhang},
  \bibinfo{person}{Zheng Liu}, {and} \bibinfo{person}{Xing Xie}.}
  \bibinfo{year}{2019}\natexlab{}.
\newblock \showarticletitle{Neural News Recommendation with Long- and
  Short-term User Representations}. In \bibinfo{booktitle}{\emph{ACL}}.
\newblock


\bibitem[\protect\citeauthoryear{Avanaki}{Avanaki}{2008}]%
        {avanaki2008exact}
\bibfield{author}{\bibinfo{person}{Alireza Avanaki}.}
  \bibinfo{year}{2008}\natexlab{}.
\newblock \showarticletitle{Exact histogram specification optimized for
  structural similarity}.
\newblock \bibinfo{journal}{\emph{arXiv preprint arXiv:0901.0065}}
  (\bibinfo{year}{2008}).
\newblock


\bibitem[\protect\citeauthoryear{Baeza-Yates, Ribeiro,
  et~al\mbox{.}}{Baeza-Yates et~al\mbox{.}}{2011}]%
        {baeza2011modern}
\bibfield{author}{\bibinfo{person}{Ricardo Baeza-Yates},
  \bibinfo{person}{Berthier de Ara{\'u}jo~Neto Ribeiro}, {et~al\mbox{.}}}
  \bibinfo{year}{2011}\natexlab{}.
\newblock \bibinfo{booktitle}{\emph{Modern information retrieval}}.
\newblock \bibinfo{publisher}{New York: ACM Press; Harlow, England:
  Addison-Wesley,}.
\newblock


\bibitem[\protect\citeauthoryear{Bahdanau, Cho, and Bengio}{Bahdanau
  et~al\mbox{.}}{2015}]%
        {bahdanau2014neural}
\bibfield{author}{\bibinfo{person}{Dzmitry Bahdanau},
  \bibinfo{person}{Kyunghyun Cho}, {and} \bibinfo{person}{Yoshua Bengio}.}
  \bibinfo{year}{2015}\natexlab{}.
\newblock \showarticletitle{Neural machine translation by jointly learning to
  align and translate}. In \bibinfo{booktitle}{\emph{ICLR}}.
\newblock


\bibitem[\protect\citeauthoryear{Barbieri, Ballesteros, Ronzano, and
  Saggion}{Barbieri et~al\mbox{.}}{2018}]%
        {barbieri2018multimodal}
\bibfield{author}{\bibinfo{person}{Francesco Barbieri}, \bibinfo{person}{Miguel
  Ballesteros}, \bibinfo{person}{Francesco Ronzano}, {and}
  \bibinfo{person}{Horacio Saggion}.} \bibinfo{year}{2018}\natexlab{}.
\newblock \showarticletitle{Multimodal Emoji Prediction}. In
  \bibinfo{booktitle}{\emph{NAACL}}. \bibinfo{publisher}{Association for
  Computational Linguistics}, \bibinfo{address}{New Orleans, Louisiana},
  \bibinfo{pages}{679--686}.
\newblock


\bibitem[\protect\citeauthoryear{Barbieri, Ballesteros, and Saggion}{Barbieri
  et~al\mbox{.}}{2017}]%
        {barbieri2017emojis}
\bibfield{author}{\bibinfo{person}{Francesco Barbieri}, \bibinfo{person}{Miguel
  Ballesteros}, {and} \bibinfo{person}{Horacio Saggion}.}
  \bibinfo{year}{2017}\natexlab{}.
\newblock \showarticletitle{Are Emojis Predictable?}. In
  \bibinfo{booktitle}{\emph{Proceedings of the 15th Conference of the
  {E}uropean Chapter of the Association for Computational Linguistics: Volume
  2, Short Papers}}. \bibinfo{publisher}{Association for Computational
  Linguistics}, \bibinfo{address}{Valencia, Spain}, \bibinfo{pages}{105--111}.
\newblock


\bibitem[\protect\citeauthoryear{Chan, Li, Yang, Chen, Hu, Zhao, and Yan}{Chan
  et~al\mbox{.}}{2019}]%
        {Chan2019Modeling}
\bibfield{author}{\bibinfo{person}{Zhangming Chan}, \bibinfo{person}{Juntao
  Li}, \bibinfo{person}{Xiaopeng Yang}, \bibinfo{person}{Xiuying Chen},
  \bibinfo{person}{Wenpeng Hu}, \bibinfo{person}{Dongyan Zhao}, {and}
  \bibinfo{person}{Rui Yan}.} \bibinfo{year}{2019}\natexlab{}.
\newblock \showarticletitle{Modeling Personalization in Continuous Space for
  Response Generation via Augmented Wasserstein Autoencoders}. In
  \bibinfo{booktitle}{\emph{EMNLP}}.
\newblock


\bibitem[\protect\citeauthoryear{Chen, Chan, Gao, Yu, Zhao, and Yan}{Chen
  et~al\mbox{.}}{2019}]%
        {Chen2019Learning}
\bibfield{author}{\bibinfo{person}{Xiuying Chen}, \bibinfo{person}{Zhangming
  Chan}, \bibinfo{person}{Shen Gao}, \bibinfo{person}{Meng-Hsuan Yu},
  \bibinfo{person}{Dongyan Zhao}, {and} \bibinfo{person}{Rui Yan}.}
  \bibinfo{year}{2019}\natexlab{}.
\newblock \showarticletitle{Learning towards Abstractive Timeline
  Summarization}. In \bibinfo{booktitle}{\emph{IJCAI}}.
\newblock


\bibitem[\protect\citeauthoryear{Chen, Xu, Zhang, Tang, Cao, Qin, and Zha}{Chen
  et~al\mbox{.}}{2018}]%
        {Chen2018Sequential}
\bibfield{author}{\bibinfo{person}{Xu Chen}, \bibinfo{person}{Hongteng Xu},
  \bibinfo{person}{Yongfeng Zhang}, \bibinfo{person}{Jiaxi Tang},
  \bibinfo{person}{Yixin Cao}, \bibinfo{person}{Zheng Qin}, {and}
  \bibinfo{person}{Hongyuan Zha}.} \bibinfo{year}{2018}\natexlab{}.
\newblock \showarticletitle{Sequential Recommendation with User Memory
  Networks}. In \bibinfo{booktitle}{\emph{WSDM}}.
\newblock


\bibitem[\protect\citeauthoryear{Chu, Vijayaraghavan, and Roy}{Chu
  et~al\mbox{.}}{2018}]%
        {Chu2018Learning}
\bibfield{author}{\bibinfo{person}{Eric Chu}, \bibinfo{person}{Prashanth
  Vijayaraghavan}, {and} \bibinfo{person}{Deb Roy}.}
  \bibinfo{year}{2018}\natexlab{}.
\newblock \showarticletitle{Learning Personas from Dialogue with Attentive
  Memory Networks}. In \bibinfo{booktitle}{\emph{EMNLP}}.
\newblock


\bibitem[\protect\citeauthoryear{Chung, Çaglar G{\"u}lçehre, Cho, and
  Bengio}{Chung et~al\mbox{.}}{2014}]%
        {Chung2014EmpiricalEO}
\bibfield{author}{\bibinfo{person}{Junyoung Chung}, \bibinfo{person}{Çaglar
  G{\"u}lçehre}, \bibinfo{person}{Kyunghyun Cho}, {and}
  \bibinfo{person}{Yoshua Bengio}.} \bibinfo{year}{2014}\natexlab{}.
\newblock \showarticletitle{Empirical Evaluation of Gated Recurrent Neural
  Networks on Sequence Modeling}. In \bibinfo{booktitle}{\emph{NIPS Workshop}}.
\newblock


\bibitem[\protect\citeauthoryear{{Das}, {Kottur}, {Gupta}, {Singh}, {Yadav},
  {Moura}, {Parikh}, and {Batra}}{{Das} et~al\mbox{.}}{2017}]%
        {das2017visual}
\bibfield{author}{\bibinfo{person}{A. {Das}}, \bibinfo{person}{S. {Kottur}},
  \bibinfo{person}{K. {Gupta}}, \bibinfo{person}{A. {Singh}},
  \bibinfo{person}{D. {Yadav}}, \bibinfo{person}{J.~M.~F. {Moura}},
  \bibinfo{person}{D. {Parikh}}, {and} \bibinfo{person}{D. {Batra}}.}
  \bibinfo{year}{2017}\natexlab{}.
\newblock \showarticletitle{Visual Dialog}. In
  \bibinfo{booktitle}{\emph{CVPR}}. \bibinfo{pages}{1080--1089}.
\newblock


\bibitem[\protect\citeauthoryear{de~Seta}{de~Seta}{2018}]%
        {Seta2018BiaoqingTC}
\bibfield{author}{\bibinfo{person}{Gabriele de Seta}.}
  \bibinfo{year}{2018}\natexlab{}.
\newblock \showarticletitle{Biaoqing: The circulation of emoticons, emoji,
  stickers, and custom images on Chinese digital media platforms}.
\newblock \bibinfo{journal}{\emph{First Monday}}  \bibinfo{volume}{23}
  (\bibinfo{year}{2018}).
\newblock


\bibitem[\protect\citeauthoryear{Ebesu, Shen, and Fang}{Ebesu
  et~al\mbox{.}}{2018}]%
        {Ebesu2018Collaborative}
\bibfield{author}{\bibinfo{person}{Travis Ebesu}, \bibinfo{person}{Bin Shen},
  {and} \bibinfo{person}{Yi Fang}.} \bibinfo{year}{2018}\natexlab{}.
\newblock \showarticletitle{Collaborative Memory Network for Recommendation
  Systems}. In \bibinfo{booktitle}{\emph{SIGIR}}.
\newblock


\bibitem[\protect\citeauthoryear{Fan, Lewis, and Dauphin}{Fan
  et~al\mbox{.}}{2018}]%
        {fan2018hierarchical}
\bibfield{author}{\bibinfo{person}{Angela Fan}, \bibinfo{person}{Mike Lewis},
  {and} \bibinfo{person}{Yann Dauphin}.} \bibinfo{year}{2018}\natexlab{}.
\newblock \showarticletitle{Hierarchical Neural Story Generation}. In
  \bibinfo{booktitle}{\emph{Proceedings of the 56th Annual Meeting of the
  Association for Computational Linguistics (Volume 1: Long Papers)}}.
  \bibinfo{publisher}{Association for Computational Linguistics},
  \bibinfo{address}{Melbourne, Australia}, \bibinfo{pages}{889--898}.
\newblock


\bibitem[\protect\citeauthoryear{Feng, Tao, Wu, Feng, Zhao, and Yan}{Feng
  et~al\mbox{.}}{2019}]%
        {Feng2019Learning}
\bibfield{author}{\bibinfo{person}{Jiazhan Feng}, \bibinfo{person}{Chongyang
  Tao}, \bibinfo{person}{Wei Wu}, \bibinfo{person}{Yansong Feng},
  \bibinfo{person}{Dongyan Zhao}, {and} \bibinfo{person}{Rui Yan}.}
  \bibinfo{year}{2019}\natexlab{}.
\newblock \showarticletitle{Learning a Matching Model with Co-teaching for
  Multi-turn Response Selection in Retrieval-based Dialogue Systems}. In
  \bibinfo{booktitle}{\emph{Proceedings of the 57th Annual Meeting of the
  Association for Computational Linguistics}}. \bibinfo{publisher}{Association
  for Computational Linguistics}, \bibinfo{address}{Florence, Italy},
  \bibinfo{pages}{3805--3815}.
\newblock


\bibitem[\protect\citeauthoryear{Gao, Ge, Chen, and Nevatia}{Gao
  et~al\mbox{.}}{2018}]%
        {Gao2018MotionAppearance}
\bibfield{author}{\bibinfo{person}{Jiyang Gao}, \bibinfo{person}{Runzhou Ge},
  \bibinfo{person}{Kan Chen}, {and} \bibinfo{person}{Ram Nevatia}.}
  \bibinfo{year}{2018}\natexlab{}.
\newblock \showarticletitle{Motion-Appearance Co-Memory Networks for Video
  Question Answering}. In \bibinfo{booktitle}{\emph{CVPR}}.
\newblock


\bibitem[\protect\citeauthoryear{Gao, You, Zhang, Wang, and Li}{Gao
  et~al\mbox{.}}{2019d}]%
        {Gao2019Multi}
\bibfield{author}{\bibinfo{person}{Peng Gao}, \bibinfo{person}{Haoxuan You},
  \bibinfo{person}{Zhanpeng Zhang}, \bibinfo{person}{Xiaogang Wang}, {and}
  \bibinfo{person}{Hongsheng Li}.} \bibinfo{year}{2019}\natexlab{d}.
\newblock \showarticletitle{Multi-Modality Latent Interaction Network for
  Visual Question Answering}. In \bibinfo{booktitle}{\emph{ICCV}}.
\newblock


\bibitem[\protect\citeauthoryear{Gao, Chen, Li, Chan, Zhao, and Yan}{Gao
  et~al\mbox{.}}{2019a}]%
        {Gao2019How}
\bibfield{author}{\bibinfo{person}{Shen Gao}, \bibinfo{person}{Xiuying Chen},
  \bibinfo{person}{Piji Li}, \bibinfo{person}{Zhangming Chan},
  \bibinfo{person}{Dongyan Zhao}, {and} \bibinfo{person}{Rui Yan}.}
  \bibinfo{year}{2019}\natexlab{a}.
\newblock \showarticletitle{How to Write Summaries with Patterns? Learning
  towards Abstractive Summarization through Prototype Editing}. In
  \bibinfo{booktitle}{\emph{Proceedings of the 2019 Conference on Empirical
  Methods in Natural Language Processing and the 9th International Joint
  Conference on Natural Language Processing (EMNLP-IJCNLP)}}.
  \bibinfo{publisher}{Association for Computational Linguistics},
  \bibinfo{address}{Hong Kong, China}, \bibinfo{pages}{3741--3751}.
\newblock


\bibitem[\protect\citeauthoryear{Gao, Chen, Li, Ren, Bing, Zhao, and Yan}{Gao
  et~al\mbox{.}}{2019b}]%
        {Gao2019Abstractive}
\bibfield{author}{\bibinfo{person}{Shen Gao}, \bibinfo{person}{Xiuying Chen},
  \bibinfo{person}{Piji Li}, \bibinfo{person}{Zhaochun Ren},
  \bibinfo{person}{Lidong Bing}, \bibinfo{person}{Dongyan Zhao}, {and}
  \bibinfo{person}{Rui Yan}.} \bibinfo{year}{2019}\natexlab{b}.
\newblock \showarticletitle{Abstractive Text Summarization by Incorporating
  Reader Comments}. In \bibinfo{booktitle}{\emph{AAAI}}.
  \bibinfo{pages}{6399--6406}.
\newblock


\bibitem[\protect\citeauthoryear{Gao, Chen, Liu, Liu, Zhao, and Yan}{Gao
  et~al\mbox{.}}{2020a}]%
        {gao2020sticker}
\bibfield{author}{\bibinfo{person}{Shen Gao}, \bibinfo{person}{Xiuying Chen},
  \bibinfo{person}{Chang Liu}, \bibinfo{person}{Li Liu},
  \bibinfo{person}{Dongyan Zhao}, {and} \bibinfo{person}{Rui Yan}.}
  \bibinfo{year}{2020}\natexlab{a}.
\newblock \showarticletitle{Learning to Respond with Stickers: A Framework of
  Unifying Multi-Modality in Multi-Turn Dialog}. In
  \bibinfo{booktitle}{\emph{The World Wide Web Conference (WWW '20)}}.
  \bibinfo{publisher}{Association for Computing Machinery},
  \bibinfo{address}{New York, NY, USA}.
\newblock


\bibitem[\protect\citeauthoryear{Gao, Chen, Ren, Zhao, and Yan}{Gao
  et~al\mbox{.}}{2020b}]%
        {Gao2020From}
\bibfield{author}{\bibinfo{person}{Shen Gao}, \bibinfo{person}{Xiuying Chen},
  \bibinfo{person}{Zhaochun Ren}, \bibinfo{person}{Dongyan Zhao}, {and}
  \bibinfo{person}{Rui Yan}.} \bibinfo{year}{2020}\natexlab{b}.
\newblock \showarticletitle{From Standard Summarization to New Tasks and
  Beyond: Summarization with Manifold Information}. In
  \bibinfo{booktitle}{\emph{IJCAI}}.
\newblock


\bibitem[\protect\citeauthoryear{Gao, Ren, Zhao, Zhao, Yin, and Yan}{Gao
  et~al\mbox{.}}{2019c}]%
        {Gao2019Product}
\bibfield{author}{\bibinfo{person}{Shen Gao}, \bibinfo{person}{Zhaochun Ren},
  \bibinfo{person}{Yihong Zhao}, \bibinfo{person}{Dongyan Zhao},
  \bibinfo{person}{Dawei Yin}, {and} \bibinfo{person}{Rui Yan}.}
  \bibinfo{year}{2019}\natexlab{c}.
\newblock \showarticletitle{Product-Aware Answer Generation in E-Commerce
  Question-Answering}. In \bibinfo{booktitle}{\emph{Proceedings of the Twelfth
  ACM International Conference on Web Search and Data Mining}} (Melbourne VIC,
  Australia) \emph{(\bibinfo{series}{WSDM '19})}.
  \bibinfo{publisher}{Association for Computing Machinery},
  \bibinfo{address}{New York, NY, USA}, \bibinfo{pages}{429--437}.
\newblock
\showISBNx{9781450359405}


\bibitem[\protect\citeauthoryear{Ge and Herring}{Ge and Herring}{2018}]%
        {Ge2018CommunicativeFO}
\bibfield{author}{\bibinfo{person}{Jing Ge} {and} \bibinfo{person}{Susan~C.
  Herring}.} \bibinfo{year}{2018}\natexlab{}.
\newblock \showarticletitle{Communicative functions of emoji sequences on Sina
  Weibo}.
\newblock \bibinfo{journal}{\emph{First Monday}}  \bibinfo{volume}{23}
  (\bibinfo{year}{2018}).
\newblock


\bibitem[\protect\citeauthoryear{Goyal, Wang, and Deng}{Goyal
  et~al\mbox{.}}{2018}]%
        {Goyal2018Think}
\bibfield{author}{\bibinfo{person}{Ankit Goyal}, \bibinfo{person}{Jian Wang},
  {and} \bibinfo{person}{Jia Deng}.} \bibinfo{year}{2018}\natexlab{}.
\newblock \showarticletitle{Think Visually: Question Answering through Virtual
  Imagery}. In \bibinfo{booktitle}{\emph{Proceedings of the 56th Annual Meeting
  of the Association for Computational Linguistics (Volume 1: Long Papers)}}.
  \bibinfo{publisher}{Association for Computational Linguistics},
  \bibinfo{address}{Melbourne, Australia}, \bibinfo{pages}{2598--2608}.
\newblock


\bibitem[\protect\citeauthoryear{Goyal, Khot, Summers-Stay, Batra, and
  Parikh}{Goyal et~al\mbox{.}}{2017}]%
        {goyal2017making}
\bibfield{author}{\bibinfo{person}{Yash Goyal}, \bibinfo{person}{Tejas Khot},
  \bibinfo{person}{Douglas Summers-Stay}, \bibinfo{person}{Dhruv Batra}, {and}
  \bibinfo{person}{Devi Parikh}.} \bibinfo{year}{2017}\natexlab{}.
\newblock \showarticletitle{Making the V in VQA matter: Elevating the role of
  image understanding in Visual Question Answering}. In
  \bibinfo{booktitle}{\emph{CVPR}}. \bibinfo{pages}{6904--6913}.
\newblock


\bibitem[\protect\citeauthoryear{Guibon, Ochs, and Bellot}{Guibon
  et~al\mbox{.}}{2018}]%
        {Guibon2018EmojiRI}
\bibfield{author}{\bibinfo{person}{Ga{\"e}l Guibon}, \bibinfo{person}{Magalie
  Ochs}, {and} \bibinfo{person}{Patrice Bellot}.}
  \bibinfo{year}{2018}\natexlab{}.
\newblock \showarticletitle{Emoji recommendation in private instant messages}.
\newblock \bibinfo{journal}{\emph{Proceedings of the 33rd Annual ACM Symposium
  on Applied Computing}} (\bibinfo{year}{2018}).
\newblock


\bibitem[\protect\citeauthoryear{Guo, Wang, and Wang}{Guo
  et~al\mbox{.}}{2019}]%
        {Guo2019Dual}
\bibfield{author}{\bibinfo{person}{Dan Guo}, \bibinfo{person}{Hui Wang}, {and}
  \bibinfo{person}{Meng Wang}.} \bibinfo{year}{2019}\natexlab{}.
\newblock \showarticletitle{Dual Visual Attention Network for Visual Dialog}.
  In \bibinfo{booktitle}{\emph{IJCAI}}.
\newblock


\bibitem[\protect\citeauthoryear{{Guo}, {Xu}, and {Tao}}{{Guo}
  et~al\mbox{.}}{2019}]%
        {guo2019image}
\bibfield{author}{\bibinfo{person}{D. {Guo}}, \bibinfo{person}{C. {Xu}}, {and}
  \bibinfo{person}{D. {Tao}}.} \bibinfo{year}{2019}\natexlab{}.
\newblock \showarticletitle{Image-Question-Answer Synergistic Network for
  Visual Dialog}. In \bibinfo{booktitle}{\emph{CVPR '19}}.
  \bibinfo{pages}{10426--10435}.
\newblock


\bibitem[\protect\citeauthoryear{He, Zhang, Ren, and Sun}{He
  et~al\mbox{.}}{2015}]%
        {he2015delving}
\bibfield{author}{\bibinfo{person}{Kaiming He}, \bibinfo{person}{Xiangyu
  Zhang}, \bibinfo{person}{Shaoqing Ren}, {and} \bibinfo{person}{Jian Sun}.}
  \bibinfo{year}{2015}\natexlab{}.
\newblock \showarticletitle{Delving deep into rectifiers: Surpassing
  human-level performance on imagenet classification}. In
  \bibinfo{booktitle}{\emph{ICCV}}. \bibinfo{pages}{1026--1034}.
\newblock


\bibitem[\protect\citeauthoryear{He, Zhang, Ren, and Sun}{He
  et~al\mbox{.}}{2016}]%
        {he2016deep}
\bibfield{author}{\bibinfo{person}{Kaiming He}, \bibinfo{person}{Xiangyu
  Zhang}, \bibinfo{person}{Shaoqing Ren}, {and} \bibinfo{person}{Jian Sun}.}
  \bibinfo{year}{2016}\natexlab{}.
\newblock \showarticletitle{Deep residual learning for image recognition}. In
  \bibinfo{booktitle}{\emph{CVPR}}. \bibinfo{pages}{770--778}.
\newblock


\bibitem[\protect\citeauthoryear{Herring and Dainas}{Herring and
  Dainas}{2017}]%
        {Herring2017NicePC}
\bibfield{author}{\bibinfo{person}{Susan~C. Herring} {and}
  \bibinfo{person}{Ashley Dainas}.} \bibinfo{year}{2017}\natexlab{}.
\newblock \showarticletitle{"Nice Picture Comment!" Graphicons in Facebook
  Comment Threads}. In \bibinfo{booktitle}{\emph{HICSS}}.
\newblock


\bibitem[\protect\citeauthoryear{Hochreiter and Schmidhuber}{Hochreiter and
  Schmidhuber}{1997}]%
        {hochreiter1997long}
\bibfield{author}{\bibinfo{person}{Sepp Hochreiter} {and}
  \bibinfo{person}{J{\"u}rgen Schmidhuber}.} \bibinfo{year}{1997}\natexlab{}.
\newblock \showarticletitle{Long short-term memory}.
\newblock \bibinfo{journal}{\emph{Neural computation}} \bibinfo{volume}{9},
  \bibinfo{number}{8} (\bibinfo{year}{1997}), \bibinfo{pages}{1735--1780}.
\newblock


\bibitem[\protect\citeauthoryear{Huang, Ren, Zhao, He, Wen, and Dong}{Huang
  et~al\mbox{.}}{2019b}]%
        {Huang2019TaxonomyAware}
\bibfield{author}{\bibinfo{person}{Jin Huang}, \bibinfo{person}{Zhaochun Ren},
  \bibinfo{person}{Wayne~Xin Zhao}, \bibinfo{person}{Gaole He},
  \bibinfo{person}{Ji-Rong Wen}, {and} \bibinfo{person}{Daxiang Dong}.}
  \bibinfo{year}{2019}\natexlab{b}.
\newblock \showarticletitle{Taxonomy-Aware Multi-Hop Reasoning Networks for
  Sequential Recommendation}. In \bibinfo{booktitle}{\emph{WSDM}}.
\newblock


\bibitem[\protect\citeauthoryear{Huang, Fang, Qian, Sang, Li, and Xu}{Huang
  et~al\mbox{.}}{2019a}]%
        {Huang2019Explainable}
\bibfield{author}{\bibinfo{person}{Xiaowen Huang}, \bibinfo{person}{Quan Fang},
  \bibinfo{person}{Shengsheng Qian}, \bibinfo{person}{Jitao Sang},
  \bibinfo{person}{Yan Li}, {and} \bibinfo{person}{Changsheng Xu}.}
  \bibinfo{year}{2019}\natexlab{a}.
\newblock \showarticletitle{Explainable Interaction-driven User Modeling over
  Knowledge Graph for Sequential Recommendation}. In
  \bibinfo{booktitle}{\emph{MM}}.
\newblock


\bibitem[\protect\citeauthoryear{Jain, Lazebnik, and Schwing}{Jain
  et~al\mbox{.}}{2018}]%
        {jain2018two}
\bibfield{author}{\bibinfo{person}{Unnat Jain}, \bibinfo{person}{Svetlana
  Lazebnik}, {and} \bibinfo{person}{Alexander~G Schwing}.}
  \bibinfo{year}{2018}\natexlab{}.
\newblock \showarticletitle{Two can play this game: visual dialog with
  discriminative question generation and answering}. In
  \bibinfo{booktitle}{\emph{ICCV}}. \bibinfo{pages}{5754--5763}.
\newblock


\bibitem[\protect\citeauthoryear{Jing, Xie, and Xing}{Jing
  et~al\mbox{.}}{2018}]%
        {Jing2018OnTA}
\bibfield{author}{\bibinfo{person}{Baoyu Jing}, \bibinfo{person}{Pengtao Xie},
  {and} \bibinfo{person}{Eric Xing}.} \bibinfo{year}{2018}\natexlab{}.
\newblock \showarticletitle{On the Automatic Generation of Medical Imaging
  Reports}. In \bibinfo{booktitle}{\emph{ACL '18}}.
  \bibinfo{publisher}{Association for Computational Linguistics},
  \bibinfo{address}{Melbourne, Australia}, \bibinfo{pages}{2577--2586}.
\newblock


\bibitem[\protect\citeauthoryear{Kim, Kim, and Kim}{Kim et~al\mbox{.}}{2019}]%
        {Kim2019Abstractive}
\bibfield{author}{\bibinfo{person}{Byeongchang Kim}, \bibinfo{person}{Hyunwoo
  Kim}, {and} \bibinfo{person}{Gunhee Kim}.} \bibinfo{year}{2019}\natexlab{}.
\newblock \showarticletitle{Abstractive Summarization of Reddit Posts with
  Multi-level Memory Networks}. In \bibinfo{booktitle}{\emph{NAACL}}.
\newblock


\bibitem[\protect\citeauthoryear{Kingma and Ba}{Kingma and Ba}{2015}]%
        {Kingma2015AdamAM}
\bibfield{author}{\bibinfo{person}{Diederik~P. Kingma} {and}
  \bibinfo{person}{Jimmy Ba}.} \bibinfo{year}{2015}\natexlab{}.
\newblock \showarticletitle{Adam: A Method for Stochastic Optimization}. In
  \bibinfo{booktitle}{\emph{ICLR}}, Vol.~\bibinfo{volume}{abs/1412.6980}.
\newblock


\bibitem[\protect\citeauthoryear{Kumar, Irsoy, Ondruska, Iyyer, Bradbury,
  Gulrajani, Zhong, Paulus, and Socher}{Kumar et~al\mbox{.}}{2016}]%
        {Kumar2016AskMA}
\bibfield{author}{\bibinfo{person}{Ankit Kumar}, \bibinfo{person}{Ozan Irsoy},
  \bibinfo{person}{Peter Ondruska}, \bibinfo{person}{Mohit Iyyer},
  \bibinfo{person}{James Bradbury}, \bibinfo{person}{Ishaan Gulrajani},
  \bibinfo{person}{Victor Zhong}, \bibinfo{person}{Romain Paulus}, {and}
  \bibinfo{person}{Richard Socher}.} \bibinfo{year}{2016}\natexlab{}.
\newblock \showarticletitle{Ask Me Anything: Dynamic Memory Networks for
  Natural Language Processing}.
\newblock \bibinfo{journal}{\emph{ArXiv}}  \bibinfo{volume}{abs/1506.07285}
  (\bibinfo{year}{2016}).
\newblock


\bibitem[\protect\citeauthoryear{Laddha, Hanoosh, and Mukherjee}{Laddha
  et~al\mbox{.}}{2019}]%
        {laddha2019understanding}
\bibfield{author}{\bibinfo{person}{Abhishek Laddha}, \bibinfo{person}{Mohamed
  Hanoosh}, {and} \bibinfo{person}{Debdoot Mukherjee}.}
  \bibinfo{year}{2019}\natexlab{}.
\newblock \showarticletitle{Understanding Chat Messages for Sticker
  Recommendation in Hike Messenger}.
\newblock \bibinfo{journal}{\emph{arXiv:1902.02704}} (\bibinfo{year}{2019}).
\newblock


\bibitem[\protect\citeauthoryear{Lei, Ji, and Li}{Lei et~al\mbox{.}}{2019}]%
        {Lei2019TiSSA}
\bibfield{author}{\bibinfo{person}{Chenyi Lei}, \bibinfo{person}{Shouling Ji},
  {and} \bibinfo{person}{Zhao Li}.} \bibinfo{year}{2019}\natexlab{}.
\newblock \showarticletitle{TiSSA: A Time Slice Self-Attention Approach for
  Modeling Sequential User Behaviors}. In \bibinfo{booktitle}{\emph{WWW}}.
\newblock


\bibitem[\protect\citeauthoryear{Lei~Ba, Kiros, and Hinton}{Lei~Ba
  et~al\mbox{.}}{2016}]%
        {lei2016layer}
\bibfield{author}{\bibinfo{person}{Jimmy Lei~Ba}, \bibinfo{person}{Jamie~Ryan
  Kiros}, {and} \bibinfo{person}{Geoffrey~E Hinton}.}
  \bibinfo{year}{2016}\natexlab{}.
\newblock \showarticletitle{Layer normalization}.
\newblock \bibinfo{journal}{\emph{arXiv preprint arXiv:1607.06450}}
  (\bibinfo{year}{2016}).
\newblock


\bibitem[\protect\citeauthoryear{Li, Qiu, Tang, Chen, Zhao, and Yan}{Li
  et~al\mbox{.}}{2019a}]%
        {Li2019Insufficient}
\bibfield{author}{\bibinfo{person}{Juntao Li}, \bibinfo{person}{Lisong Qiu},
  \bibinfo{person}{Bo Tang}, \bibinfo{person}{Dongmin Chen},
  \bibinfo{person}{Dongyan Zhao}, {and} \bibinfo{person}{Rui Yan}.}
  \bibinfo{year}{2019}\natexlab{a}.
\newblock \showarticletitle{Insufficient Data Can Also Rock! Learning to
  Converse Using Smaller Data with Augmentation}. In
  \bibinfo{booktitle}{\emph{AAAI}}.
\newblock


\bibitem[\protect\citeauthoryear{Li, Song, Gao, Liu, Huang, He, and Gan}{Li
  et~al\mbox{.}}{2019b}]%
        {li2019beyond}
\bibfield{author}{\bibinfo{person}{Xiangpeng Li}, \bibinfo{person}{Jingkuan
  Song}, \bibinfo{person}{Lianli Gao}, \bibinfo{person}{Xianglong Liu},
  \bibinfo{person}{Wenbing Huang}, \bibinfo{person}{Xiangnan He}, {and}
  \bibinfo{person}{Chuang Gan}.} \bibinfo{year}{2019}\natexlab{b}.
\newblock \showarticletitle{Beyond RNNs: Positional Self-Attention with
  Co-Attention for Video Question Answering}. In
  \bibinfo{booktitle}{\emph{AAAI}}.
\newblock


\bibitem[\protect\citeauthoryear{Li, Duan, Zhou, Chu, Ouyang, Wang, and
  Zhou}{Li et~al\mbox{.}}{2018}]%
        {Li2018Visual}
\bibfield{author}{\bibinfo{person}{Yikang Li}, \bibinfo{person}{Nan Duan},
  \bibinfo{person}{Bolei Zhou}, \bibinfo{person}{Xiao Chu},
  \bibinfo{person}{Wanli Ouyang}, \bibinfo{person}{Xiaogang Wang}, {and}
  \bibinfo{person}{Ming Zhou}.} \bibinfo{year}{2018}\natexlab{}.
\newblock \showarticletitle{Visual Question Generation as Dual Task of Visual
  Question Answering}. In \bibinfo{booktitle}{\emph{CVPR}}.
\newblock


\bibitem[\protect\citeauthoryear{Lu, Kannan, Yang, Parikh, and Batra}{Lu
  et~al\mbox{.}}{2017}]%
        {lu2017best}
\bibfield{author}{\bibinfo{person}{Jiasen Lu}, \bibinfo{person}{Anitha Kannan},
  \bibinfo{person}{Jianwei Yang}, \bibinfo{person}{Devi Parikh}, {and}
  \bibinfo{person}{Dhruv Batra}.} \bibinfo{year}{2017}\natexlab{}.
\newblock \showarticletitle{Best of both worlds: Transferring knowledge from
  discriminative learning to a generative visual dialog model}. In
  \bibinfo{booktitle}{\emph{NIPS}}. \bibinfo{pages}{314--324}.
\newblock


\bibitem[\protect\citeauthoryear{Ma, Shen, Dick, Wu, Wang, van~den Hengel, and
  Reid}{Ma et~al\mbox{.}}{2018}]%
        {Ma2018Visual}
\bibfield{author}{\bibinfo{person}{Chao Ma}, \bibinfo{person}{Chunhua Shen},
  \bibinfo{person}{Anthony Dick}, \bibinfo{person}{Qi Wu},
  \bibinfo{person}{Peng Wang}, \bibinfo{person}{Anton van~den Hengel}, {and}
  \bibinfo{person}{Ian Reid}.} \bibinfo{year}{2018}\natexlab{}.
\newblock \showarticletitle{Visual Question Answering With Memory-Augmented
  Networks}. In \bibinfo{booktitle}{\emph{CVPR}}.
\newblock


\bibitem[\protect\citeauthoryear{Malinowski, Rohrbach, and Fritz}{Malinowski
  et~al\mbox{.}}{2015}]%
        {malinowski2015ask}
\bibfield{author}{\bibinfo{person}{Mateusz Malinowski}, \bibinfo{person}{Marcus
  Rohrbach}, {and} \bibinfo{person}{Mario Fritz}.}
  \bibinfo{year}{2015}\natexlab{}.
\newblock \showarticletitle{Ask your neurons: A neural-based approach to
  answering questions about images}. In \bibinfo{booktitle}{\emph{ICCV}}.
  \bibinfo{pages}{1--9}.
\newblock


\bibitem[\protect\citeauthoryear{Maruf and Haffari}{Maruf and Haffari}{2018}]%
        {Maruf2018Document}
\bibfield{author}{\bibinfo{person}{Sameen Maruf} {and}
  \bibinfo{person}{Gholamreza Haffari}.} \bibinfo{year}{2018}\natexlab{}.
\newblock \showarticletitle{Document Context Neural Machine Translation with
  Memory Networks}. In \bibinfo{booktitle}{\emph{ACL}}.
\newblock


\bibitem[\protect\citeauthoryear{Miller, Fisch, Dodge, Karimi, Bordes, and
  Weston}{Miller et~al\mbox{.}}{2016}]%
        {Miller2016KeyValueMN}
\bibfield{author}{\bibinfo{person}{Alexander~H. Miller}, \bibinfo{person}{Adam
  Fisch}, \bibinfo{person}{Jesse Dodge}, \bibinfo{person}{Amir-Hossein Karimi},
  \bibinfo{person}{Antoine Bordes}, {and} \bibinfo{person}{Jason Weston}.}
  \bibinfo{year}{2016}\natexlab{}.
\newblock \showarticletitle{Key-Value Memory Networks for Directly Reading
  Documents}.
\newblock \bibinfo{journal}{\emph{ArXiv}}  \bibinfo{volume}{abs/1606.03126}
  (\bibinfo{year}{2016}).
\newblock


\bibitem[\protect\citeauthoryear{Murahari, Chattopadhyay, Batra, Parikh, and
  Das}{Murahari et~al\mbox{.}}{2019}]%
        {Murahari2019Improving}
\bibfield{author}{\bibinfo{person}{Vishvak Murahari},
  \bibinfo{person}{Prithvijit Chattopadhyay}, \bibinfo{person}{Dhruv Batra},
  \bibinfo{person}{Devi Parikh}, {and} \bibinfo{person}{Abhishek Das}.}
  \bibinfo{year}{2019}\natexlab{}.
\newblock \showarticletitle{Improving Generative Visual Dialog by Answering
  Diverse Questions}. In \bibinfo{booktitle}{\emph{EMNLP}}.
\newblock


\bibitem[\protect\citeauthoryear{Nair and Hinton}{Nair and Hinton}{2010}]%
        {Nair2010RectifiedLU}
\bibfield{author}{\bibinfo{person}{Vinod Nair} {and}
  \bibinfo{person}{Geoffrey~E. Hinton}.} \bibinfo{year}{2010}\natexlab{}.
\newblock \showarticletitle{Rectified Linear Units Improve Restricted Boltzmann
  Machines}. In \bibinfo{booktitle}{\emph{ICML}}.
\newblock


\bibitem[\protect\citeauthoryear{{Noh}, {Kim}, {Mun}, and {Han}}{{Noh}
  et~al\mbox{.}}{2019}]%
        {Noh2019Transfer}
\bibfield{author}{\bibinfo{person}{H. {Noh}}, \bibinfo{person}{T. {Kim}},
  \bibinfo{person}{J. {Mun}}, {and} \bibinfo{person}{B. {Han}}.}
  \bibinfo{year}{2019}\natexlab{}.
\newblock \showarticletitle{Transfer Learning via Unsupervised Task Discovery
  for Visual Question Answering}. In \bibinfo{booktitle}{\emph{2019 IEEE/CVF
  Conference on Computer Vision and Pattern Recognition (CVPR)}}.
  \bibinfo{pages}{8377--8386}.
\newblock
\showISSN{1063-6919}


\bibitem[\protect\citeauthoryear{Pavez, Allende, and Allende-Cid}{Pavez
  et~al\mbox{.}}{2018}]%
        {Pavez2018Working}
\bibfield{author}{\bibinfo{person}{Juan Pavez}, \bibinfo{person}{Héctor
  Allende}, {and} \bibinfo{person}{Héctor Allende-Cid}.}
  \bibinfo{year}{2018}\natexlab{}.
\newblock \showarticletitle{Working Memory Networks: Augmenting Memory Networks
  with a Relational Reasoning Module}. In \bibinfo{booktitle}{\emph{ACL}}.
\newblock


\bibitem[\protect\citeauthoryear{Pi, Bian, Zhou, Zhu, and Gai}{Pi
  et~al\mbox{.}}{2019}]%
        {Pi2019Practice}
\bibfield{author}{\bibinfo{person}{Qi Pi}, \bibinfo{person}{Weijie Bian},
  \bibinfo{person}{Guorui Zhou}, \bibinfo{person}{Xiaoqiang Zhu}, {and}
  \bibinfo{person}{Kun Gai}.} \bibinfo{year}{2019}\natexlab{}.
\newblock \showarticletitle{Practice on Long Sequential User Behavior Modeling
  for Click-Through Rate Prediction}. In \bibinfo{booktitle}{\emph{KDD}}.
\newblock


\bibitem[\protect\citeauthoryear{Ren, Qin, Fang, Zhang, Zheng, Bian, Zhou, Xu,
  Yu, Zhu, and Gai}{Ren et~al\mbox{.}}{2019b}]%
        {Ren2019Lifelong}
\bibfield{author}{\bibinfo{person}{Kan Ren}, \bibinfo{person}{Jiarui Qin},
  \bibinfo{person}{Yuchen Fang}, \bibinfo{person}{Weinan Zhang},
  \bibinfo{person}{Lei Zheng}, \bibinfo{person}{Weijie Bian},
  \bibinfo{person}{Guorui Zhou}, \bibinfo{person}{Jian Xu},
  \bibinfo{person}{Yong Yu}, \bibinfo{person}{Xiaoqiang Zhu}, {and}
  \bibinfo{person}{Kun Gai}.} \bibinfo{year}{2019}\natexlab{b}.
\newblock \showarticletitle{Lifelong Sequential Modeling with Personalized
  Memorization for User Response Prediction}. In
  \bibinfo{booktitle}{\emph{SIGIR}}.
\newblock


\bibitem[\protect\citeauthoryear{Ren, Chen, Li, Ren, Ma, and de~Rijke}{Ren
  et~al\mbox{.}}{2019a}]%
        {Ren2019RepeatNet}
\bibfield{author}{\bibinfo{person}{Pengjie Ren}, \bibinfo{person}{Zhumin Chen},
  \bibinfo{person}{Jing Li}, \bibinfo{person}{Zhaochun Ren},
  \bibinfo{person}{Jun Ma}, {and} \bibinfo{person}{Maarten de Rijke}.}
  \bibinfo{year}{2019}\natexlab{a}.
\newblock \showarticletitle{RepeatNet: A Repeat Aware Neural Recommendation
  Machine for Session-Based Recommendation}. In
  \bibinfo{booktitle}{\emph{AAAI}}.
\newblock


\bibitem[\protect\citeauthoryear{{Sha}, {Hu}, and {Chao}}{{Sha}
  et~al\mbox{.}}{2018}]%
        {Chao2018Cross}
\bibfield{author}{\bibinfo{person}{F. {Sha}}, \bibinfo{person}{H. {Hu}}, {and}
  \bibinfo{person}{W. {Chao}}.} \bibinfo{year}{2018}\natexlab{}.
\newblock \showarticletitle{Cross-Dataset Adaptation for Visual Question
  Answering}. In \bibinfo{booktitle}{\emph{2018 IEEE/CVF Conference on Computer
  Vision and Pattern Recognition}}. \bibinfo{pages}{5716--5725}.
\newblock
\showISSN{1063-6919}


\bibitem[\protect\citeauthoryear{Su, Zhu, Dong, Cai, Chen, and Li}{Su
  et~al\mbox{.}}{2018}]%
        {Su2018Learning}
\bibfield{author}{\bibinfo{person}{Zhou Su}, \bibinfo{person}{Chen Zhu},
  \bibinfo{person}{Yinpeng Dong}, \bibinfo{person}{Dongqi Cai},
  \bibinfo{person}{Yurong Chen}, {and} \bibinfo{person}{Jianguo Li}.}
  \bibinfo{year}{2018}\natexlab{}.
\newblock \showarticletitle{Learning Visual Knowledge Memory Networks for
  Visual Question Answering}. In \bibinfo{booktitle}{\emph{CVPR}}.
\newblock


\bibitem[\protect\citeauthoryear{Sukhbaatar, Szlam, Weston, and
  Fergus}{Sukhbaatar et~al\mbox{.}}{2015}]%
        {Sukhbaatar2015EndToEndMN}
\bibfield{author}{\bibinfo{person}{Sainbayar Sukhbaatar},
  \bibinfo{person}{Arthur Szlam}, \bibinfo{person}{Jason Weston}, {and}
  \bibinfo{person}{Rob Fergus}.} \bibinfo{year}{2015}\natexlab{}.
\newblock \showarticletitle{End-To-End Memory Networks}. In
  \bibinfo{booktitle}{\emph{NIPS}}.
\newblock


\bibitem[\protect\citeauthoryear{Szegedy, Vanhoucke, Ioffe, Shlens, and
  Wojna}{Szegedy et~al\mbox{.}}{2016}]%
        {szegedy2016rethinking}
\bibfield{author}{\bibinfo{person}{Christian Szegedy}, \bibinfo{person}{Vincent
  Vanhoucke}, \bibinfo{person}{Sergey Ioffe}, \bibinfo{person}{Jon Shlens},
  {and} \bibinfo{person}{Zbigniew Wojna}.} \bibinfo{year}{2016}\natexlab{}.
\newblock \showarticletitle{Rethinking the inception architecture for computer
  vision}. In \bibinfo{booktitle}{\emph{CVPR}}. \bibinfo{pages}{2818--2826}.
\newblock


\bibitem[\protect\citeauthoryear{Tao, Gao, Shang, Wu, Zhao, and Yan}{Tao
  et~al\mbox{.}}{2018}]%
        {Tao2018Get}
\bibfield{author}{\bibinfo{person}{Chongyang Tao}, \bibinfo{person}{Shen Gao},
  \bibinfo{person}{Mingyue Shang}, \bibinfo{person}{Wei Wu},
  \bibinfo{person}{Dongyan Zhao}, {and} \bibinfo{person}{Rui Yan}.}
  \bibinfo{year}{2018}\natexlab{}.
\newblock \showarticletitle{Get The Point of My Utterance! Learning Towards
  Effective Responses with Multi-Head Attention Mechanism}. In
  \bibinfo{booktitle}{\emph{IJCAI}}.
\newblock


\bibitem[\protect\citeauthoryear{Tao, Wu, Xu, Hu, Zhao, and Yan}{Tao
  et~al\mbox{.}}{2019b}]%
        {tao2019multi}
\bibfield{author}{\bibinfo{person}{Chongyang Tao}, \bibinfo{person}{Wei Wu},
  \bibinfo{person}{Can Xu}, \bibinfo{person}{Wenpeng Hu},
  \bibinfo{person}{Dongyan Zhao}, {and} \bibinfo{person}{Rui Yan}.}
  \bibinfo{year}{2019}\natexlab{b}.
\newblock \showarticletitle{Multi-Representation Fusion Network for Multi-Turn
  Response Selection in Retrieval-Based Chatbots}. In
  \bibinfo{booktitle}{\emph{WSDM}}. ACM, \bibinfo{pages}{267--275}.
\newblock


\bibitem[\protect\citeauthoryear{Tao, Wu, Xu, Hu, Zhao, and Yan}{Tao
  et~al\mbox{.}}{2019c}]%
        {Tao2019One}
\bibfield{author}{\bibinfo{person}{Chongyang Tao}, \bibinfo{person}{Wei Wu},
  \bibinfo{person}{Can Xu}, \bibinfo{person}{Wenpeng Hu},
  \bibinfo{person}{Dongyan Zhao}, {and} \bibinfo{person}{Rui Yan}.}
  \bibinfo{year}{2019}\natexlab{c}.
\newblock \showarticletitle{One Time of Interaction May Not Be Enough: Go Deep
  with an Interaction-over-Interaction Network for Response Selection in
  Dialogues}. In \bibinfo{booktitle}{\emph{Proceedings of the 57th Annual
  Meeting of the Association for Computational Linguistics}}.
  \bibinfo{publisher}{Association for Computational Linguistics},
  \bibinfo{address}{Florence, Italy}, \bibinfo{pages}{1--11}.
\newblock


\bibitem[\protect\citeauthoryear{Tao, Li, Wang, Fang, Yang, Zhao, and Fu}{Tao
  et~al\mbox{.}}{2019a}]%
        {Tao2019Log2Intent}
\bibfield{author}{\bibinfo{person}{Zhiqiang Tao}, \bibinfo{person}{Sheng Li},
  \bibinfo{person}{Zhaowen Wang}, \bibinfo{person}{Chen Fang},
  \bibinfo{person}{Longqi Yang}, \bibinfo{person}{Handong Zhao}, {and}
  \bibinfo{person}{Yun Fu}.} \bibinfo{year}{2019}\natexlab{a}.
\newblock \showarticletitle{Log2Intent: Towards Interpretable User Modeling via
  Recurrent Semantics Memory Unit}. In \bibinfo{booktitle}{\emph{KDD}}.
\newblock


\bibitem[\protect\citeauthoryear{Vaswani, Shazeer, Parmar, Uszkoreit, Jones,
  Gomez, Kaiser, and Polosukhin}{Vaswani et~al\mbox{.}}{2017}]%
        {vaswani2017attention}
\bibfield{author}{\bibinfo{person}{Ashish Vaswani}, \bibinfo{person}{Noam
  Shazeer}, \bibinfo{person}{Niki Parmar}, \bibinfo{person}{Jakob Uszkoreit},
  \bibinfo{person}{Llion Jones}, \bibinfo{person}{Aidan~N Gomez},
  \bibinfo{person}{{\L}ukasz Kaiser}, {and} \bibinfo{person}{Illia
  Polosukhin}.} \bibinfo{year}{2017}\natexlab{}.
\newblock \showarticletitle{Attention is all you need}. In
  \bibinfo{booktitle}{\emph{NIPS}}. \bibinfo{pages}{5998--6008}.
\newblock


\bibitem[\protect\citeauthoryear{Wang, Wu, Shen, Dick, and van~den Hengel}{Wang
  et~al\mbox{.}}{2017}]%
        {Wang2017Explicit}
\bibfield{author}{\bibinfo{person}{Peng Wang}, \bibinfo{person}{Qi Wu},
  \bibinfo{person}{Chunhua Shen}, \bibinfo{person}{Anthony Dick}, {and}
  \bibinfo{person}{Anton van~den Hengel}.} \bibinfo{year}{2017}\natexlab{}.
\newblock \showarticletitle{Explicit Knowledge-based Reasoning for Visual
  Question Answering}. In \bibinfo{booktitle}{\emph{Proceedings of the
  Twenty-Sixth International Joint Conference on Artificial Intelligence,
  {IJCAI-17}}}. \bibinfo{pages}{1290--1296}.
\newblock


\bibitem[\protect\citeauthoryear{Wang, Yin, Hu, Lian, Wang, and Huang}{Wang
  et~al\mbox{.}}{2018}]%
        {Wang2018Neural}
\bibfield{author}{\bibinfo{person}{Qinyong Wang}, \bibinfo{person}{Hongzhi
  Yin}, \bibinfo{person}{Zhiting Hu}, \bibinfo{person}{Defu Lian},
  \bibinfo{person}{Hao Wang}, {and} \bibinfo{person}{Zi Huang}.}
  \bibinfo{year}{2018}\natexlab{}.
\newblock \showarticletitle{Neural Memory Streaming Recommender Networks with
  Adversarial Training}. In \bibinfo{booktitle}{\emph{KDD}}.
\newblock


\bibitem[\protect\citeauthoryear{Wang and Jiang}{Wang and Jiang}{2017}]%
        {Wang2016ACM}
\bibfield{author}{\bibinfo{person}{Shuohang Wang} {and} \bibinfo{person}{Jing
  Jiang}.} \bibinfo{year}{2017}\natexlab{}.
\newblock \showarticletitle{A Compare-Aggregate Model for Matching Text
  Sequences}.
\newblock \bibinfo{journal}{\emph{ICLR}} (\bibinfo{year}{2017}).
\newblock


\bibitem[\protect\citeauthoryear{Wang, Bovik, Sheikh, Simoncelli,
  et~al\mbox{.}}{Wang et~al\mbox{.}}{2004}]%
        {wang2004image}
\bibfield{author}{\bibinfo{person}{Zhou Wang}, \bibinfo{person}{Alan~C Bovik},
  \bibinfo{person}{Hamid~R Sheikh}, \bibinfo{person}{Eero~P Simoncelli},
  {et~al\mbox{.}}} \bibinfo{year}{2004}\natexlab{}.
\newblock \showarticletitle{Image quality assessment: from error visibility to
  structural similarity}.
\newblock \bibinfo{journal}{\emph{IEEE transactions on image processing}}
  \bibinfo{volume}{13}, \bibinfo{number}{4} (\bibinfo{year}{2004}),
  \bibinfo{pages}{600--612}.
\newblock


\bibitem[\protect\citeauthoryear{Wu, Liu, Wang, and Dong}{Wu
  et~al\mbox{.}}{2018a}]%
        {Wu2018ChainOR}
\bibfield{author}{\bibinfo{person}{Chenfei Wu}, \bibinfo{person}{Jinlai Liu},
  \bibinfo{person}{Xiaojie Wang}, {and} \bibinfo{person}{Xuan Dong}.}
  \bibinfo{year}{2018}\natexlab{a}.
\newblock \showarticletitle{Chain of Reasoning for Visual Question Answering}.
\newblock In \bibinfo{booktitle}{\emph{Advances in Neural Information
  Processing Systems 31}}, \bibfield{editor}{\bibinfo{person}{S.~Bengio},
  \bibinfo{person}{H.~Wallach}, \bibinfo{person}{H.~Larochelle},
  \bibinfo{person}{K.~Grauman}, \bibinfo{person}{N.~Cesa-Bianchi}, {and}
  \bibinfo{person}{R.~Garnett}} (Eds.). \bibinfo{publisher}{Curran Associates,
  Inc.}, \bibinfo{pages}{275--285}.
\newblock


\bibitem[\protect\citeauthoryear{Wu, Liu, Wang, and Dong}{Wu
  et~al\mbox{.}}{2018b}]%
        {Wu2018ObjectDifferenceAA}
\bibfield{author}{\bibinfo{person}{Chenfei Wu}, \bibinfo{person}{Jinlai Liu},
  \bibinfo{person}{Xiaojie Wang}, {and} \bibinfo{person}{Xuan Dong}.}
  \bibinfo{year}{2018}\natexlab{b}.
\newblock \showarticletitle{Object-Difference Attention: A Simple Relational
  Attention for Visual Question Answering}. In
  \bibinfo{booktitle}{\emph{Proceedings of the 26th ACM International
  Conference on Multimedia}} (Seoul, Republic of Korea)
  \emph{(\bibinfo{series}{MM ’18})}. \bibinfo{publisher}{Association for
  Computing Machinery}, \bibinfo{address}{New York, NY, USA},
  \bibinfo{pages}{519–527}.
\newblock
\showISBNx{9781450356657}


\bibitem[\protect\citeauthoryear{Wu, Wu, An, Qi, Huang, Huang, and Xie}{Wu
  et~al\mbox{.}}{2019b}]%
        {Wu2019Neural}
\bibfield{author}{\bibinfo{person}{Chuhan Wu}, \bibinfo{person}{Fangzhao Wu},
  \bibinfo{person}{Mingxiao An}, \bibinfo{person}{Tao Qi},
  \bibinfo{person}{Jianqiang Huang}, \bibinfo{person}{Yongfeng Huang}, {and}
  \bibinfo{person}{Xing Xie}.} \bibinfo{year}{2019}\natexlab{b}.
\newblock \showarticletitle{Neural News Recommendation with Heterogeneous User
  Behavior}. In \bibinfo{booktitle}{\emph{EMNLP}}.
\newblock


\bibitem[\protect\citeauthoryear{Wu, Socher, and Xiong}{Wu
  et~al\mbox{.}}{2019a}]%
        {Wu2019Globaltolocal}
\bibfield{author}{\bibinfo{person}{Chien-Sheng Wu}, \bibinfo{person}{Richard
  Socher}, {and} \bibinfo{person}{Caiming Xiong}.}
  \bibinfo{year}{2019}\natexlab{a}.
\newblock \showarticletitle{Global-to-local Memory Pointer Networks for
  Task-Oriented Dialogue}. In \bibinfo{booktitle}{\emph{ICLR}}.
\newblock


\bibitem[\protect\citeauthoryear{Wu, Wang, Shen, Dick, and van~den Hengel}{Wu
  et~al\mbox{.}}{2016}]%
        {wu2016ask}
\bibfield{author}{\bibinfo{person}{Qi Wu}, \bibinfo{person}{Peng Wang},
  \bibinfo{person}{Chunhua Shen}, \bibinfo{person}{Anthony Dick}, {and}
  \bibinfo{person}{Anton van~den Hengel}.} \bibinfo{year}{2016}\natexlab{}.
\newblock \showarticletitle{Ask me anything: Free-form visual question
  answering based on knowledge from external sources}. In
  \bibinfo{booktitle}{\emph{CVPR}}. \bibinfo{pages}{4622--4630}.
\newblock


\bibitem[\protect\citeauthoryear{Wu, Wang, Shen, Reid, and van~den Hengel}{Wu
  et~al\mbox{.}}{2018c}]%
        {wu2018you}
\bibfield{author}{\bibinfo{person}{Qi Wu}, \bibinfo{person}{Peng Wang},
  \bibinfo{person}{Chunhua Shen}, \bibinfo{person}{Ian Reid}, {and}
  \bibinfo{person}{Anton van~den Hengel}.} \bibinfo{year}{2018}\natexlab{c}.
\newblock \showarticletitle{Are you talking to me? reasoned visual dialog
  generation through adversarial learning}. In
  \bibinfo{booktitle}{\emph{ICCV}}. \bibinfo{pages}{6106--6115}.
\newblock


\bibitem[\protect\citeauthoryear{Wu, Wu, Xing, Zhou, and Li}{Wu
  et~al\mbox{.}}{2017}]%
        {Wu2017SequentialMN}
\bibfield{author}{\bibinfo{person}{Yu Wu}, \bibinfo{person}{Wei Wu},
  \bibinfo{person}{Chen Xing}, \bibinfo{person}{Ming Zhou}, {and}
  \bibinfo{person}{Zhoujun Li}.} \bibinfo{year}{2017}\natexlab{}.
\newblock \showarticletitle{Sequential Matching Network: A New Architecture for
  Multi-turn Response Selection in Retrieval-Based Chatbots}. In
  \bibinfo{booktitle}{\emph{ACL}}. \bibinfo{publisher}{Association for
  Computational Linguistics}, \bibinfo{address}{Vancouver, Canada},
  \bibinfo{pages}{496--505}.
\newblock


\bibitem[\protect\citeauthoryear{Xie, Liu, Yan, and Sun}{Xie
  et~al\mbox{.}}{2016}]%
        {xie2016neural}
\bibfield{author}{\bibinfo{person}{Ruobing Xie}, \bibinfo{person}{Zhiyuan Liu},
  \bibinfo{person}{Rui Yan}, {and} \bibinfo{person}{Maosong Sun}.}
  \bibinfo{year}{2016}\natexlab{}.
\newblock \showarticletitle{Neural emoji recommendation in dialogue systems}.
\newblock \bibinfo{journal}{\emph{arXiv preprint arXiv:1612.04609}}
  (\bibinfo{year}{2016}).
\newblock


\bibitem[\protect\citeauthoryear{Xiong, Merity, and Socher}{Xiong
  et~al\mbox{.}}{2016a}]%
        {Xiong2016DynamicMN}
\bibfield{author}{\bibinfo{person}{Caiming Xiong}, \bibinfo{person}{Stephen
  Merity}, {and} \bibinfo{person}{Richard Socher}.}
  \bibinfo{year}{2016}\natexlab{a}.
\newblock \showarticletitle{Dynamic Memory Networks for Visual and Textual
  Question Answering}.
\newblock \bibinfo{journal}{\emph{ArXiv}}  \bibinfo{volume}{abs/1603.01417}
  (\bibinfo{year}{2016}).
\newblock


\bibitem[\protect\citeauthoryear{Xiong, Merity, and Socher}{Xiong
  et~al\mbox{.}}{2016b}]%
        {xiong2016dynamic}
\bibfield{author}{\bibinfo{person}{Caiming Xiong}, \bibinfo{person}{Stephen
  Merity}, {and} \bibinfo{person}{Richard Socher}.}
  \bibinfo{year}{2016}\natexlab{b}.
\newblock \showarticletitle{Dynamic memory networks for visual and textual
  question answering}. In \bibinfo{booktitle}{\emph{ICML}}.
  \bibinfo{pages}{2397--2406}.
\newblock


\bibitem[\protect\citeauthoryear{Yan, Song, and Wu}{Yan et~al\mbox{.}}{2016}]%
        {Yan2016LearningTR}
\bibfield{author}{\bibinfo{person}{Rui Yan}, \bibinfo{person}{Yiping Song},
  {and} \bibinfo{person}{Hua Wu}.} \bibinfo{year}{2016}\natexlab{}.
\newblock \showarticletitle{Learning to Respond with Deep Neural Networks for
  Retrieval-Based Human-Computer Conversation System}. In
  \bibinfo{booktitle}{\emph{Proceedings of the 39th International ACM SIGIR
  Conference on Research and Development in Information Retrieval}} (Pisa,
  Italy) \emph{(\bibinfo{series}{SIGIR '16})}. \bibinfo{publisher}{Association
  for Computing Machinery}, \bibinfo{address}{New York, NY, USA},
  \bibinfo{pages}{55--64}.
\newblock
\showISBNx{9781450340694}


\bibitem[\protect\citeauthoryear{Yan and Zhao}{Yan and Zhao}{2018}]%
        {Yan2018Coupled}
\bibfield{author}{\bibinfo{person}{Rui Yan} {and} \bibinfo{person}{Dongyan
  Zhao}.} \bibinfo{year}{2018}\natexlab{}.
\newblock \showarticletitle{Coupled Context Modeling for Deep Chit-Chat:
  Towards Conversations between Human and Computer}. In
  \bibinfo{booktitle}{\emph{Proceedings of the 24th ACM SIGKDD International
  Conference on Knowledge Discovery \& Data Mining}} (London, United Kingdom)
  \emph{(\bibinfo{series}{KDD '18})}. \bibinfo{publisher}{Association for
  Computing Machinery}, \bibinfo{address}{New York, NY, USA},
  \bibinfo{pages}{2574--2583}.
\newblock
\showISBNx{9781450355520}


\bibitem[\protect\citeauthoryear{Yan, Zhao, and E.}{Yan et~al\mbox{.}}{2017}]%
        {Yan2017Joint}
\bibfield{author}{\bibinfo{person}{Rui Yan}, \bibinfo{person}{Dongyan Zhao},
  {and} \bibinfo{person}{Weinan E.}} \bibinfo{year}{2017}\natexlab{}.
\newblock \showarticletitle{Joint Learning of Response Ranking and Next
  Utterance Suggestion in Human-Computer Conversation System}. In
  \bibinfo{booktitle}{\emph{Proceedings of the 40th International ACM SIGIR
  Conference on Research and Development in Information Retrieval}} (Shinjuku,
  Tokyo, Japan) \emph{(\bibinfo{series}{SIGIR '17})}.
  \bibinfo{publisher}{Association for Computing Machinery},
  \bibinfo{address}{New York, NY, USA}, \bibinfo{pages}{685--694}.
\newblock
\showISBNx{9781450350228}


\bibitem[\protect\citeauthoryear{Yang, Yan, Yu, Li, and Chiu}{Yang
  et~al\mbox{.}}{2017}]%
        {Yang2017Multi}
\bibfield{author}{\bibinfo{person}{Chunfeng Yang}, \bibinfo{person}{Huan Yan},
  \bibinfo{person}{Donghan Yu}, \bibinfo{person}{Yong Li}, {and}
  \bibinfo{person}{Dah~Ming Chiu}.} \bibinfo{year}{2017}\natexlab{}.
\newblock \showarticletitle{Multi-site User Behavior Modeling and Its
  Application in Video Recommendation}. In \bibinfo{booktitle}{\emph{SIGIR}}.
\newblock


\bibitem[\protect\citeauthoryear{Yu, Lian, Mahmoody, Liu, and Xie}{Yu
  et~al\mbox{.}}{2019}]%
        {Yu2019Adaptive}
\bibfield{author}{\bibinfo{person}{Zeping Yu}, \bibinfo{person}{Jianxun Lian},
  \bibinfo{person}{Ahmad Mahmoody}, \bibinfo{person}{Gongshen Liu}, {and}
  \bibinfo{person}{Xing Xie}.} \bibinfo{year}{2019}\natexlab{}.
\newblock \showarticletitle{Adaptive User Modeling with Long and Short-Term
  Preferences for Personalized Recommendation}. In
  \bibinfo{booktitle}{\emph{IJCAI}}.
\newblock


\bibitem[\protect\citeauthoryear{Zhao, Liu, Chao, and Qian}{Zhao
  et~al\mbox{.}}{2020}]%
        {Zhao2020CAPERCP}
\bibfield{author}{\bibinfo{person}{Guoshuai Zhao}, \bibinfo{person}{Zhidan
  Liu}, \bibinfo{person}{Yulu Chao}, {and} \bibinfo{person}{Xueming Qian}.}
  \bibinfo{year}{2020}\natexlab{}.
\newblock \showarticletitle{CAPER: Context-Aware Personalized Emoji
  Recommendation}.
\newblock \bibinfo{journal}{\emph{IEEE Transactions on Knowledge and Data
  Engineering}} (\bibinfo{year}{2020}), \bibinfo{pages}{1--1}.
\newblock


\bibitem[\protect\citeauthoryear{Zhou, Dong, Wu, Zhao, Yu, Tian, Liu, and
  Yan}{Zhou et~al\mbox{.}}{2016}]%
        {zhou2016multi}
\bibfield{author}{\bibinfo{person}{Xiangyang Zhou}, \bibinfo{person}{Daxiang
  Dong}, \bibinfo{person}{Hua Wu}, \bibinfo{person}{Shiqi Zhao},
  \bibinfo{person}{Dianhai Yu}, \bibinfo{person}{Hao Tian},
  \bibinfo{person}{Xuan Liu}, {and} \bibinfo{person}{Rui Yan}.}
  \bibinfo{year}{2016}\natexlab{}.
\newblock \showarticletitle{Multi-view Response Selection for Human-Computer
  Conversation}. In \bibinfo{booktitle}{\emph{EMNLP '16}}.
  \bibinfo{publisher}{Association for Computational Linguistics},
  \bibinfo{address}{Austin, Texas}, \bibinfo{pages}{372--381}.
\newblock


\bibitem[\protect\citeauthoryear{Zhou, Li, Dong, Liu, Chen, Zhao, Yu, and
  Wu}{Zhou et~al\mbox{.}}{2018}]%
        {zhou2018multi}
\bibfield{author}{\bibinfo{person}{Xiangyang Zhou}, \bibinfo{person}{Lu Li},
  \bibinfo{person}{Daxiang Dong}, \bibinfo{person}{Yi Liu},
  \bibinfo{person}{Ying Chen}, \bibinfo{person}{Wayne~Xin Zhao},
  \bibinfo{person}{Dianhai Yu}, {and} \bibinfo{person}{Hua Wu}.}
  \bibinfo{year}{2018}\natexlab{}.
\newblock \showarticletitle{Multi-turn response selection for chatbots with
  deep attention matching network}. In \bibinfo{booktitle}{\emph{ACL}},
  Vol.~\bibinfo{volume}{1}. \bibinfo{pages}{1118--1127}.
\newblock


\bibitem[\protect\citeauthoryear{Zhou, Mascolo, and Zhao}{Zhou
  et~al\mbox{.}}{2019}]%
        {Zhou2019TopicEnhanced}
\bibfield{author}{\bibinfo{person}{Xiao Zhou}, \bibinfo{person}{Cecilia
  Mascolo}, {and} \bibinfo{person}{Zhongxiang Zhao}.}
  \bibinfo{year}{2019}\natexlab{}.
\newblock \showarticletitle{Topic-Enhanced Memory Networks for Personalised
  Point-of-Interest Recommendation}. In \bibinfo{booktitle}{\emph{KDD}}.
\newblock


\bibitem[\protect\citeauthoryear{Zhou and Wang}{Zhou and Wang}{2018}]%
        {Zhou2018MojiTalk}
\bibfield{author}{\bibinfo{person}{Xianda Zhou} {and}
  \bibinfo{person}{William~Yang Wang}.} \bibinfo{year}{2018}\natexlab{}.
\newblock \showarticletitle{MojiTalk: Generating Emotional Responses at Scale}.
  In \bibinfo{booktitle}{\emph{ACL}}.
\newblock


\bibitem[\protect\citeauthoryear{Zhu, Cao, Liu, Yang, Ying, and Xiong}{Zhu
  et~al\mbox{.}}{2020}]%
        {Zhu2020Sequential}
\bibfield{author}{\bibinfo{person}{Nengjun Zhu}, \bibinfo{person}{Jian Cao},
  \bibinfo{person}{Yanchi Liu}, \bibinfo{person}{Yang Yang},
  \bibinfo{person}{Haochao Ying}, {and} \bibinfo{person}{Hui Xiong}.}
  \bibinfo{year}{2020}\natexlab{}.
\newblock \showarticletitle{Sequential Modeling of Hierarchical User Intention
  and Preference for Next-item Recommendation}. In
  \bibinfo{booktitle}{\emph{WSDM}}.
\newblock


\bibitem[\protect\citeauthoryear{Zhu, Li, Liao, Wang, Guan, Liu, and Cai}{Zhu
  et~al\mbox{.}}{2017}]%
        {Zhu2017What}
\bibfield{author}{\bibinfo{person}{Yu Zhu}, \bibinfo{person}{Hao Li},
  \bibinfo{person}{Yikang Liao}, \bibinfo{person}{Beidou Wang},
  \bibinfo{person}{Ziyu Guan}, \bibinfo{person}{Haifeng Liu}, {and}
  \bibinfo{person}{Deng Cai}.} \bibinfo{year}{2017}\natexlab{}.
\newblock \showarticletitle{What to Do Next: Modeling User Behaviors by
  Time-LSTM}. In \bibinfo{booktitle}{\emph{IJCAI}}.
\newblock


\bibitem[\protect\citeauthoryear{Żołna and Romański}{Żołna and
  Romański}{2017}]%
        {zolna2017User}
\bibfield{author}{\bibinfo{person}{Konrad Żołna} {and}
  \bibinfo{person}{Bartłomiej Romański}.} \bibinfo{year}{2017}\natexlab{}.
\newblock \showarticletitle{User Modeling Using LSTM Networks}. In
  \bibinfo{booktitle}{\emph{AAAI}}.
\newblock


\end{thebibliography}

\end{document}